\documentclass[final,5p,times,twocolumn]{elsarticle}    

\usepackage{amssymb}
\usepackage{graphicx}
\usepackage{xurl}
\usepackage{color}
\usepackage{float}

\usepackage{amsmath,amsfonts}
\usepackage{enumitem}
\usepackage{multirow}
\usepackage{subfig}
\usepackage{tikz,forest}
\usepackage{booktabs}
\usepackage{pdflscape}
\usepackage{array}
\usepackage{makecell}
\usepackage{caption}
\usepackage{comment}
\usepackage[makeroom]{cancel}
\usepackage{ulem} 
\usepackage{hyperref}

\newcommand{\PreserveBackslash}[1]{\let\temp=\\#1\let\\=\temp}
\newcolumntype{C}[1]{>{\PreserveBackslash\centering}p{#1}}
\newcolumntype{R}[1]{>{\PreserveBackslash\raggedleft}p{#1}}
\newcolumntype{L}[1]{>{\PreserveBackslash\raggedright}p{#1}}

\usepackage{tikz}

\definecolor{darkblue}{rgb}{0.0, 0.0, 0.55}
\definecolor{darkcyan}{rgb}{0.0, 0.55, 0.55}

\journal{Neurocomputing}

\makeatletter
\def\ps@pprintTitle{%
    \let\@oddhead\@empty
    \let\@evenhead\@empty
    \let\@oddfoot\@empty
    \let\@evenfoot\@empty
}
\makeatother

\begin{document}

\begin{frontmatter}

\title{Using Offline Data to Speed Up Reinforcement Learning in Procedurally Generated Environments\tnoteref{label1}}

\tnotetext[label1]{This paper has been accepted for publication in \textbf{Neurocomputing}, \\DOI: \url{https://doi.org/10.1016/j.neucom.2024.129079}.}

\author[tecn]{Alain Andres\corref{cor1}} 
\author[uoe]{Lukas Schäfer}
\author[uoe]{Stefano V. Albrecht}
\author[tecn,upv]{Javier Del Ser}

\cortext[cor1]{Corresponding author: alain.andres@tecnalia.com}

\affiliation[tecn]{organization={TECNALIA, Basque Research and Technology Alliance (BRTA)},
            city={Donostia-San Sebastian},
            postcode={20009}, 
            country={Spain}}
\affiliation[upv]{organization={University of the Basque Country (UPV/EHU)},
            city={Bilbao},
            postcode={48013}, 
            country={Spain}}
\affiliation[uoe]{organization={University of Edinburgh},
            addressline={Informatics Forum}, 
            city={Edinburgh},
            postcode={EH8 9AB}, 
            country={United Kingdom}}

\begin{abstract}
One of the key challenges of \textit{Reinforcement Learning} (RL) is the ability of an agent to generalize its learned policy to unseen settings. Moreover, training an RL agent requires large numbers of interactions with the environment. Motivated by the success of \textit{Imitation Learning} (IL), we conduct a study to investigate whether an agent can leverage offline data in the form of trajectories to improve the sample-efficiency in procedurally generated environments. We consider two settings of using IL from offline data for RL: (1) pre-training a policy before online RL training and (2) concurrently training a policy with online RL and IL from offline data. We analyze the impact of the quality (optimality of trajectories), quantity and diversity of available offline trajectories on the effectiveness of both approaches. Across four well-known sparse reward tasks in the MiniGrid environment, we find that using IL for both pre-training and concurrently during online RL training, consistently improves sample-efficiency, and in some tasks achieves higher returns compared to using either IL or RL alone. Furthermore, we show that training a policy from as few as two trajectories can make the difference between learning an optimal policy at the end of online training and not learning at all. 
Evaluation in two tasks of the Procgen environment further highlights that the diversity of the training data is more important than its quality. Our findings motivate the widespread adoption of IL for pre-training and concurrent IL in procedurally generated environments whenever offline trajectories are available or can be generated.

\end{abstract}

\end{frontmatter}


\section{Introduction}

\textit{Reinforcement Learning} (RL) is widely used for sequential decision making in various fields, including healthcare ~\cite{gottesman_guidelines_2019}, energy~\cite{fu_applications_2022} and robotics ~\cite{zhu_dexterous_2018}. Traditionally, RL algorithms are trained and evaluated in the same single task, with the goal of maximizing the cumulative reward over time. However, the variability of real-world problems poses a challenge for these agents, as they may not generalize well to new (unseen) scenarios \cite{reed_generalist_2022}. 
This generalization challenge is critical in applications such as industry, where RL models are used for real-time scheduling problems ~\cite{ECHEVERRIA2025109488} or for accurately distinguishing test and control lines in gold immunochromatographic strip images~\cite{zeng2021deep}. Similarly, in robotics, RL-driven control systems for tasks like manipulation ~\cite{10611477} and grasping ~\cite{kalashnikov2018scalable} must generalize across different scenarios, robotic assets, and object types to ensure safe and reliable performance in real-world deployments.
To address this issue, recent research in RL has studied the ability of agents to generalize to varying yet similar tasks that can differ in either the state space, dynamics of the environment, the agent's action space and/or even the reward function \cite{kirk_survey_2022}. 

One way of evaluating the generalization capability of agents is by training agents in procedurally content generated (PCG) environments. Any PCG task constitutes a set of levels over which the learned policy has to generalize. 
Completing the levels of a single task requires a common skill,
but may, for example, vary in the agent's initial location, the layout of its environment, colors and locations of objects the agent can interact with. Such variability prevents the agent from memorizing specific trajectories (overfitting) ~\cite{cobbe_leveraging_2020}; instead, PCG environments force the agent to learn relevant representations and policies which effectively generalize across all levels of a task. 

However, PCG environments often require large a\-mounts of interactions to train an effective policy \cite{jiang_prioritized_2021}. 
Our work finds its motivation in the availability of offline data in many real-world settings, where one of the main objectives is to decrease the number of agent-environment interactions due to economic, safety and time constraints. Specifically, we study the effectiveness of using \textit{Imitation Learning} (IL) on offline data to improve the converged performance and sample-efficiency of an RL agent in PCG environments. The contributions of our work are threefold:
\begin{enumerate}
    \item We analyze how offline data can be used to pre-train a policy to kickstart the learning of an agent.
    \item We study how offline data can be combined with the online collected experiences for concurrent IL and RL training to improve sample efficiency.
    \item We investigate how the quality, quantity and diversity of the offline data affects the learning process.
\end{enumerate}

We collect a dataset of offline trajectories by training an agent with a self-IL approach specifically designed for PCG environments (RAPID ~\cite{zha_rank_2021}) and 
storing the best trajectories seen so far at three checkpoints during training.
Each of these three datasets contains trajectories of varying quality as measured by the performance of the policy at that point during the agent's learning process. We use \textit{Behavior Cloning} (BC) as a form of IL to train the agent on this data before (\textit{pre-training}) and concurrently to the online RL training. Finally, we examine the results in various tasks of two PCG benchmarks: MiniGrid~\cite{minigrid} and Procgen~\cite{cobbe_leveraging_2020}.
Our results demonstrate that leveraging offline data significantly reduces the number of interactions required to learn an optimal policy across all analyzed tasks. Specifically, we arrive at the following novel insights with respect to the state of the art in sample efficiency, offline data and IL (later reviewed in Section \ref{sec:relatedwork}): 
\begin{itemize}[leftmargin=*]
    \item Pre-training with offline data provides a strong initialization policy, effectively kickstarting the agent’s learning process and enhancing early performance. 
    \item Concurrently training with IL and RL further improves sample efficiency and robustness, allowing the agent to reach optimal performance with fewer interactions. 
    \item Most importantly, our results reveal that diversity in demonstration trajectories is more beneficial for generalization than high-quality, near-optimal trajectories. Training with a wider range of experiences across levels better equips the agent to handle new, unseen scenarios. 
\end{itemize}

These findings offer novel perspectives into the role of demonstration diversity in generalization, extending beyond the traditional focus on demonstration quality. In practice, our approach provides a more efficient framework for RL in real-world applications such as robotics, energy management, and industrial automation, where interaction costs are high and environments resemble PCG setups, with variable conditions that make generalization a mandatory capability for the learned agents.

The rest of the manuscript is structured as follows: Section \ref{sec:relatedwork} revisits the literature related to sample efficiency, offline data for RL and IL, whereas Section \ref{sec:background} poses fundamental concepts in RL, PCG and IL. Next, we elaborate on the specific PCG benchmarks and their generalization requirements (Section \ref{subsec:environment}), and describe the data collection and learning methodology considered in our study (Section \ref{sec:methodology}). Results of the performed experiments are presented and discussed in Section \ref{subsec:results}. Finally, conclusions are drawn in Section \ref{sec:conclus}, together with an outline of future research directions.

\section{Related Work} \label{sec:relatedwork}

This section revisits briefly different topics of relevance for our study,
including the sample efficiency in PCG environments (Section \ref{sec:sampleefficiency}), the use of offline data for RL (Section \ref{sec:offlinedata}) and IL (Section \ref{sec:ILreview}). Finally, Section \ref{sec:contrib} builds upon the reviewed literature to expose the contribution of this manuscript to the field.

\subsection{Sample-efficiency in PCG environments} \label{sec:sampleefficiency}

Off-policy algorithms are naturally suitable to make use of data collected by an arbitrary behavior policy, and are often found to be more sample efficient in the number of agent-environment interactions due to the application of a replay buffer~\cite{doro_sample-efficient_2023}. However, they can exhibit larger instabilities and tend to be more sensitive to hyperparameters than on-policy solutions \cite{gu_q-prop_2017}. These issues are further exacerbated in PCG environments \cite{ehrenberg_study_2022}, where comparably little research exists using off-policy (e.g. DQN~\cite{mnih_human-level_2015}, SAC~\cite{haarnoja_soft_2018}) algorithms in comparison with on-policy algorithms (e.g. PPO~\cite{schulman_proximal_2017}, IMPALA ~\cite{espeholt_impala_2018}, PPG~\cite{cobbe_phasic_2020}). In fact, off-policy algorithms have only been applied to solve tasks that are comparably easily solved by on-policy algorithms~\cite{seurin_im_2020,kessler_same_2022,nguyen_leveraging_2022}. Therefore, a large amount of algorithmic approaches have been focused on how to improve the sample-efficiency of on-policy algorithms by incentivizing exploration with either intrinsic rewards that model the curiosity~\cite{raileanu_ride_2020,zhang_bebold_2020,flet-berliac_adversarially_2021,schafer_decoupled_2022} or using self-IL techniques~\cite{zha_rank_2021,pshikhachev_self-imitation_2021,andres_towards_2022,andres_enhanced_2023,lin_taking_2024} which augment online RL training with BC from self-collected trajectories.


\subsection{Offline Data for RL} \label{sec:offlinedata}
One way to enhance the sample-efficiency and performance of a RL agent is by enabling them to learn from existing datasets (with no online interactions). Such data can be gathered in various ways, including supervision of a human in order to maximize the return of demonstrations \cite{liu_curriculum_2022}, without supervision while trying to maximize the data coverage~\cite{liu_aps_2021} or the discovery of skills~\cite{liu_behavior_2021}, among others. 

Such offline data has been shown effective in offline-to-online RL settings \cite{xie_pretraining_2022}, where the offline data is used to pre-train a given policy, and then the policy is further fine-tuned during the online learning stage~\cite{nair_awac_2021,lu_aw-opt_2021,song_hybrid_2022,zhang_policy_2023,wexler_analyzing_2022}. However, its success is subject to the distribution shift between the offline data and the data collected during the online interaction with the environment. In fact, distribution shift can be minimized using prioritization techniques~\cite{yue_boosting_2022,gupta_can_2022} or enforcing constraints on the learned policy~\cite{kumar_conservative_2020,kostrikov_offline_2021,wexler_analyzing_2022}. However, these works rely on off-policy RL solutions which have been shown to be ineffective in many PCG tasks~\cite{ehrenberg_study_2022,cobbe_leveraging_2020,mohanty_measuring_2021}.

\subsubsection{Imitation Learning for Reinforcement Learning} \label{sec:ILreview}

IL offers a simplified approach to utilizing offline data, requiring only state and action pairs. This method focuses on mimicking demonstrated behaviors, often thro\-ugh supervised learning techniques like BC \cite{nair_awac_2021,rajeswaran_learning_2018,gupta_relay_2019,liu_curriculum_2022}, although other non-supervised paradigms might also be adopted \cite{ho_generative_2016,reddy_sqil_2019}. However, its effectiveness is sensitive to the quality of the data as given by its optimality~\cite{kumar_when_2022}. Moreover, IL faces issues regarding the distribution shift between the provided offline data and the data collected during online execution ~\cite{nair_awac_2021,zhang_policy_2023}. Nevertheless, this distribution shift is further complicated when considering PCG environments, where providing demonstrations can overfit the agent to solve some levels that do not match with the generalization requirements of the entire level distribution.

\subsection{Our Contribution} \label{sec:contrib}
Although previous approaches combined IL (using offline data) with on-policy RL (applied through online gathered data), studies focusing on these techniques within PCG environments -- where the agent generalization is critical -- are scarce. 
Concurrently with our work, \cite{xu_improved_2023} investigated a similar setup in the MiniGrid benchmark. However, they applied an additional self-IL loss with prioritization~\cite{oh_self-imitation_2018}, resulting in more frequent updates delivered to the agent. In \cite{jia2022improving}, a PPO agent is first trained in Procgen until reaching a learning convergence plateau, after which demonstrations from specific level subsets are used with IL to further enhance its overall performance. Additionally, \cite{mediratta2024gengap} has recently highlighted the performance differences between PPO, BC, and offline RL methods in Procgen, showing that online methods outperform offline strategies when generalization is needed. However, they do not examine the impact of combining PPO with BC.

Differentially, in this study we rigorously investigate the effectiveness of combining IL with RL, focusing on variations in the quality, quantity, and diversity of the offline demonstrations. While our approach can be applies to any general RL algorithm, we concentrate on on-policy algorithms -- specifically PPO, which has demonstrated the best performance in prior studies. These attributes play a pivotal role in leveraging the benefits of IL, particularly in PCG environments where achieving generalization is not merely advantageous, but necessary. Importantly, our research explores the relevance of these demonstrations attributes when applying IL either for pre-training, or when being integrated concurrently with RL during the online phase. This two-fold focus allows drawing insights into the optimization of sample efficiency and robustness in highly variable PCG environments.

\section{Background} \label{sec:background}
\subsection{Partially Observable Markov Decision Process}
We define a RL problem as a Markov Decision Process (MDP) given by a tuple \{{$\mathcal{S},\mathcal{A},\mathcal{P},\mathcal{R}, \gamma$}\}, where $\mathcal{S}$ represents the state space, $\mathcal{A}$ is the action space, $\mathcal{P}: \mathcal{S}\times\mathcal{A}\times\mathcal{S} \rightarrow [0,1]$ is the state-transition probability function, $\mathcal{R}:\mathcal{S}\times \mathcal{A} \times \mathcal{S} \rightarrow \mathbb{R}$ represents the reward function, and $\gamma \in [0,1)$ denotes the discount factor. At every time step $t$, the agent observes a state $s_t \in \mathcal{S}$ and selects an action $a_t$ sampled from its policy $a_t \sim \pi(\cdot|s_t)$. Given the current state $s_t$ and selected action $a_t$ the environment transitions to a new state $s_{t+1} \sim \mathcal{P}(s_t,a_t)$ and the agent receives a reward $r_t=\mathcal{R}(s_t,a_t,s_{t+1})$. In partially observable environments where the agent might only observe a part of the state, the environment can be formalized as a Partially Observable Markov Decision Process (POMPD) ~\cite{kaelbling_planning_1998}. A POMDP extends the MPD formalism to a 7-tuple \{{$\mathcal{S},\mathcal{A},\mathcal{P},\mathcal{R}, \gamma,\mathcal{O}, \Omega$}\} where $\Omega$ represents the observation space and $\mathcal{O}: \mathcal{S} \times \mathcal{A} \times \Omega \rightarrow [0,1]$ represents the observation function that maps a state and action to a distribution over observations. In a POMDP, the agent only receives observations 
$o_t \sim \mathcal{O}(s_t, a_t)$ based on the current state and selected action, and conditions its policy on the episodic history of observations.

\subsection{Procedural Content Generation} \label{subsec:pcg}
In this work, we focus on procedurally generated environments which require agents to learn policies that generalize across a collection of levels that optimize a given objective.

Formally, a task $T$ is composed of a collection of different levels $l \in \mathcal{L}(T)$, where each level is considered a POMDP and $\mathcal{L}(T)$ represents the whole distribution of levels for task $T$. The levels are generated with a seed, ID or parameter vector that makes them differ from other levels with respect to their state spaces $\mathcal{S}$ and observation spaces $\Omega$~\cite{kirk_survey_2022}. 

Unlike traditional environments, where the agent's goal is to maximize the return in a static setting with limited variation between episodes (e.g., minor changes like the initial state $s_0$ or the location of the goal), PCG tasks introduce diversity by sampling levels from a broader distribution $\mathcal{L}(T)$. 
This requires agents to generalize across $\mathcal{L}(T)$ rather than overfitting to specific levels, necessitating exploration strategies that can adapt to varying layouts and objectives.

As a consequence, this inherent diversity in PCG environments redefines the agent's learning objective. Instead of optimizing the actions to be taken in a single scenario, the agent must maximize the expected discounted returns over the whole level distribution, i.e.:
\begin{equation}
    \mathbb{E}_{\mathcal{L}(T)}[G_t] = \mathbb{E}_{\mathcal{L}(T)}\left[\sum_{t=0}^N \gamma^t \mathcal{R}(s_t, a_t,s_{t+1})\right]
\end{equation}
where $N$ is the episode length and $\mathcal{R}(s_t,a_t,s_{t+1})$ is the reward at time step $t$. $G_t$ denotes the discounted return after time step $t$.


\subsection{Imitation Learning} \label{subsec:il}
Imitation Learning can be applied in several ways. We adopt BC using the log loss function (also called \textit{cross-entropy loss})~\cite{osa_algorithmic_2018} for discrete actions:
\begin{equation}\label{eq:bc}
    L_{BC} =  - \frac{1}{|B|} \sum_{(s, a) \sim B} \ln(\pi(a|s))
\end{equation}
where $B$ is a batch of state action pairs 
$\{s,a\}$
containing experiences to be imitated, $|B|$ denotes its size, and $\pi$ is the policy being trained, with $\pi(a|s)$ indicating the probability of selecting action $a$ in state $s$. Prior works combine this BC loss with the online RL loss for a single backpropagation and optimization objective~\cite{hester_deep_2017,vecerik_leveraging_2018}, whereas we separate the optimization for BC and RL~\cite{zha_rank_2021,andres_towards_2022}\footnote{This separation does not affect pre-training with IL because no RL updates are computed in that stage.}. This allows controlling the number of optimization steps and the learning frequency of each optimization objective. 

\subsubsection{Self-Imitation Learning}\label{sec:sil}

When expert data is not available, the agent can be trained with self-Imitation Learning (self-IL). This para\-digm attempts to learn a policy based on past successful trajectories collected by the agent itself~\cite{oh_self-imitation_2018,zha_rank_2021,andres_towards_2022}, so that the agent can improve its behavior with actions that led to promising outcomes. RAPID~\cite{zha_rank_2021} determines the success of a trajectory based on the following weighted score:
\begin{equation}\label{eq:rapid_scores}
    S = w_0\cdot S_{ext} + w_1\cdot S_{local} + w_2\cdot S_{global}
\end{equation}
where $S_{ext}$ refers to the extrinsic return of the episode, $S_{local}$ represents the diversity of states within the episo\-de, and $S_{global}$ represents the long-term exploration as given by state visitation counts~\cite{bellemare_unifying_2016,ostrovski_count-based_2017}. RAPID ranks trajectories based on their weighted score and stores the trajectories with highest scores in a replay buffer. Throughout training, a random batch of trajectories is sampled uniformly at random and the BC loss is minimized for the given samples.

\section{Procedurally Generated Environments} \label{subsec:environment}

We train and evaluate our results in multiple PCG tasks of both MiniGrid~\cite{minigrid} and Procgen~\cite{cobbe_leveraging_2020} benchmarks.
These benchmarks are widely adopted and used in the RL community for evaluating generalization due to their diverse task configurations \cite{kirk_survey_2022}.
This section details the objectives, how the levels are constituted (i.e., state space $\mathcal{S}$), the available action space ($\mathcal{A}$) and the reward function ($\mathcal{R}$) of the considered tasks in MiniGrid (Section \ref{subsec:minigrid}) and Procgen (Section \ref{subsec:procgen}). Moreover, we explain the relation between the selected train level distribution, and the pursued generalization skills (Section \ref{subsec:generalization}).

\subsection{MiniGrid} \label{subsec:minigrid}
\begin{figure}[t]
    \centering
    \includegraphics[width=0.475\columnwidth]{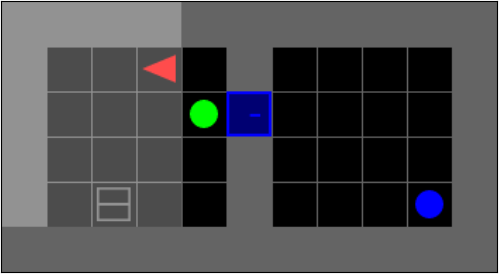}
    \includegraphics[width=0.475\columnwidth]{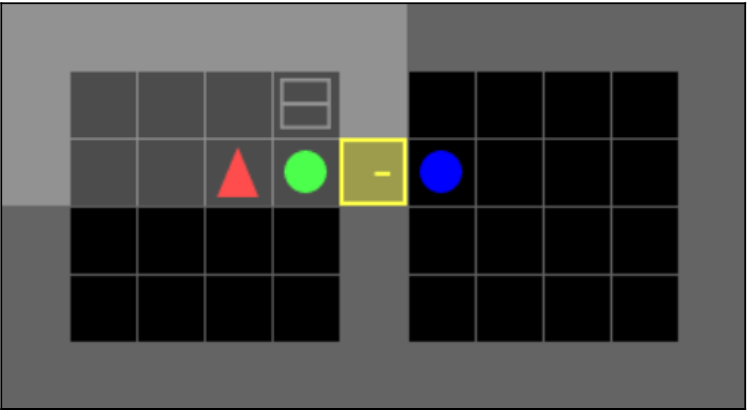}
    \includegraphics[width=0.475\columnwidth]{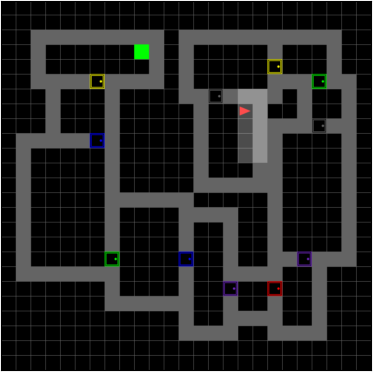}
    \includegraphics[width=0.475\columnwidth]{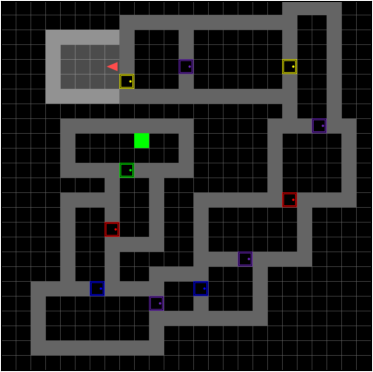}
    \caption{Two different levels of \texttt{O1Dlhb} (top) and \texttt{MN12S10} (bottom) tasks from the MiniGrid benchmark. The agent has only access to the bright area highlighted at its front. Variations in the agent's spawn position, door colors, and target locations across levels give rise to the procedurally generated content.
    }
    \label{fig:o1dlhb_mn12s10}
\end{figure}
In MiniGrid scenarios, each level presents unique variations, such as the agent’s spawn location, orientation, object positions and colors, and overall maze layout, creating diverse challenges for the agent. To accomplish a specified mission in each level, the agent navigates the grid-world using a set of 7 discrete actions ($\mathcal{A}$). While the complete state ($\mathcal{S}$) encompasses the entire grid layout, including all objects and properties, agents typically only receive local observations of a $7 \times 7$ grid $(\Omega)$ centered on their position. These observations include information about nearby object identifiers, colors, and properties.


In this paper we focus on \texttt{ObstructedMaze} (e.g., \texttt{O1Dlhb}) and \texttt{MultiRoom} (e.g., \texttt{MN12S10}) tasks, depicted in Figure ~\ref{fig:o1dlhb_mn12s10}, which require the agent to acquire diverse skills. For instance, in \texttt{O1Dlhb}, the agent (represented as a red triangle) must move the ball, uncover the key under the box, pick up the key, open the door, discard the key and pick the blue ball. In contrast, \texttt{MN12S10} requires the agent to advance through multiple rooms by opening intervening doors until it reaches the designated green square.

All considered tasks have sparse rewards because the agent only receives a non-zero reward if the task objective is completed in less than a predefined number of steps. The reward for task completion ($s_{t+1}$ is a terminal state) is given by:
\begin{equation}
    \mathcal{R}(s_t, a_t,s_{t+1})=1 - 0.9\cdot \frac{t}{t_{max}},
    \label{eq:reward_function}
\end{equation}
with $t$ being the current time step, and $t_{max}$ denoting the maximum number of steps per episode. The maximum number of steps varies with the specific task (e.g., \texttt{O1Dlhb}: 288, \texttt{MN12S10}: 240). Additionally, the terminal reward is only received if the task is completed before the maximum number of steps is reached ($t < t_{max}$) and and scales linearly with the number of time steps required by the agent to complete the task, namely, the faster the agent reaches the objective, the higher the terminal reward becomes. The reward at any other time step is zero.

\subsection{Procgen}\label{subsec:procgen}
\begin{figure}[t]
    \centering
    \begin{tabular}{cc}
         \texttt{Ninja} & \texttt{Climber} 
         \\
         \includegraphics[width=0.465\columnwidth]{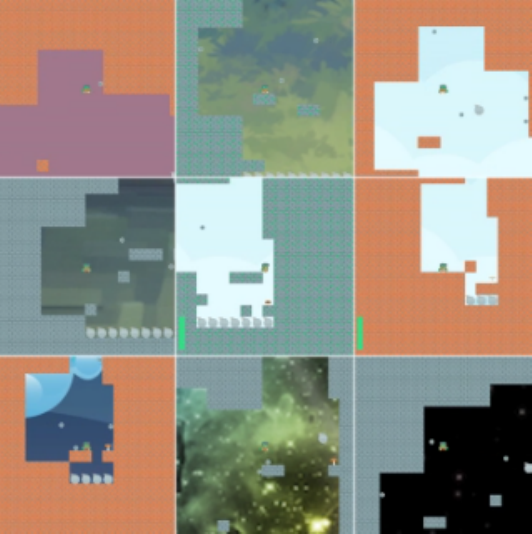} &  
        \includegraphics[width=0.45\columnwidth]{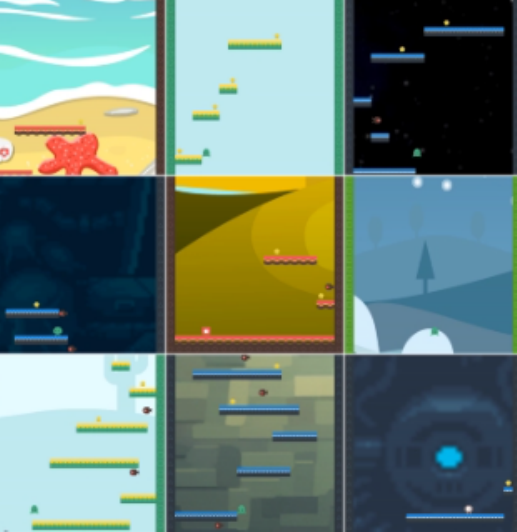}        
    \end{tabular}
    \caption{Different levels of \texttt{Ninja} (left) and \texttt{Climber} (right) tasks from the Procgen benchmark.
    Variations in game assets such as bombs, platform configurations, and background underscore the PCG elements in each task.
    }
    \label{fig:ninja_climber_envs_example}
\end{figure}

\begin{figure*}[ht]
    \centering
    \resizebox{2\columnwidth}{!}{\begin{tabular}{ccccc}

        & \multicolumn{1}{c}{Procgen} & & \multicolumn{2}{c}{MiniGrid}
        \\
        \cmidrule{2-2} \cmidrule{4-5}
        \multirow{2}{*}{\rotatebox[origin=c]{90}{Return}} 
        &
        \includegraphics[width=0.3\textwidth]{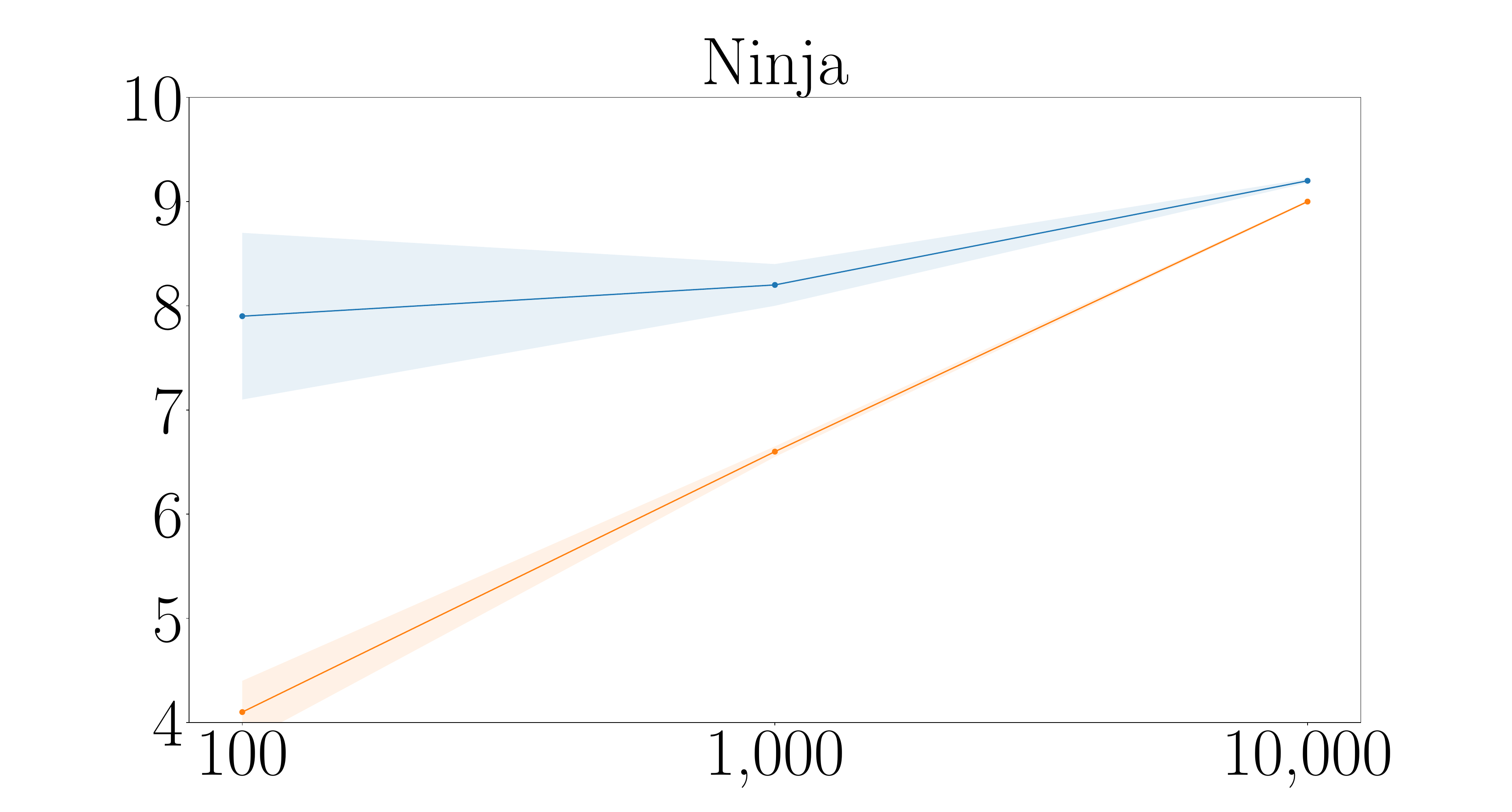} & &
        \includegraphics[width=0.3\textwidth]{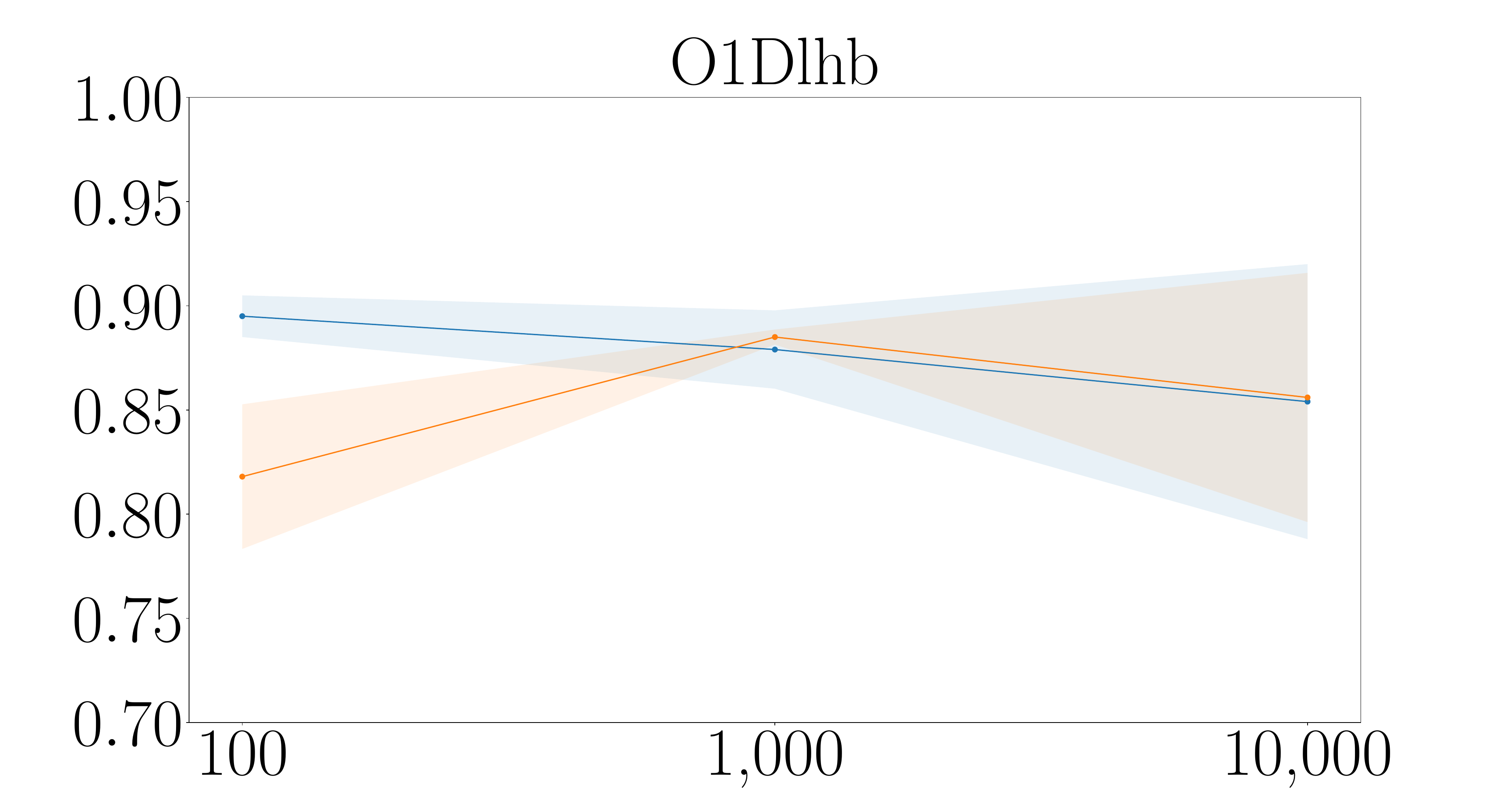} &
        \includegraphics[width=0.3\textwidth]{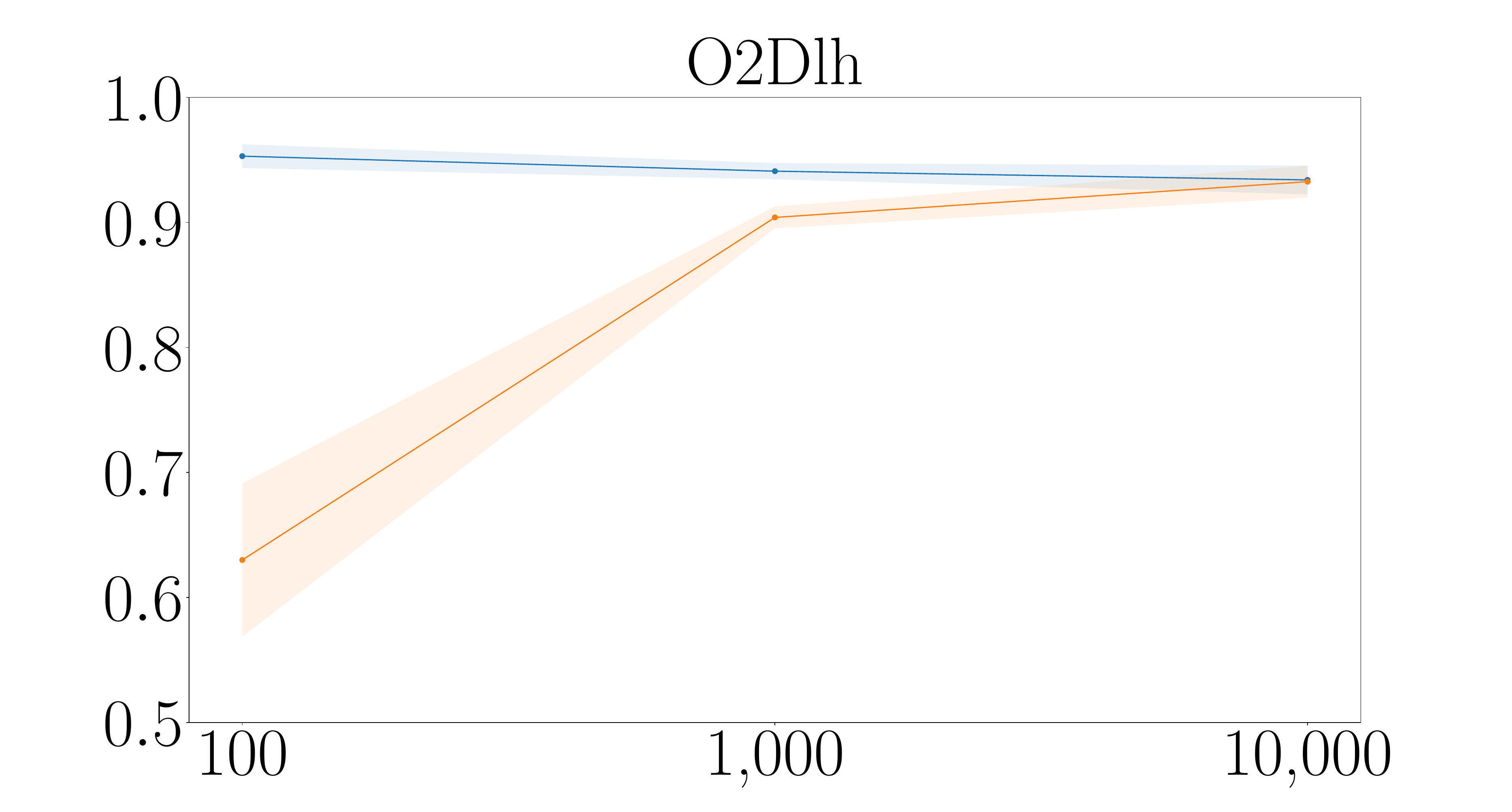} 
        \\
        & 
        \includegraphics[width=0.3\textwidth]{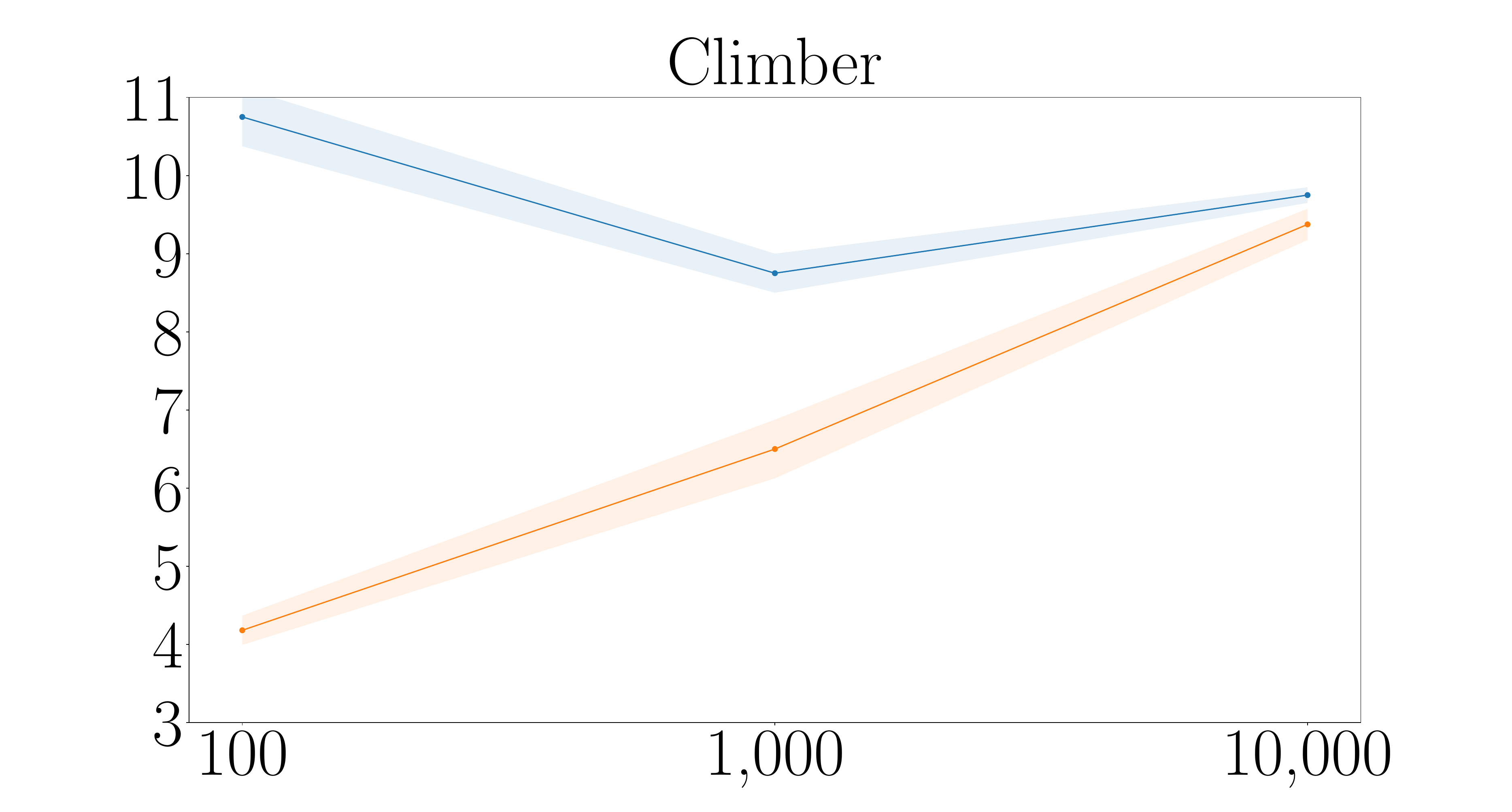} & &
        \includegraphics[width=0.3\textwidth]{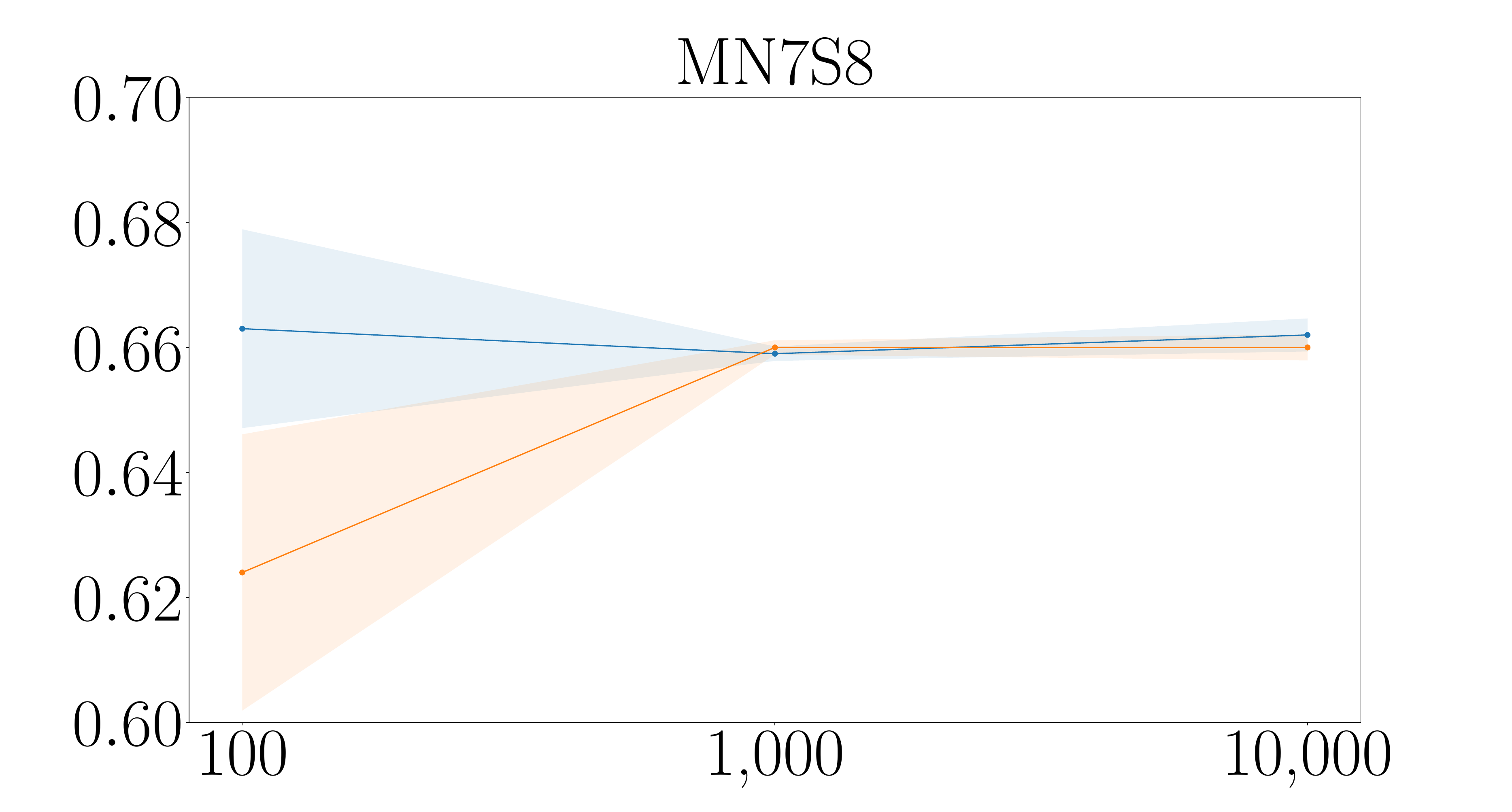} &
        \includegraphics[width=0.3\textwidth]{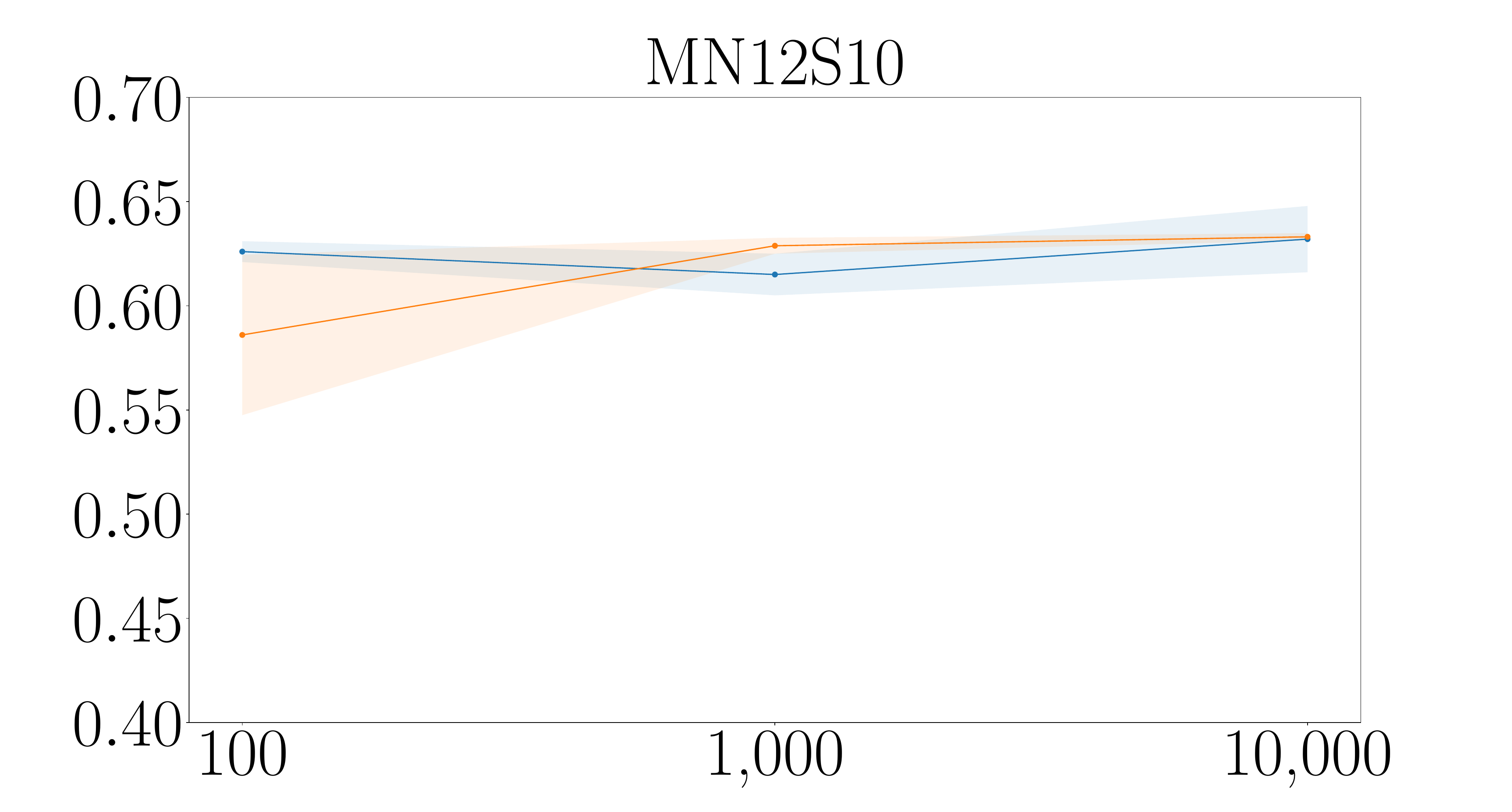} 
        \\
        & \multicolumn{4}{c}{Number of Levels}
    \end{tabular}}    
    \caption{
    Training performance (blue) and testing performance (orange) given by episodic returns depending on the number of levels used during training across multiple PCG tasks. The solid lines represent the average return, while the shaded areas indicate the standard deviation. 
    Testing performance is evaluated on levels that were not seen during the training phase. For Procgen tasks \texttt{Ninja} and \texttt{Climber} (data reproduced from \cite{cobbe_leveraging_2020}), shown on the left, the curves were obtained by training the agent with PPO for 200M time steps. In contrast, for Minigrid tasks \texttt{O1Dlhb}, \texttt{O2Dlh}, \texttt{MN7S8}, and \texttt{MN12S10}, the curves were derived from training the agent with an approach that combines IL with Intrinsic Motivation \cite{andres_towards_2022} for 10M to 20M time steps.
    }
    \label{fig:number_levels_performance}
\end{figure*}
In Procgen, each level's layout ($\mathcal{S}$) varies in terms of the game assets, backgrounds, and other elements that govern the number of entities, the horizon for task completion, and the overall level difficulty. All environments within Procgen employ a discrete 15-dimensional action space ($\mathcal{A}$). Similarly to MiniGrid, the agent in Procgen receives only a partial 64 $\times$ 64 RGB observation ($\Omega$) at each time step, limiting its perception of the environment. As a result, the agent must infer critical information about the level’s structure, such as directions to rewarding elements or obstacles that may lie outside its direct field of view.

We focus our evaluation on two specific Procgen tasks for their inherent exploration challenges: \texttt{Ninja} and \texttt{Climber}, shown in Figure \ref{fig:ninja_climber_envs_example}. In \texttt{Ninja}, the agent has to jump across narrow ledges to collect a mushroom located at the end of the level, while avoiding bomb obstacles. The agent can neutralize bombs by selecting an action to throw stars. The agent earns a reward of +10 upon mushroom collection (regardless of the steps taken) or +0 otherwise ($\mathcal{R}_{ninja}$). In the \texttt{Climber} task, the agent's objective is to ascend a series of platforms, gather stars, and evade scattered flying monsters. A default reward of +10 is granted upon the level's completion. Additionally, each level contains a variable number of stars, and the agent receives an incremental reward of +1 for each star collected. Consequently, the total reward ($\mathcal{R}_{climber}$) that the agent can achieve ranges between 10 and 17, depending on the number of stars available.

\subsection{Generalization Requirements: Train-Test Level Distributions} \label{subsec:generalization}

As previously introduced in Section \ref{subsec:pcg}, the goal when training an agent in PCG environments is not to memorize how to solve specific trajectories, but to learn relevant skills so that the agent can generalize to similar levels not seen during the training phase. 

One challenge to this sought generalization is that agents tend to overfit to specific levels encountered during training when trained on smaller sets of levels. This results in a notable gap in performance between training and testing levels, also referred to as the \textit{generalization gap}~\cite{cobbe_leveraging_2020}.

One strategy to reduce the generalization gap is to train on a large number of levels. By training on many levels, the agent encounters more diversity of levels during training. This increases the chance of learning robust policies that generalize effectively to testing levels,  since the agent may have encountered similar levels before. 
Figure \ref{fig:number_levels_performance} visualizes this correlation between the number of training levels and the agent's performance across held-out testing levels. For instance, in MiniGrid tasks, training on 1,000 levels (namely, $|\mathcal{L}_{train}(T)|=1,000$) is sufficient to achieve similar performance in previously unseen testing levels compared to training levels. On the other hand, in the Procgen tasks which exhibit more variability than those corresponding to MiniGrid, minimizing the generalization gap requires training over 10,000 levels.

Unfortunately, training on a larger number of levels inherently increases the complexity of the learning process, usually requiring more agent-environment interactions. Consequently, it becomes appealing to explore methods that enhance sample efficiency in these scenarios. In this context, we will study how IL can be effectively combined with RL training to reduce this computational requirement.

\section{Methodology} \label{sec:methodology}
In this section we outline our approach including the data collection 
 (Section \ref{sec:data_collection}) and IL techniques applied for pre- and concurrent training (Section \ref{sec:learning_offline_IL}).

\begin{table*}[t]
  \centering
  \captionsetup{width=\textwidth}
  \caption{Summary of the buffers collected for 4 MiniGrid tasks: \texttt{O1Dlhb}, \texttt{O2Dlh}, \texttt{MN7S8}, \texttt{MN12S10}. We provide statistics of the quantity and diversity of the data as given by the total number of stored levels ($\#_{levels}$) and the mean number of trajectories per level ($\mu_{\tau/level}$). The quality of stored trajectories is given by their mean number of experiences ($\mu_{\{s,a\}}$) and returns ($\mu_{G(\tau)}$). The last rows correspond to the expected optimal returns ($\mathbb{E}^*[G(\tau)]$) and optimal number of steps ($\mathbb{E}^*[lenght(\tau)]$) required to solve these tasks, where the expectation is over the entire level distribution of the task at hand. Each of those buffers contain 10,000 experience tuples.}
  \label{tab:buffer_metada}
      \begin{tabular}{lccccccccccccccc}
        \toprule
         & \multicolumn{3}{c}{O1Dlhb} & & \multicolumn{3}{c}{O2Dlh} & & \multicolumn{3}{c}{MN7S8} & & \multicolumn{3}{c}{MN12S10} \\
         \cmidrule{2-4} \cmidrule{6-8} \cmidrule{10-12} \cmidrule{14-16}
        & 10\% & 60\% & 90\% & & 10\% & 60\% & 90\% & & 10\% & 60\% & 90\% & & 10\% & 60\% & 90\% \\ \midrule
        $\#_{levels}$ & 88 & 259 & 250 & & 68 & 296 & 292 & & 115 & 193 & 210 & & 70 & 100 & 101\\
        $\mu_{\tau/level}$ & 1.01 & 1.03 & 2.18 & & 1 & 1.07 & 2.39 & & 1 & 1.02 & 1.13 & & 1 & 1.04 & 1.26 \\
        $\mu_{\{s,a\}}$ & 112.4 & 37.3 & 18.34 & & 147.1 & 31.5 & 14.3 & & 86.9 & 50.8 & 41.8 & & 142.8 & 96.2 & 78.1 \\
        $\mu_{G(\tau)}$ & 0.63 & 0.88 & 0.94 & & 0.74 & 0.95 & 0.98 & & 0.42 & 0.67 & 0.73 & & 0.45 & 0.64 & 0.71 \\ 
        \midrule
        $\mathbb{E}^*[G(\tau)]$ & \multicolumn{3}{c}{0.92} & & \multicolumn{3}{c}{0.95} & & \multicolumn{3}{c}{0.67} & &\multicolumn{3}{c}{0.65} \\
        $\mathbb{E}^*[length(\tau)]$ & \multicolumn{3}{c}{25.6} & & \multicolumn{3}{c}{32} & & \multicolumn{3}{c}{51.3} & &\multicolumn{3}{c}{93.3} \\
        \bottomrule
      \end{tabular}
\end{table*}

\subsection{Data Collection}\label{sec:data_collection}

Unlike in other IL works where expert demonstrations are given to the agent, we initially train an agent until convergence\footnote{The purpose of this agent is solely to act as a demonstrator for collecting trajectories, and is not utilized in any other way for our study.} using RAPID~\cite{zha_rank_2021}, as outlined in Section \ref{sec:sil}. 
For each specific task, we store a replay buffer containing a maximum of 10,000 experience tuples (i.e., the state-action pair $\{s, a\}$) as datasets of trajectories at three different checkpoints during training.
These checkpoints are selected either when a certain amount of interactions have been completed, or when the agent achieves a certain level of expertise (i.e., when an average return threshold is met). 

\subsubsection{MiniGrid}

For MiniGrid, the three checkpoints correspond to the agent first achieving evaluation returns of 0.1, 0.6 and 0.9 in \texttt{ObstructedMaze} scenarios, and 0.06, 0.4 and 0.6 in \texttt{MultiRoom}, respectively. These returns represent approximately 10\%, 60\% and 90\% of the expected optimal returns for those tasks. For the data collection, the RAPID agent is trained for 10M steps in the \texttt{MN7S8}, \texttt{MN12S10} and \texttt{O1Dlhb} tasks, which was sufficient to converge to the optimal policy. In \texttt{O2Dlh}, we find that RAPID is unable to reach the optimal policy. Therefore, we additionally incentivize exploration using intrinsic motivation~\cite{andres_towards_2022} and train for 30M steps, after which the agent achieves close to optimal performance. 
Table~\ref{tab:buffer_metada} outlines more details about the quality, quantity and diversity of each collected dataset in Minigrid. We can see that the more steps are required for a single trajectory, the fewer total trajectories can be stored in the dataset. This trend holds since the total number of experience tuples stored in each dataset is limited to 10,000 experiences. Therefore, in environments where a decrease in the number of steps per trajectory leads to higher returns, such as Minigrid, offline datasets comprising higher-quality trajectories (with fewer steps per trajectory) contain a larger number of total trajectories compared to datasets with trajectories of lower quality.

\subsubsection{Procgen}

On the other hand, the Procgen state space includes RGB observations. However, RAPID is not designed to handle non-discrete state spaces due to the local and global exploration scores depending on discrete states. Consequently, RAPID is limited in Procgen to prioritize experiences only based on the extrinsic score $S_{ext}$ (i.e., $w_1=0$ and $w_2=0$ in Equation \eqref{eq:rapid_scores}). Under these limitations, RAPID exhibits little to no improvement compared to PPO, and is unable to converge to the expected optimal policy. Due to these challenges, we collected datasets in Procgen using RAPID, but instead of storing datasets at checkpoints corresponding to the agent achieving certain performance thresholds, we store datasets at predetermined time steps during training. Specifically, we store the lower-, medium-, and higher-quality datasets at 7.5M, 15M, and 25M time steps of training. In both \texttt{Ninja} and \texttt{Climber} tasks, we train RAPID for a total amount of 25M steps. Table~\ref{tab:buffer_metada_procgen} outlines more details about the quality, quantity, and diversity of collected datasets in both Procgen tasks.
\begin{table}[t]
  \centering
  \captionsetup{width=\columnwidth}
  \caption{Summary of the buffers collected for 2 Procgen tasks: \texttt{Ninja} and \texttt{Climber}. We provide statistics of the quantity and diversity of the data as given by the total number of stored levels ($\#_{levels}$) and the mean number of trajectories per level ($\mu_{\tau/level}$). The quality of stored trajectories is given by their mean number of experiences ($\mu_{\{s,a\}}$) and returns ($\mu_{G(\tau)}$). The last row correspond to the expected optimal returns ($\mathbb{E}^*[G(\tau)]$) required to solve these tasks, where the expectation is over the entire level distribution of the task at hand. Each of those buffers contain 10,000 experience tuples.}
  \label{tab:buffer_metada_procgen}
  \resizebox{\columnwidth}{!}{
      \begin{tabular}{lccccccc}
        \toprule
         & \multicolumn{3}{c}{Ninja} & & \multicolumn{3}{c}{Climber} \\
        \cmidrule{2-4} \cmidrule{6-8}
        & 7.5M & 15M & 25M & & 7.5M & 15M & 25M \\ \midrule
        $\#_{levels}$ & 62 & 67 & 74 & & 3 & 2 & 4 \\
        $\mu_{\tau/level}$ & 2.06 & 1.77 & 1.69 & & 5.33 & 11.5 & 4.75\\
        $\mu_{\{s,a\}}$& 78.13 & 84.03 & 80 & & 625 & 434.78 & 526.31 \\
        $\mu_{G(\tau)}$ & 10 & 10 & 10 & & 15.09 & 16.0 & 16.75 \\ 
        \midrule
        $\mathbb{E}^*[G(\tau)]$ & \multicolumn{3}{c}{10} & & \multicolumn{3}{c}{12.6} \\
        \bottomrule
      \end{tabular}
  }
\end{table}

\begin{figure*}[t]
    \centering
    \includegraphics[width=\textwidth]{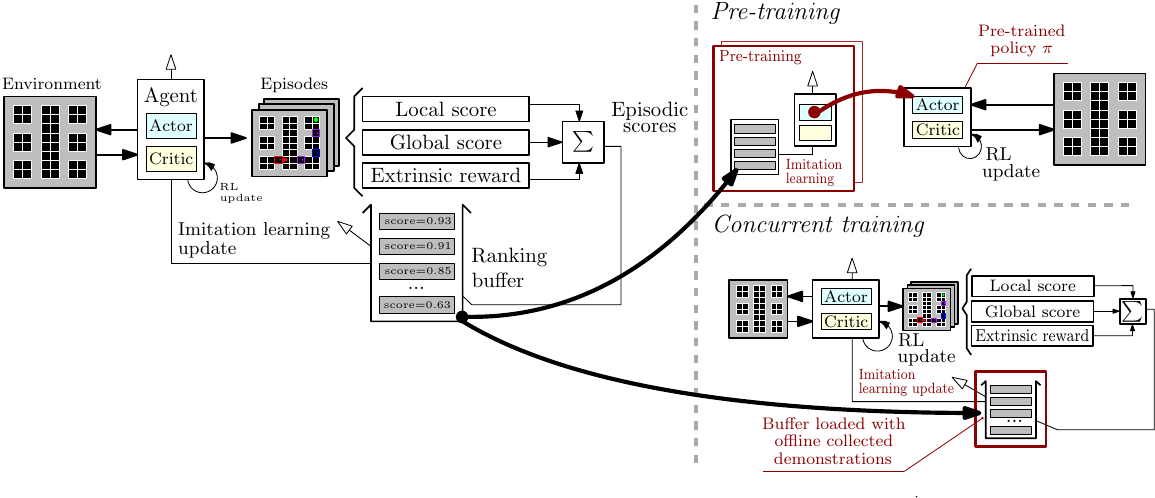}
    \caption{Our proposed evaluation framework. On the left, we train an agent with RAPID to collect datasets of varying quality. On the top right, we use IL just to pre-train a policy which is then used as initialization for the RL training. Alternatively, on the bottom right, we concurrently train the policy with RL and IL by initializing the buffer with the offline collected demonstrations.}
    \label{fig:proposed_methodology}
\end{figure*}
\subsection{Learning from Offline Data} \label{sec:learning_offline_IL}

In this study we consider two approaches to leverage offline data to improve RL agents, visualised in Figure \ref{fig:proposed_methodology}: pre-training and concurrent IL.
\begin{itemize}[leftmargin=*]
    \item \textit{Pre-training}: Prior to interacting with the environment, the pre-training approach samples batches of state-action pairs uniformly at random from a selected dataset (independently of the trajectory that they belong to).
    Each batch is used to minimize the BC loss as described in Equation ~\ref{eq:bc}.
    We complete a fixed number of such optimization steps during this phase.
    After pre-training, no more IL is applied, and the agent is trained using standard RL while interacting with the environment.
    \vspace{2mm}

    \item \textit{Concurrent training}: For concurrent training, both the IL and RL losses are utilized during online training. The RL agent's policy is randomly initialized and trained from online interactions with the environment (as usual).
    An RL update involves processing a collected rollout through multiple optimization steps, which are determined by specific hyperparameters of the chosen RL algorithm (e.g., number of epochs, number of minibatches). Concurrently, following each RL update, an IL update is conducted to further enhance the learning process. This IL update involves sampling uniformly at random a specified number of batches (as explained above), with each batch undergoing a separate optimization step. The number of these batches is configurable, allowing us to fine-tune the balance between the impacts of IL and RL within our training strategy.
    
    Additionally, during the online training phase, if a collected trajectory has a higher prioritization score than other trajectories in the buffer according to Equation~\eqref{eq:rapid_scores}, it is added to the buffer, replacing trajectories with lower scores.
\end{itemize}

\section{Evaluation Results}\label{subsec:results}

In order to understand how IL impacts the described offline-to-online paradigm in PCG environments, we pose the following research questions:
\begin{enumerate}
    \item Does pre-training a RL agent with IL improve sample efficiency or converged performance?
    \item Can IL from offline trajectories be concurrently used to train an agent alongside online RL? 
    \item How many levels and trajectories are needed for effective pre-training? And for concurrent training? 
    \item How does the quality of demonstrations affect the low demonstration regime?
\end{enumerate}

Informed by prior work identifying better performance of on-policy over off-policy algorithms in PCG environments ~\cite{ehrenberg_study_2022,cobbe_leveraging_2020,mohanty_measuring_2021}, we use PPO \cite{schulman_proximal_2017} for the online RL training. We compare the results of agents with and without pre-training and concurrent IL with three baselines: i) only training agents online with RL (using PPO), ii) only training agents offline with IL (using BC)\footnote{Train on the demonstrations provided in each buffer using BC as in Equation \eqref{eq:bc}; no RL is applied.}, and iii) training agents with the RAPID self-IL approach~\cite{zha_rank_2021}. We refer to \ref{app:hyperparams} for further information regarding the selected hyperparameters and configuration.

In our experiments with MiniGrid tasks, we used 10,000 training levels to ensure robust generalization as detailed in Section \ref{subsec:generalization}. For Procgen, we trained the agent on 200 levels aligning with the standard established by previous research \cite{jiang_prioritized_2021,raileanu_decoupling_2021}, although it might be insufficient for optimal generalization.
Unless otherwise stated, all the provided plots report the mean and standard deviation of the average return throughout training, computed over the past 100 train levels/episodes, across 3 different independent runs. 

Next, we present the results for the two benchmarks considered in this study: MiniGrid (Section \ref{subsec:minigrid_results}) and Procgen (Section \ref{subsec:procgen_results}). Within each section, we address the impact of employing IL in both pre-training and concurrently with RL during the online phase. Moreover, we explore the significance of diversity in demonstration selection in MiniGrid, particularly when dealing with a limited number of demonstrations (Section \ref{subsubsec:results_div_minigrid}). We also investigate the implications of demonstration composition in Procgen tasks (Section \ref{subsubsec:procgen_buffer_diversity_1ep}), where the levels vary more in their agent observations, making generalization across levels more challenging.

\subsection{MiniGrid} \label{subsec:minigrid_results}
\subsubsection{Pre-training with Imitation Learning}\label{subsec:results_pretraining}

\begin{figure}[t]
    \centering
    \includegraphics[width=\columnwidth]{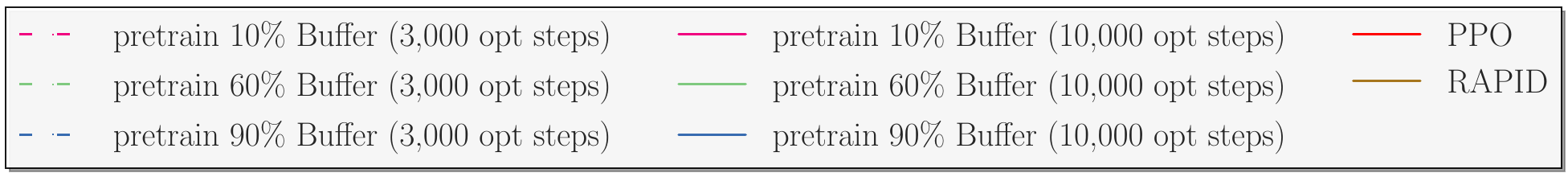}
    \\
    \includegraphics[width=\columnwidth]{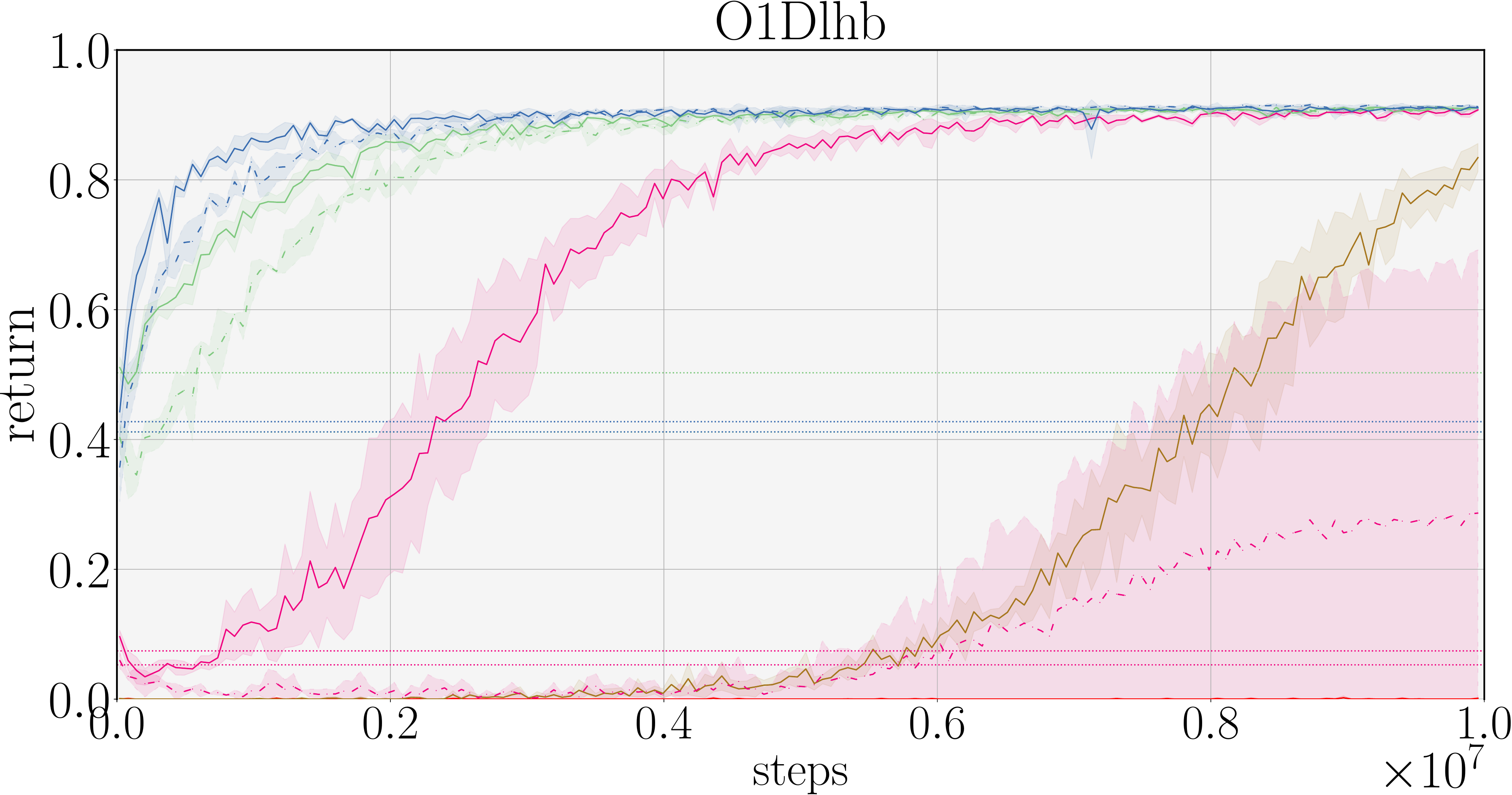}
    \includegraphics[width=\columnwidth]{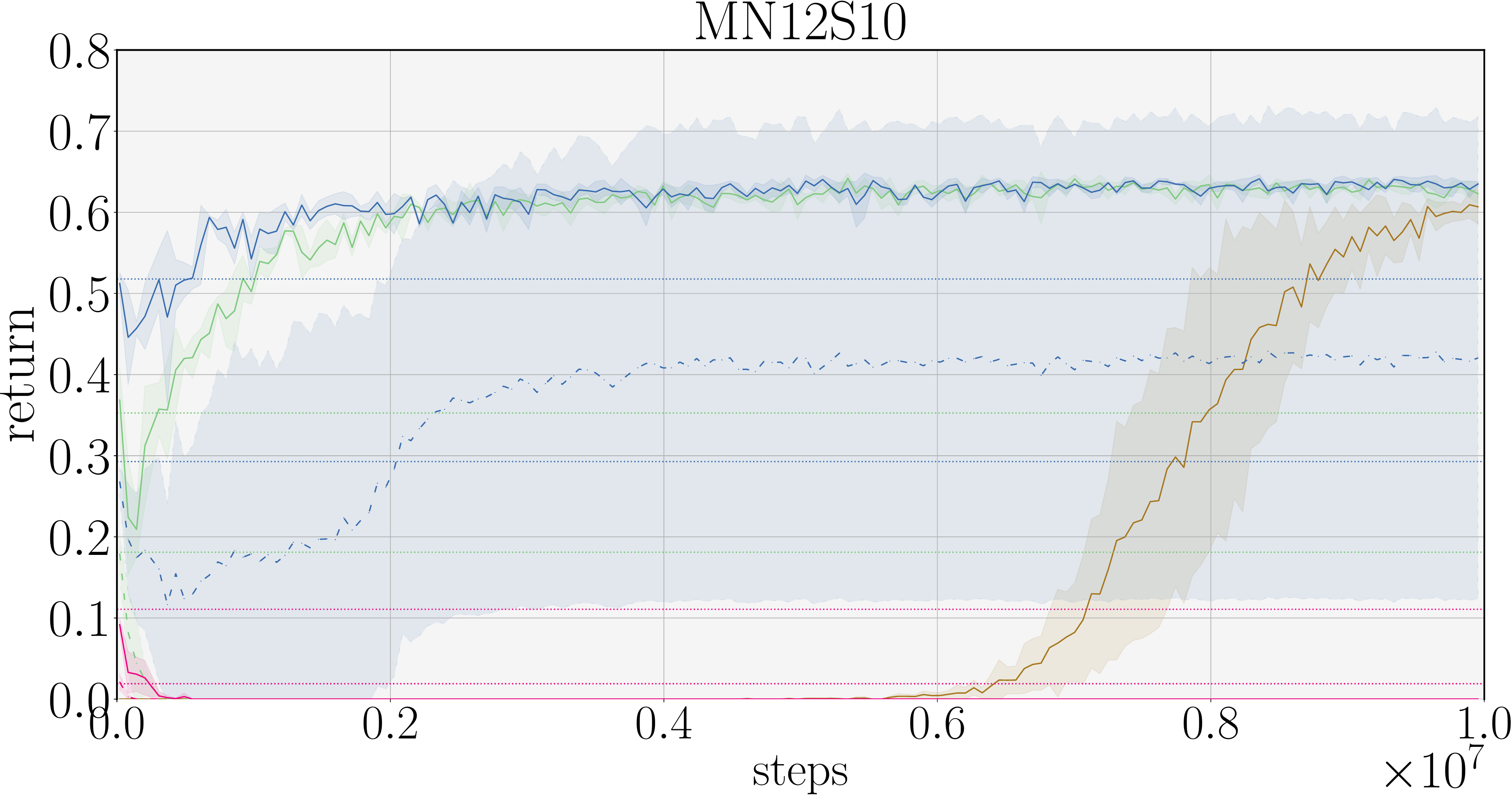} 
    \\
    \caption{Performance of the agent when pre-training with IL before the RL training phase in \texttt{O1Dlhb} (top) and \texttt{MN12S10} (bottom). The horizontal dashed lines represent the pre-trained policies' return over the entire distribution of levels --trained solely with BC--that serve as initialization point for the training phase. Depending the task and the demonstrations used, the employed number of pre-training 
    optimization steps
    (3,000 or 10,000) affects more/less the performance. Notice the x-axis provides the number of interactions/steps of the agent (after the pre-training phase).}
    \label{fig:pre_training}
\end{figure}

Figure \ref{fig:pre_training} presents the agent's performance in MiniGrid when leveraging IL for policy pre-training. 
Initializing the agent with a pre-trained policy makes it perform better from the beginning when compared to learning from scratch, resulting in improved sample efficiency and overall performance compared to both PPO (in red) and RAPID (in brown). 

The speed of convergence when using pre-training further benefits from higher quality demonstrations and increased number of pre-training optimization steps.
However, independently of the chosen demonstrations, policies derived through BC after pre-training (as shown by the horizontal lines in Figure~\ref{fig:pre_training}.), without further fine-tuning, are insufficient for generalization in the entire level distribution, and require additional training with RL to approach an optimal policy. This poor performance of the agent after BC pre-training can be explained by the composition of the offline datasets, which only contain experiences from a small subset of levels. Therefore, after training with IL on these experiences, the agent might be able to solve the levels represented in the offline dataset, but exhibit poor performance on many other levels, leading to overall low returns. This is further shown by Table~\ref{tab:buffer_metada} where we can see that offline datasets only contain experiences of between 70 and 300 levels, which have been shown to be insufficient to close the generalization gap (Figure~\ref{fig:number_levels_performance}).
\paragraph{Initial Policy Quality and Generalization} The quality of trajectories in the offline datasets plays a crucial role in the performance of agents: higher quality trajectories consistently result in better agent performance. For example, when using 3,000 
optimization steps,
agents trained with the 60\% and 10\% buffers in \texttt{MN12S10} yield $\approx 0.2$ and $\approx 0.1$ returns after pre-training, respectively. However, these returns quickly collapse to $0$ during the online training phase. In contrast, in \texttt{O1Dlhb} both 90\% and 60\% buffers initialize agents with higher return values ($\approx$0.4), and these agents successfully learn an optimal policy.

Moreover, the speed of convergence towards an optimal solution is directly proportional to the buffer quality: agents using the 90\% buffer converge faster than those with the 60\% buffer, which in turn outpace agents trained with a 10\% quality buffer. This can be clearly observed in \texttt{O1Dlhb}, where the agent requires approximately 0.6M, 1.6M and 4.5M steps to obtain an average return of 0.8 with the 90\%, 60\% and 10\% buffers, respectively.

\paragraph{Number of IL optimization steps}
We find that increasing the number of optimization steps in pre-training from 3,000 to 10,000 significantly improves the performance right after pre-training and during RL fine-tuning. This is, the number of optimization steps utilized for pre-training plays a crucial role in determining the agent's performance. 

For instance, when using 3,000 pre-training optimization steps, only the pre-training from the higher-quality 60\% and 90\% buffers (\texttt{O1Dlhb}) and with the 90\% buffer (\texttt{MN12S10}) was effective. Yet, if the count of such steps is increased
\footnote{Note that the agent uses the same offline buffers and identical number of online interactions; only the number of IL optimization steps at pre-training are increased.} to 10,000, agents are able to extract more valuable insights from offline data, which translates into a higher quality policy and an improved performance. As a consequence, the agent manages to successfully learn even from the suboptimal 60\% buffer in \texttt{MN12S10}, and from the 60\% and 10\% buffer in \texttt{O1Dlhb}. A tangible improvement can be seen in \texttt{MN12S10} with the 60\% buffer, where the agent's initial return performance surged to $\sim$0.35, a marked increase from the earlier $\sim$0.2. 

\paragraph{Distribution Shift Adaptation} The overall difficulty of a task, compounded by the allowed maximum number of steps and the expected length of an optimal trajectory, can significantly hinder the effectiveness of pre-training the agent's policy. This is particularly evinced in MiniGrid scenarios, and specially in the \texttt{MN12S10} task, where even if nearly completing a task by traversing almost the entire maze, an agent receives no rewards if it fails to accomplish the goal within the maximum allowed time steps ($=t_{max}$). As a result, the agent's networks are updated to classify near-success trajectories as failures. 

To illustrate this, let us consider the extent to which a given trajectory can deviate from the optimal policy's trajectory in terms of time steps. On the one hand, in \texttt{O1Dlhb} the agent is allowed to take up to $\times11.5$ longer (worse) than the optimal policy's trajectory to solve the task. On the other hand, in \texttt{MN12S10} this gap is reduced to $\times2.5$
\footnote{The expected episode length for the optimal policy in an average \texttt{O1Dlhb} level is $\mathbb{E}^*[length(\tau)]=25$. Considering that the agent has $t_{max}=288$ time steps to solve the task, then the agent is allowed to take $288/25 \approx 11.5 \times$ longer trajectories than the optimal. In \texttt{MN12S10}, the expected episode length is $t_{max}=240$ and the optimal policy requires $93$ time steps, which results in a ratio of $240/93 \approx 2.5 \times$ possible longer trajectories with respect to the optimal.}.
This suggests a tighter episode duration in \texttt{MN12S10} might hinder the agent's adaptability during online training.

\begin{figure}[t]
    \centering
    \includegraphics[width=\columnwidth]{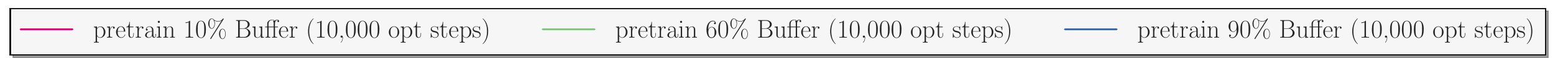}
    \includegraphics[width=\columnwidth]{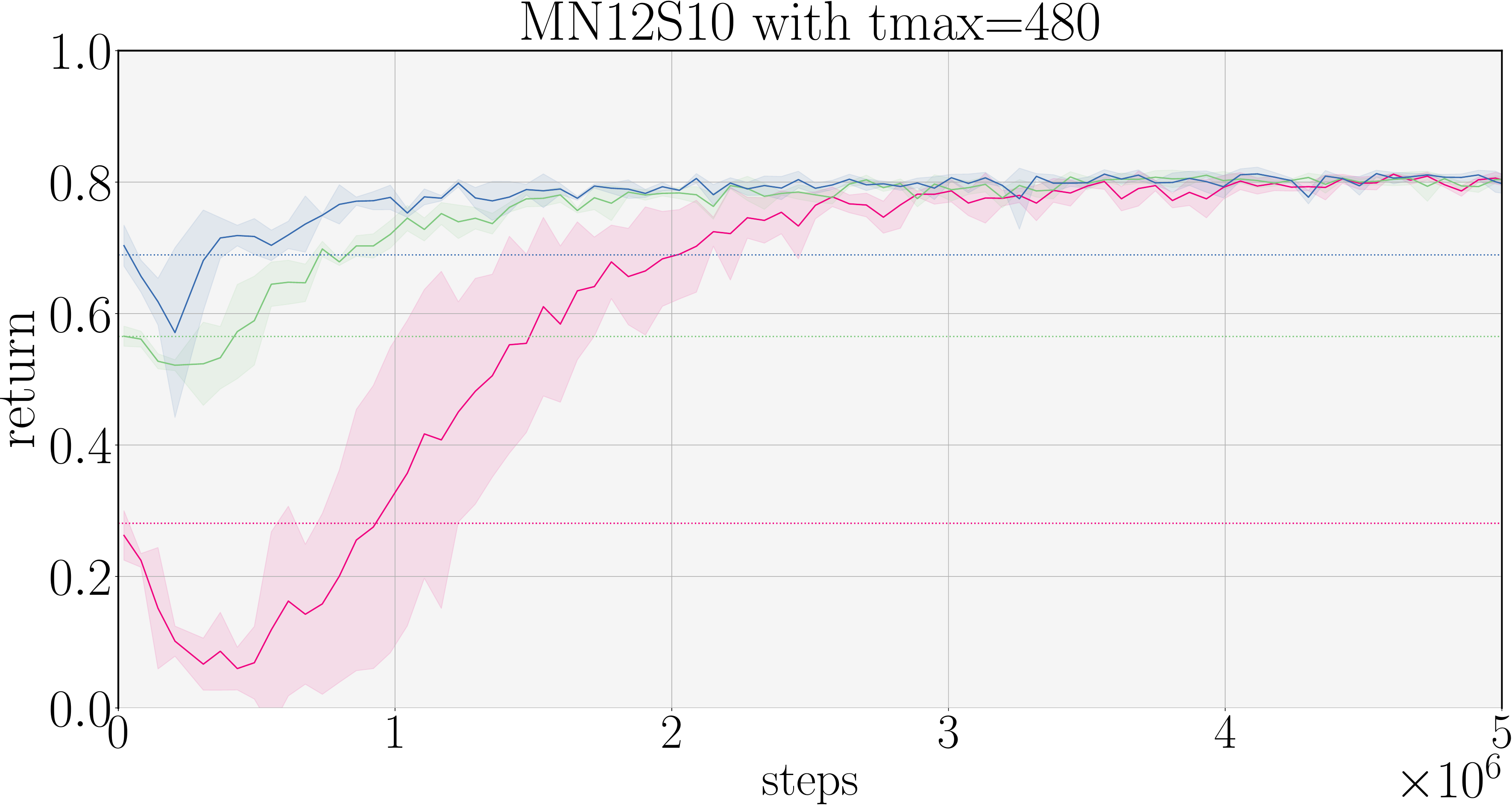}
    \caption{Performance of the agent in \texttt{MN12S10} when increasing $t_{max}$ from 240 to 480 and using Imitation Learning during pre-training phase.}
    \label{fig:mn12s10_extended_maxsteps}
\end{figure}

To investigate this hypothesis, we train agents in the \texttt{MN12S10} task with a maximum episode length of $t_{max}=480$ in contrast to the original $t_{max}=240$ (with which the agent previously failed to learn)\footnote{Note that $t_{max}$ modification changes the expected optimal return from 0.65 shown in Table ~\ref{tab:buffer_metada} to $\sim$ 0.82., as $\mathcal{R}$ depends on that value.}. Figure ~\ref{fig:mn12s10_extended_maxsteps} shows the learning curve in this modified task, indicating that this simplification allows the agent to explore and adapt to the new distribution of levels, leading to higher returns. Whereas the agent failed in the original task when pre-trained with the 10\% buffer (pink), the same approach is successful once the maximum number of time steps is increased. Similarly, the initial drop in episodic returns after pre-training observed with the 60\% buffer (green) is averted.

\subsubsection{Concurrent Online RL and IL}\label{subsec:results_concurrrent}

\begin{figure}[t]
    \centering
    \includegraphics[width=\columnwidth]{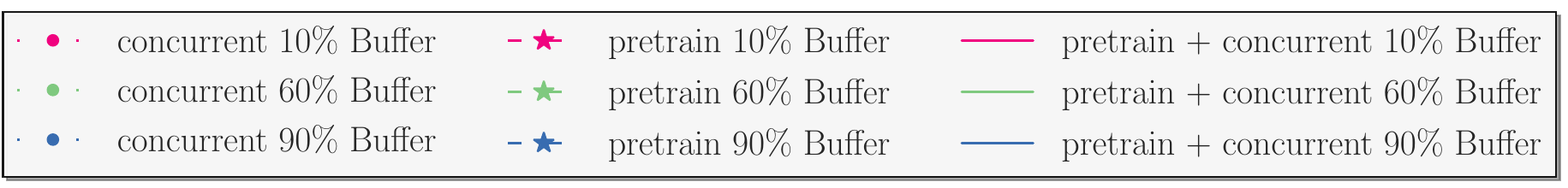}
    \\
    \includegraphics[width=\columnwidth]{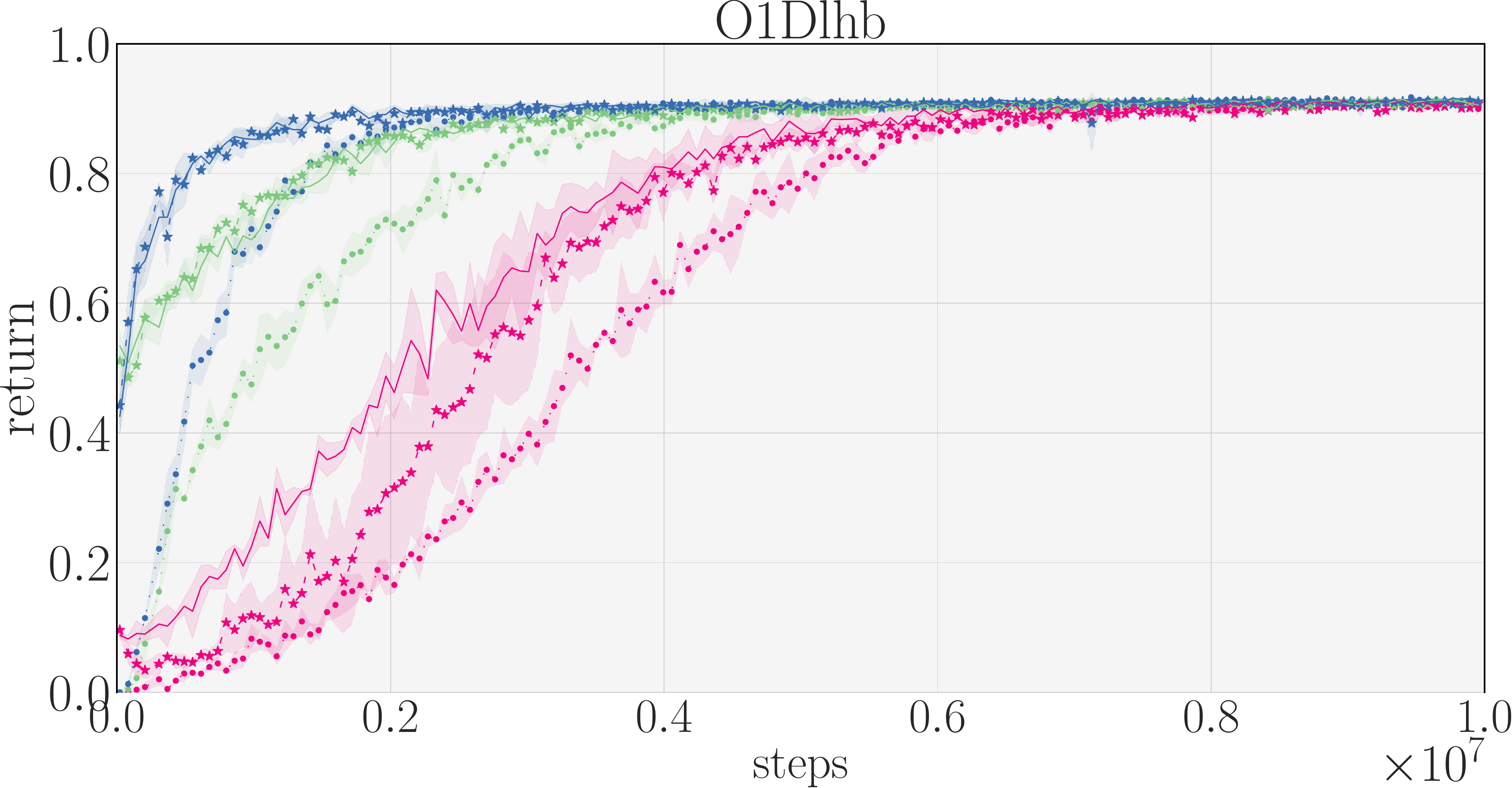}
    \includegraphics[width=\columnwidth]{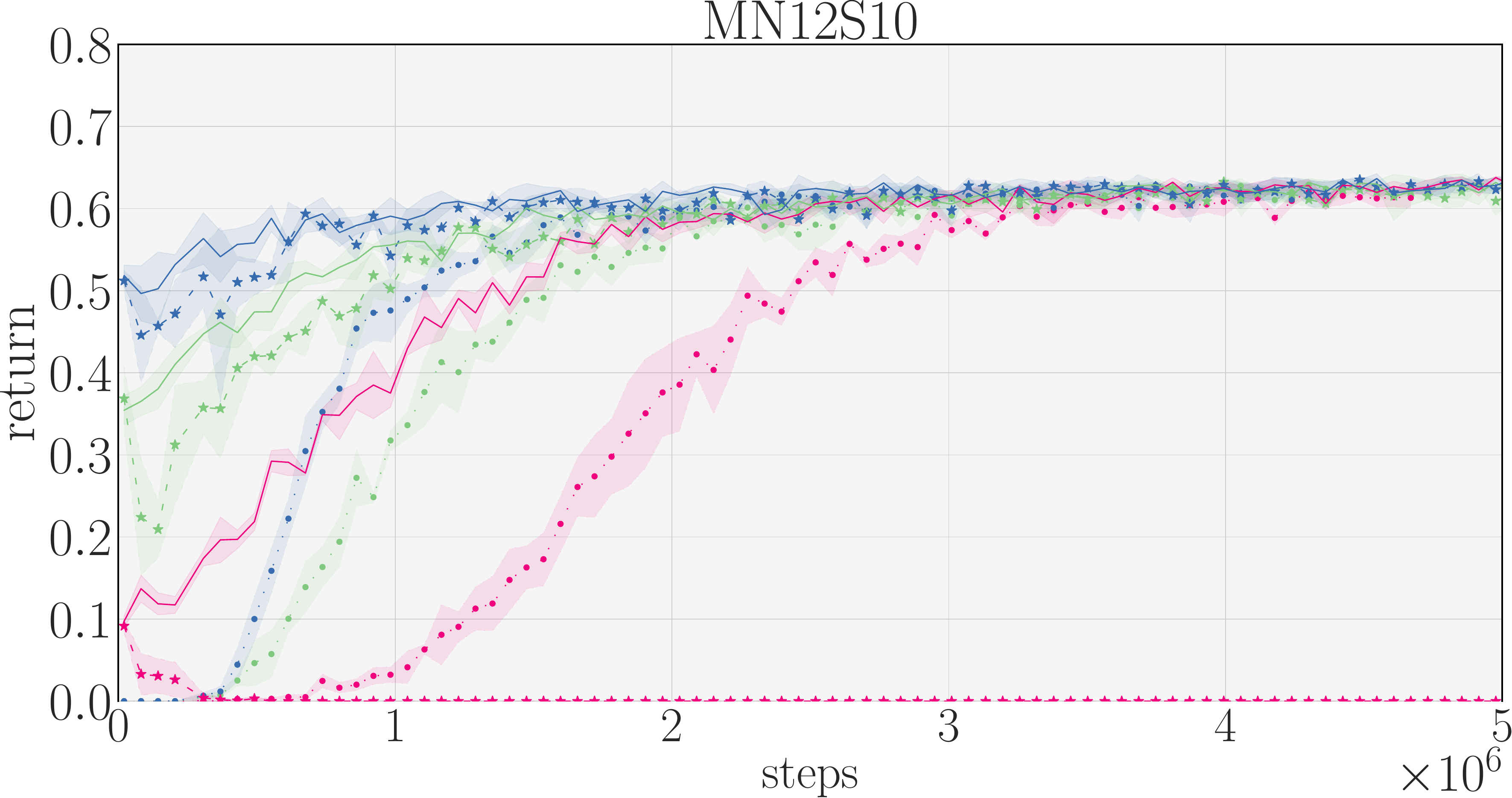}
    \\
    \caption{Performance of the agent when randomly initializing the policy and using either only RL (dotted curves) or both IL and RL losses (solid curves) online during the training in \texttt{O1Dlhb} (top) and \texttt{MN12S10} (bottom). The obtained results are compared when IL is just used for pre-training (dashed curves). With low-quality demonstrations (i.e., 10\% Buffer), the best results are retrieved when IL is used at both pre-training and at the main training phase.}
    \label{fig:concurrent_online}
\end{figure}

In Figure~\ref{fig:concurrent_online}, we compare the impact of concurrently training the agent with IL and RL during online training in MiniGrid. We find that agents trained with concurrent IL manage to solve all the task and with all the analyzed buffers even when pre-training exposed difficulties to learn (e.g., in \texttt{MN12S10} with the 10\% buffer, pink). Thus, concurrently using RL and IL exhibits better robustness, despite not having any prior knowledge at the beginning of the training phase. On the contrary, due to the significant jumpstart obtained by the pre-trained policies, pre-training can result in better sample-efficiency.

\paragraph{Pre-training + Concurrent RL and IL} The previous results are further improved by combining both pre-training (in order to benefit from the kickstart and sample efficiency) and concurrent IL during online training (for robustness). The solid lines in Figure ~\ref{fig:concurrent_online} show that using IL in pre-training and also concurrently with RL during the online phase results in robust convergence to the optimal policies, with less number of online interactions in both tasks.

\begin{figure*}[h!]
    \centering
    \resizebox{2\columnwidth}{!}{\begin{tabular}{c|cccc}
    \multicolumn{5}{c}{\includegraphics[width=\columnwidth]{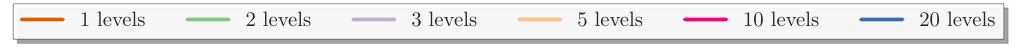}} 
    \\
    & \texttt{O1Dlhb} & \texttt{O2Dlh} & \texttt{MN7S8} & \texttt{MN12S10}  
    \vspace{-1mm}
    \\
    \rotatebox[origin=l]{90}{90\% Buffer} &
    \includegraphics[width=0.475\columnwidth]{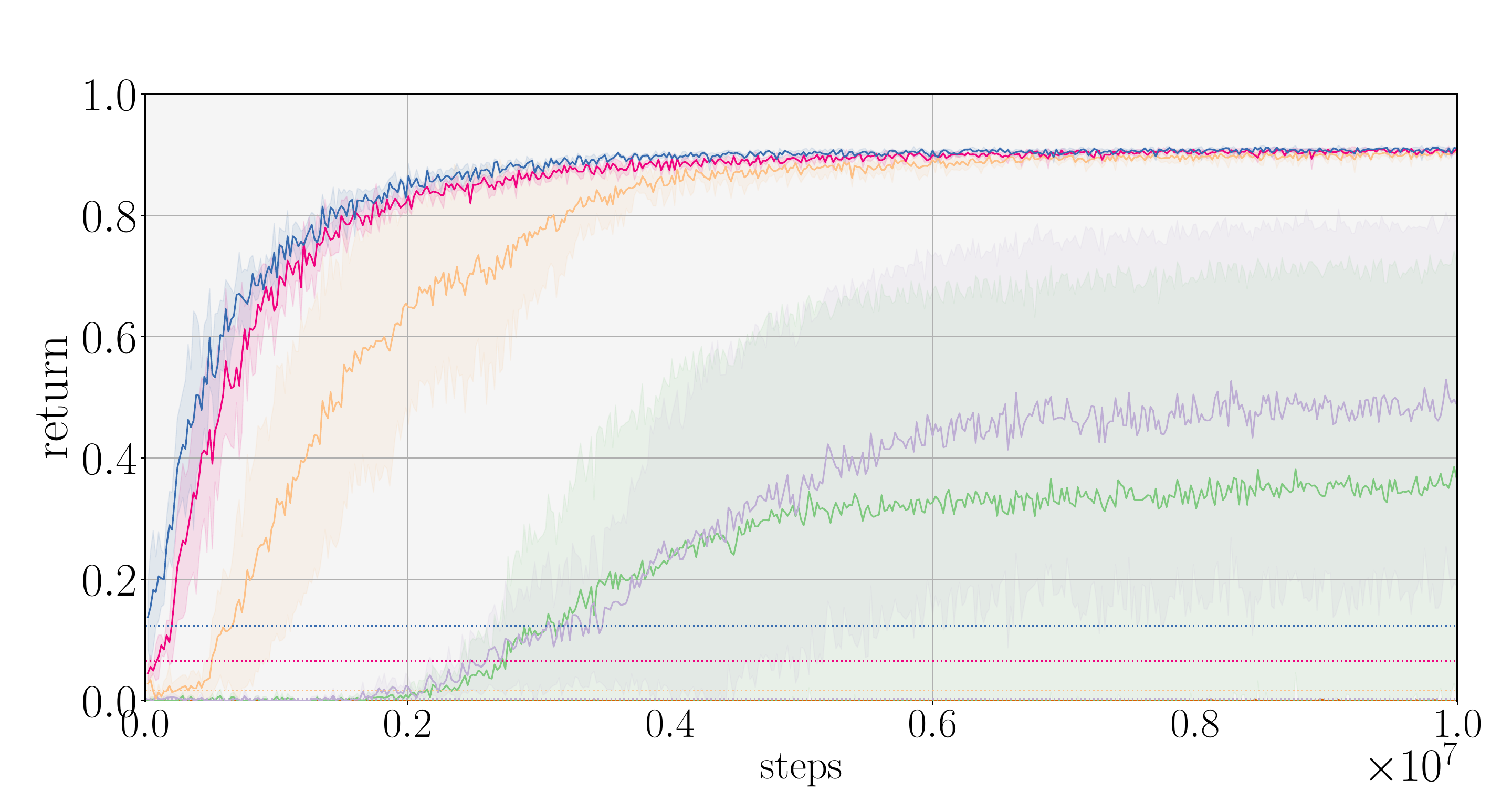} &
    \includegraphics[width=0.475\columnwidth]{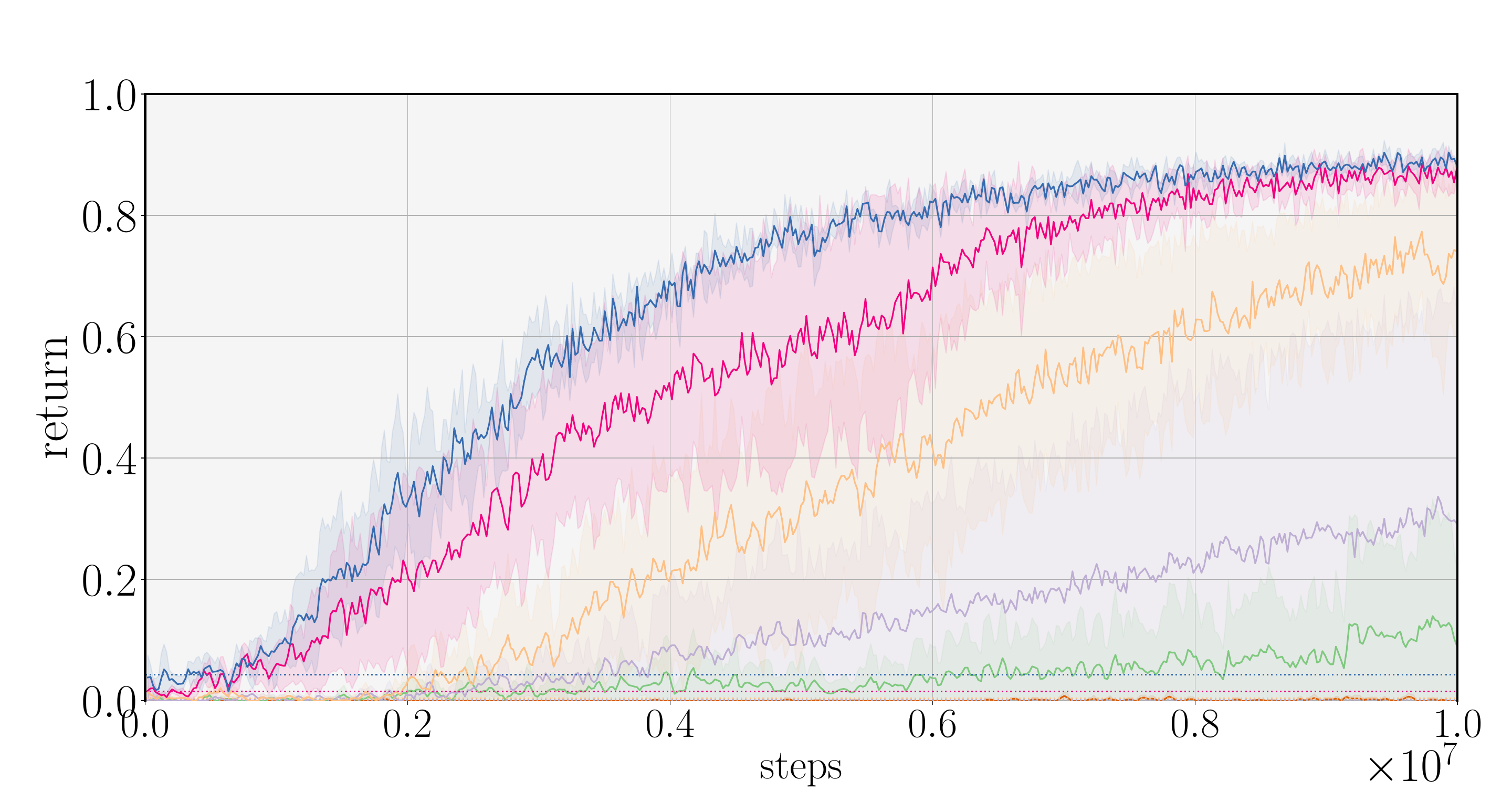} &
    \includegraphics[width=0.475\columnwidth]{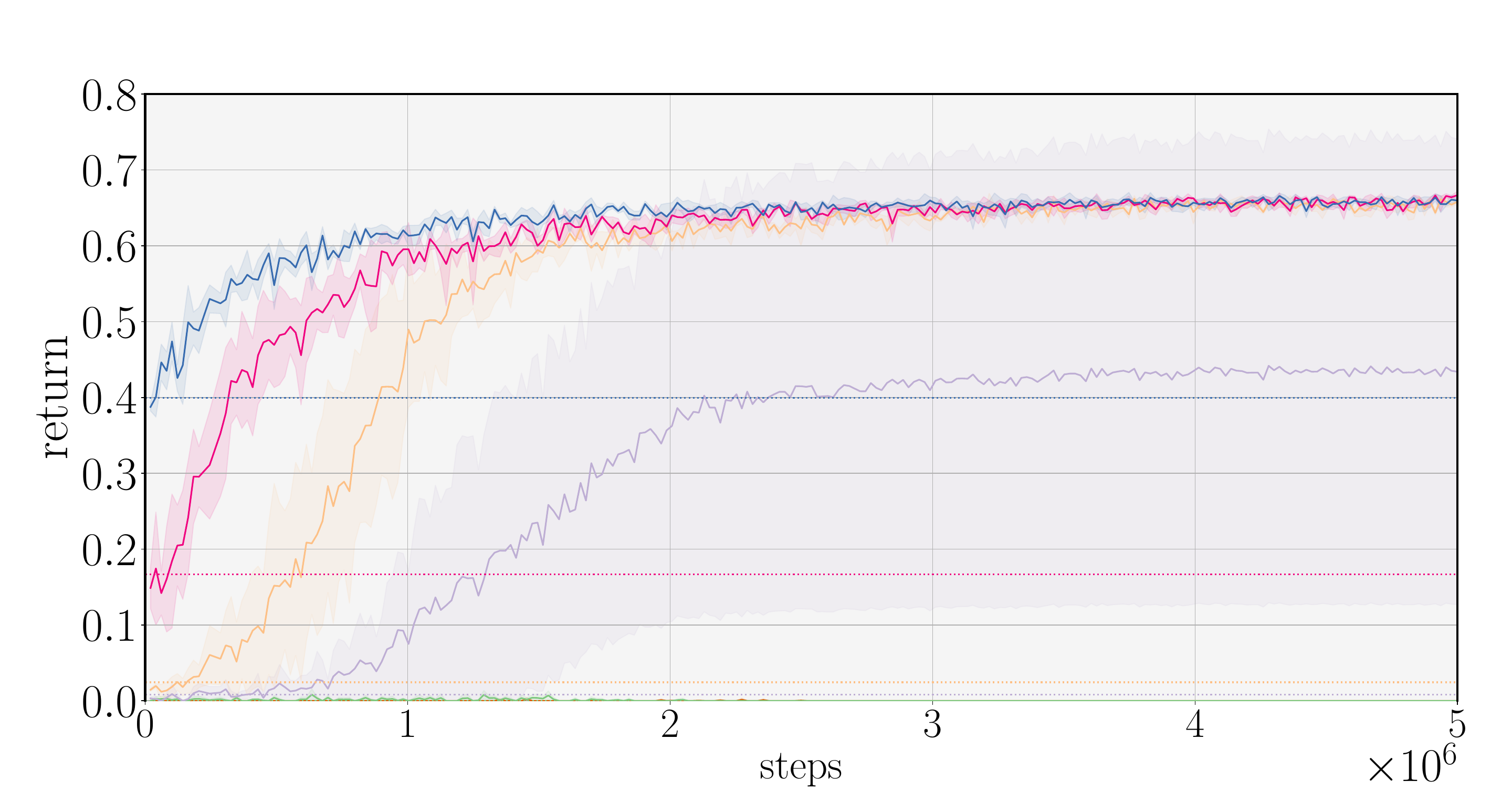} &
    \includegraphics[width=0.475\columnwidth]{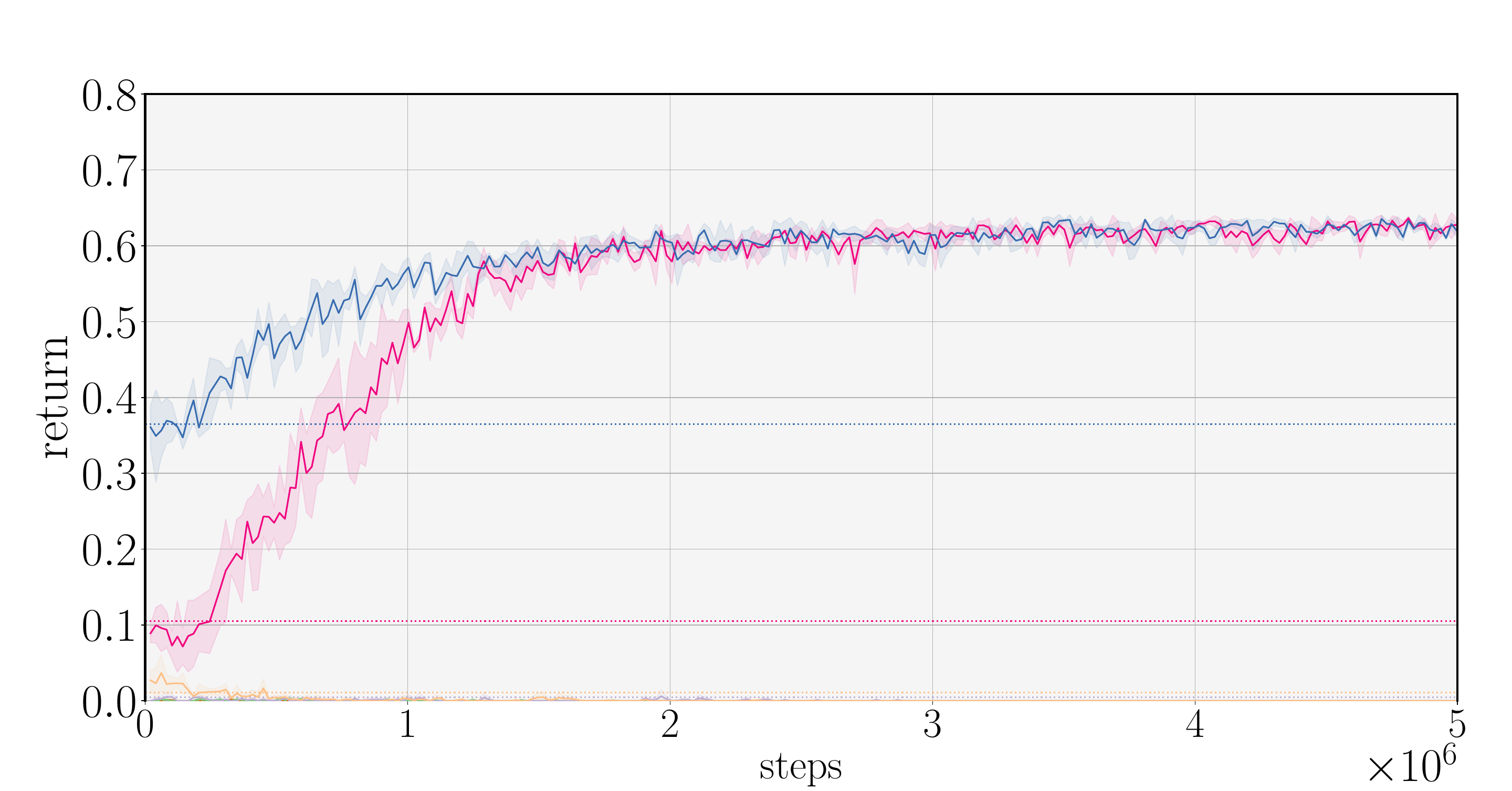}
    \\
    \rotatebox[origin=l]{90}{60\% Buffer}&
    \includegraphics[width=0.475\columnwidth]{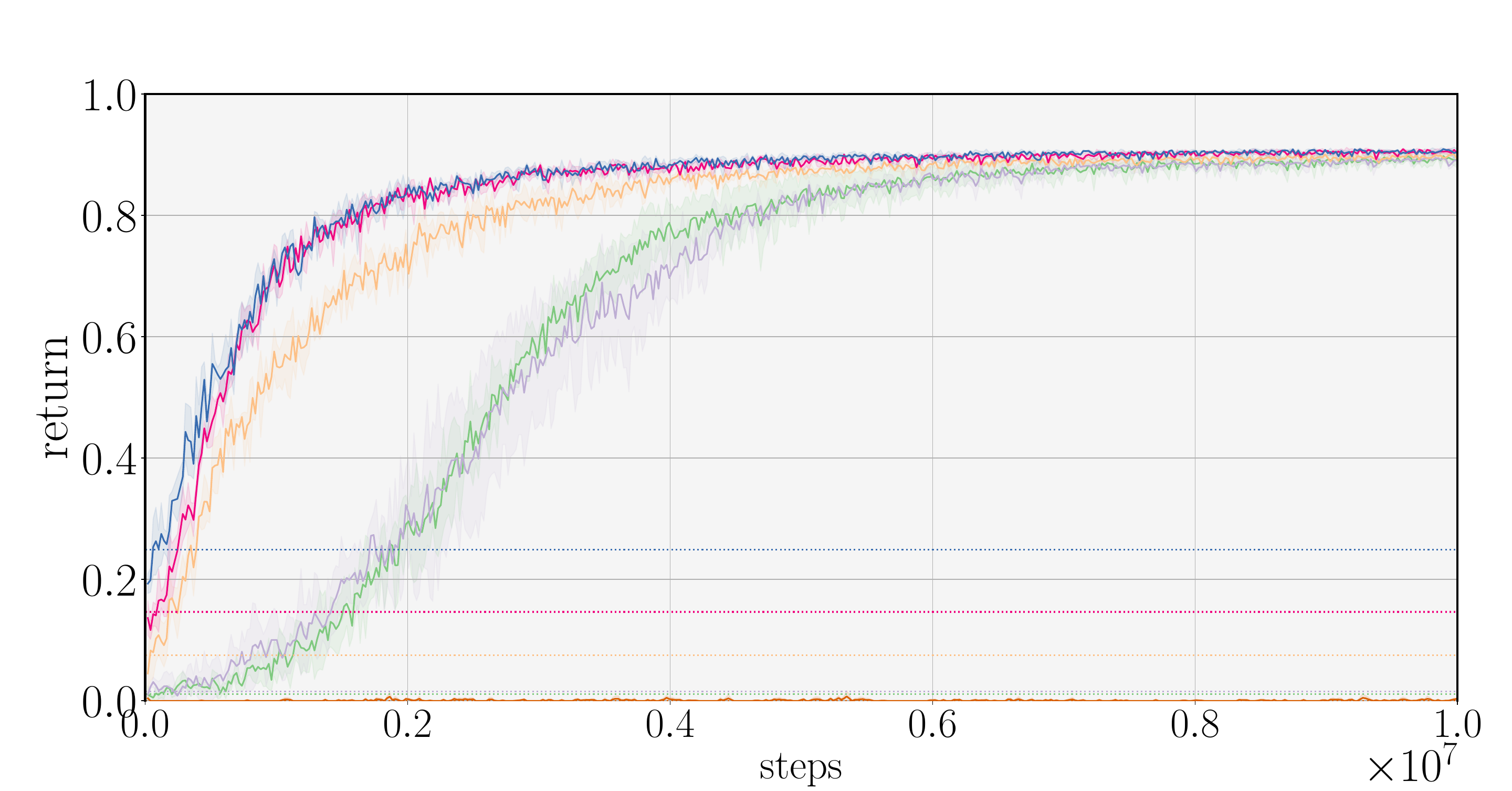} &
    \includegraphics[width=0.475\columnwidth]{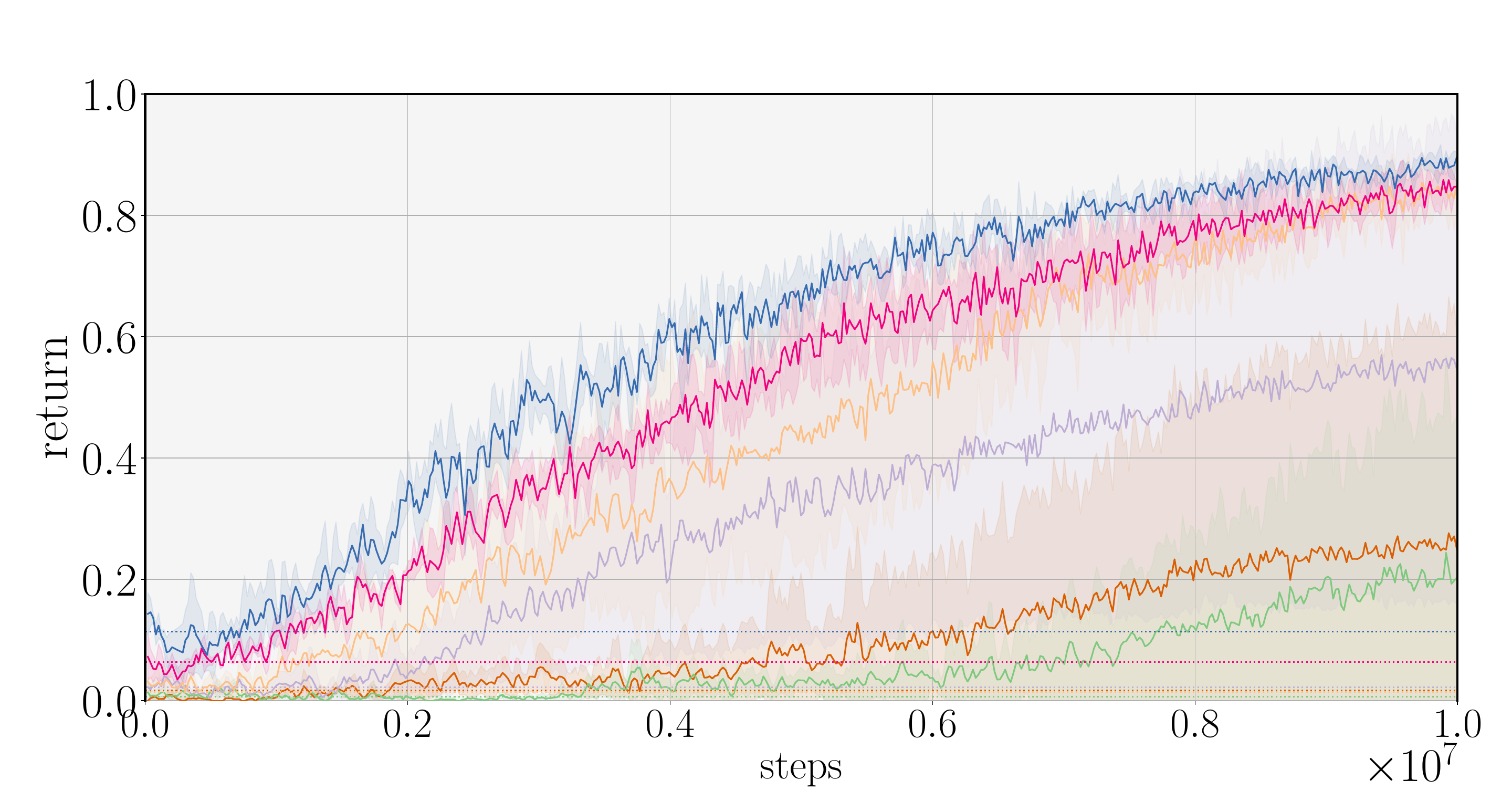} &
    \includegraphics[width=0.475\columnwidth]{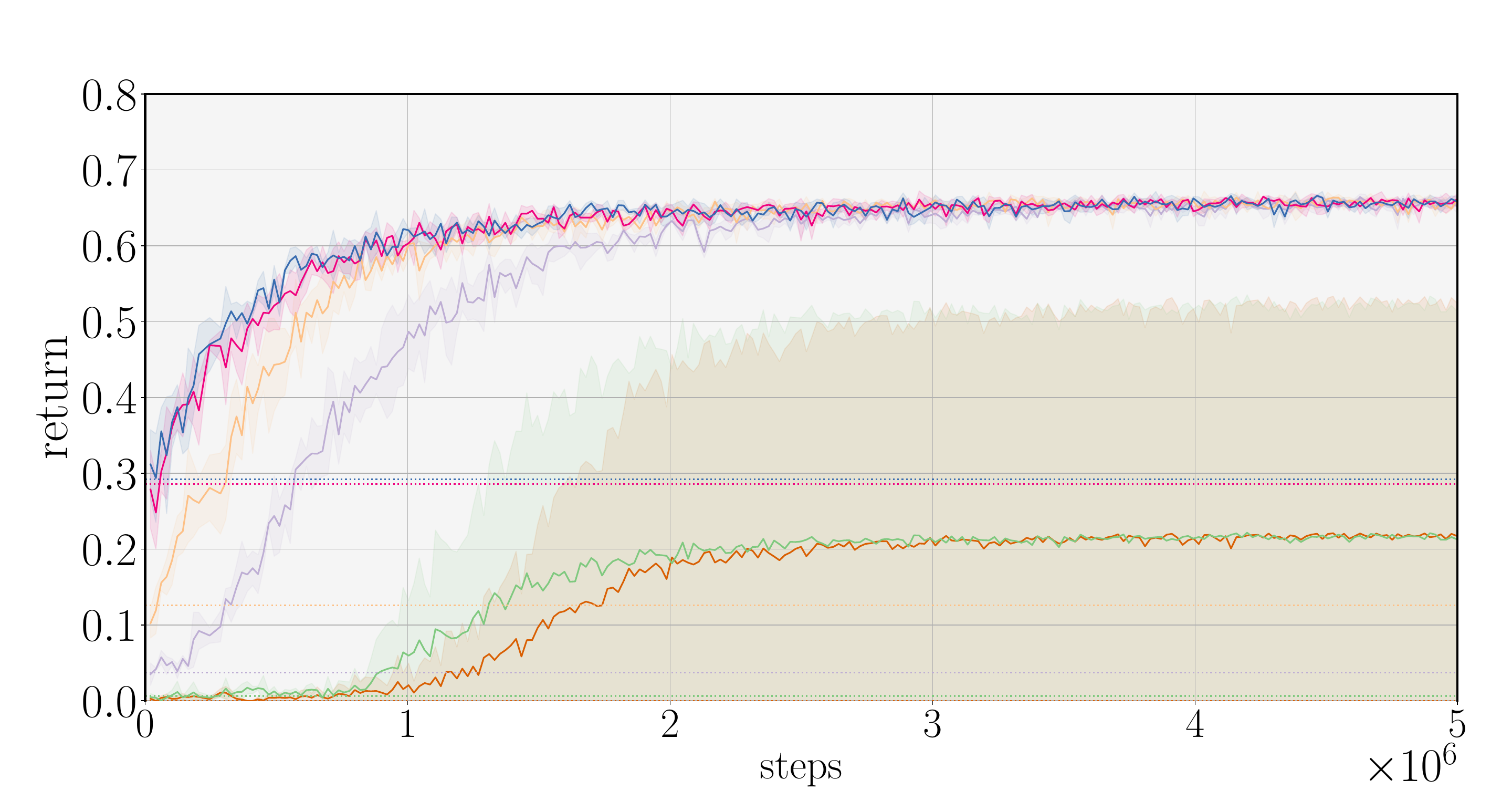} &
    \includegraphics[width=0.475\columnwidth]{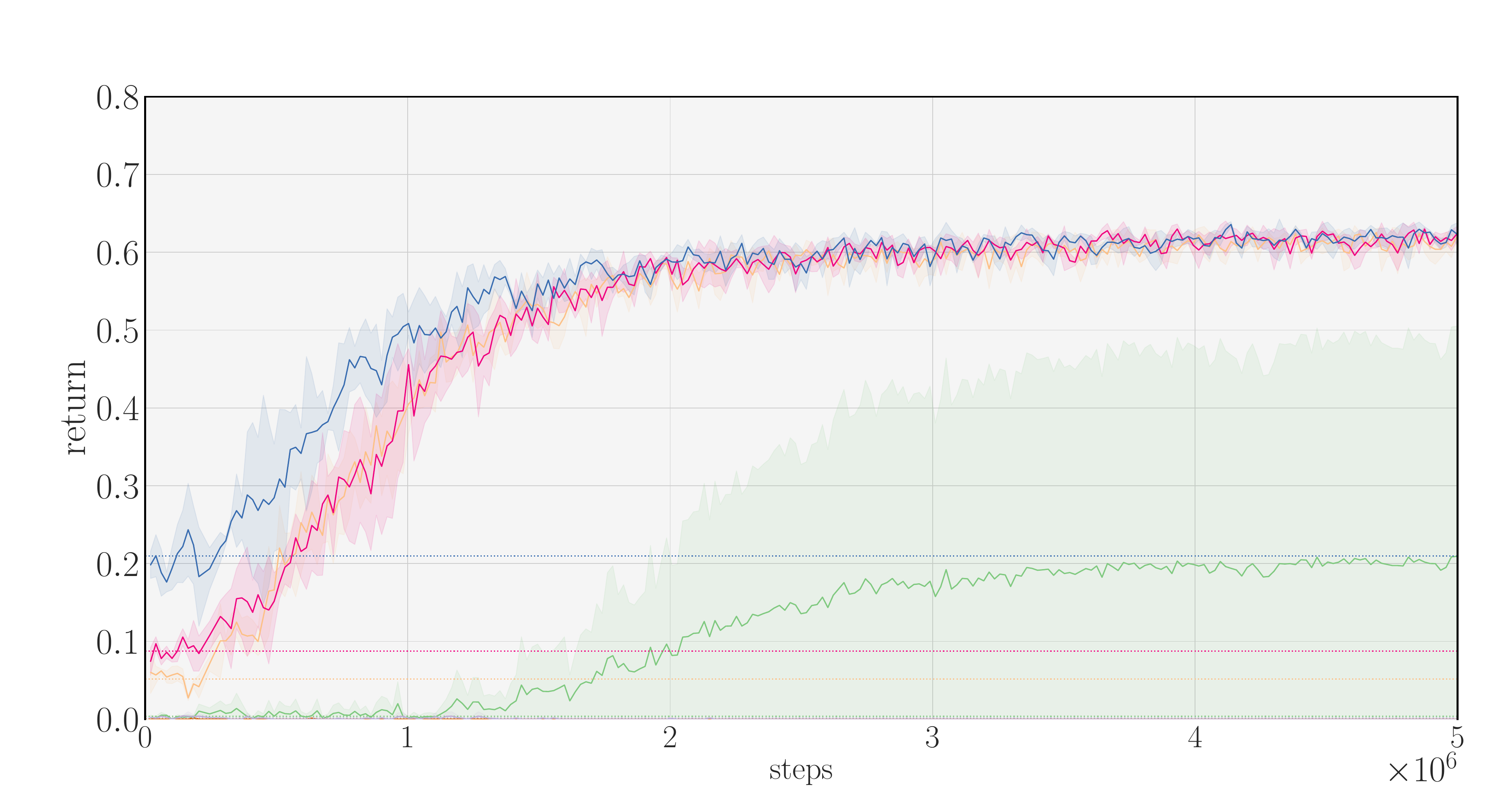}
    \\
    \rotatebox[origin=l]{90}{10\% Buffer}&
    \includegraphics[width=0.475\columnwidth]{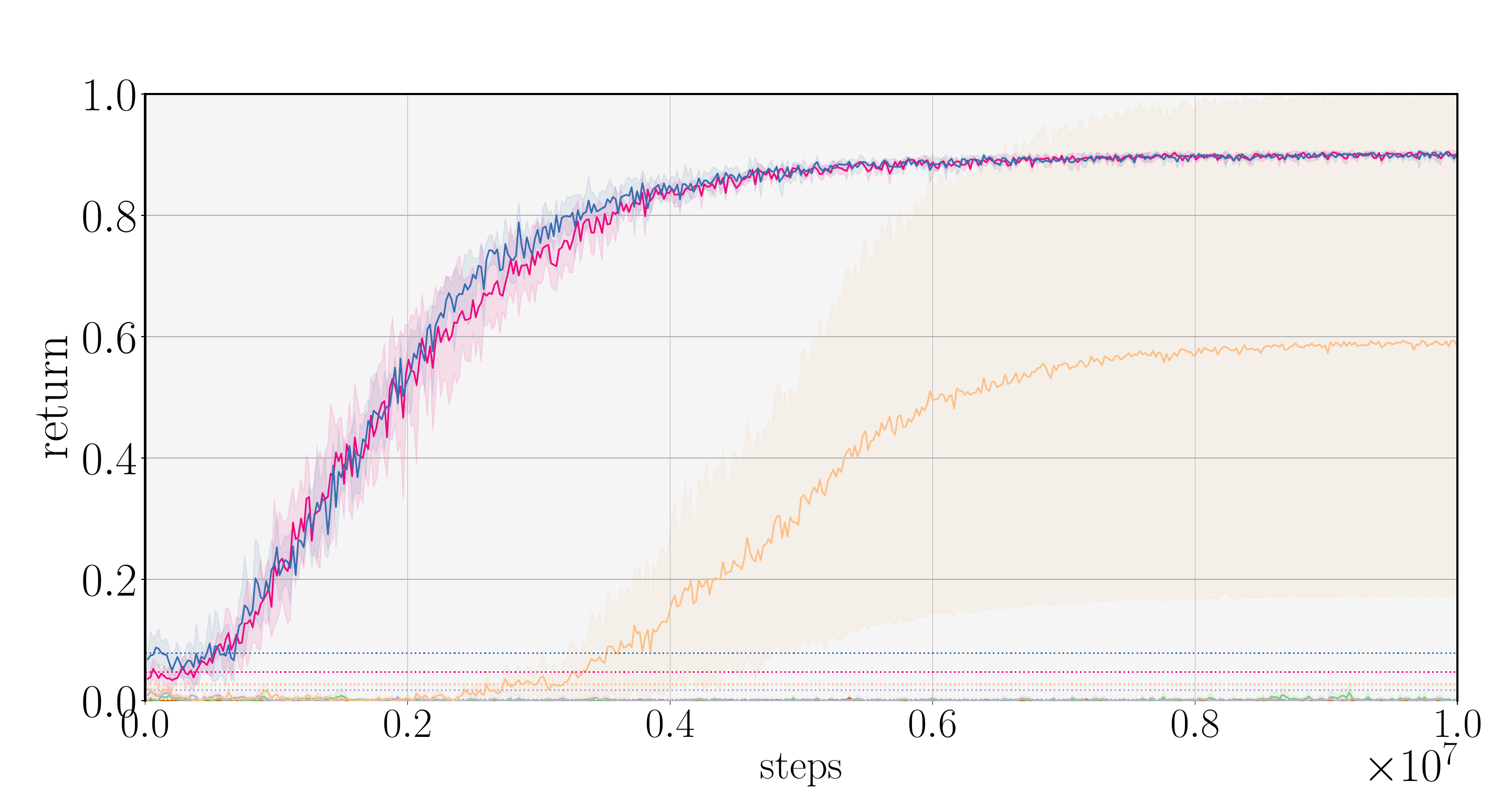} &
    \includegraphics[width=0.475\columnwidth]{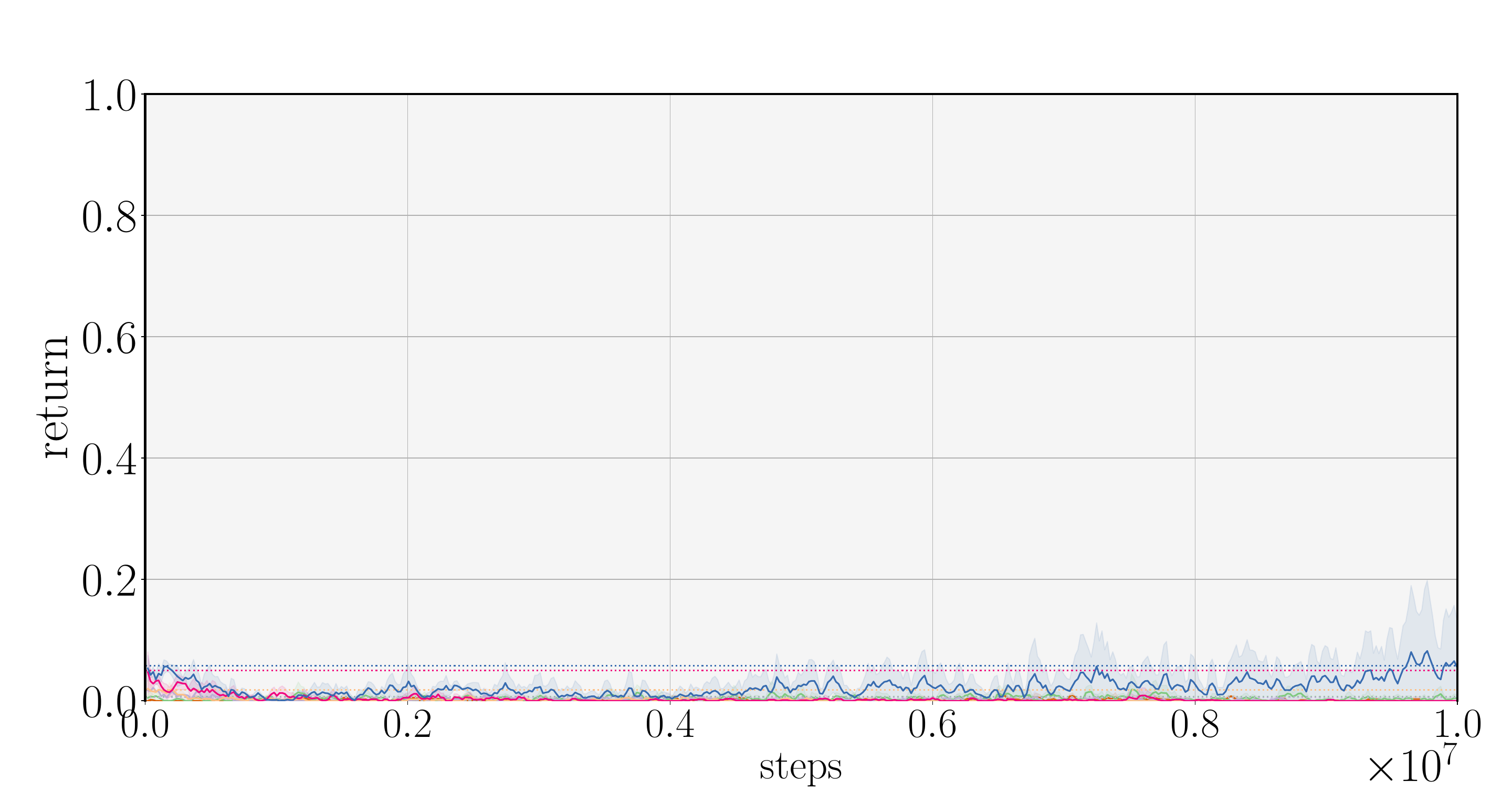} &
    \includegraphics[width=0.475\columnwidth]{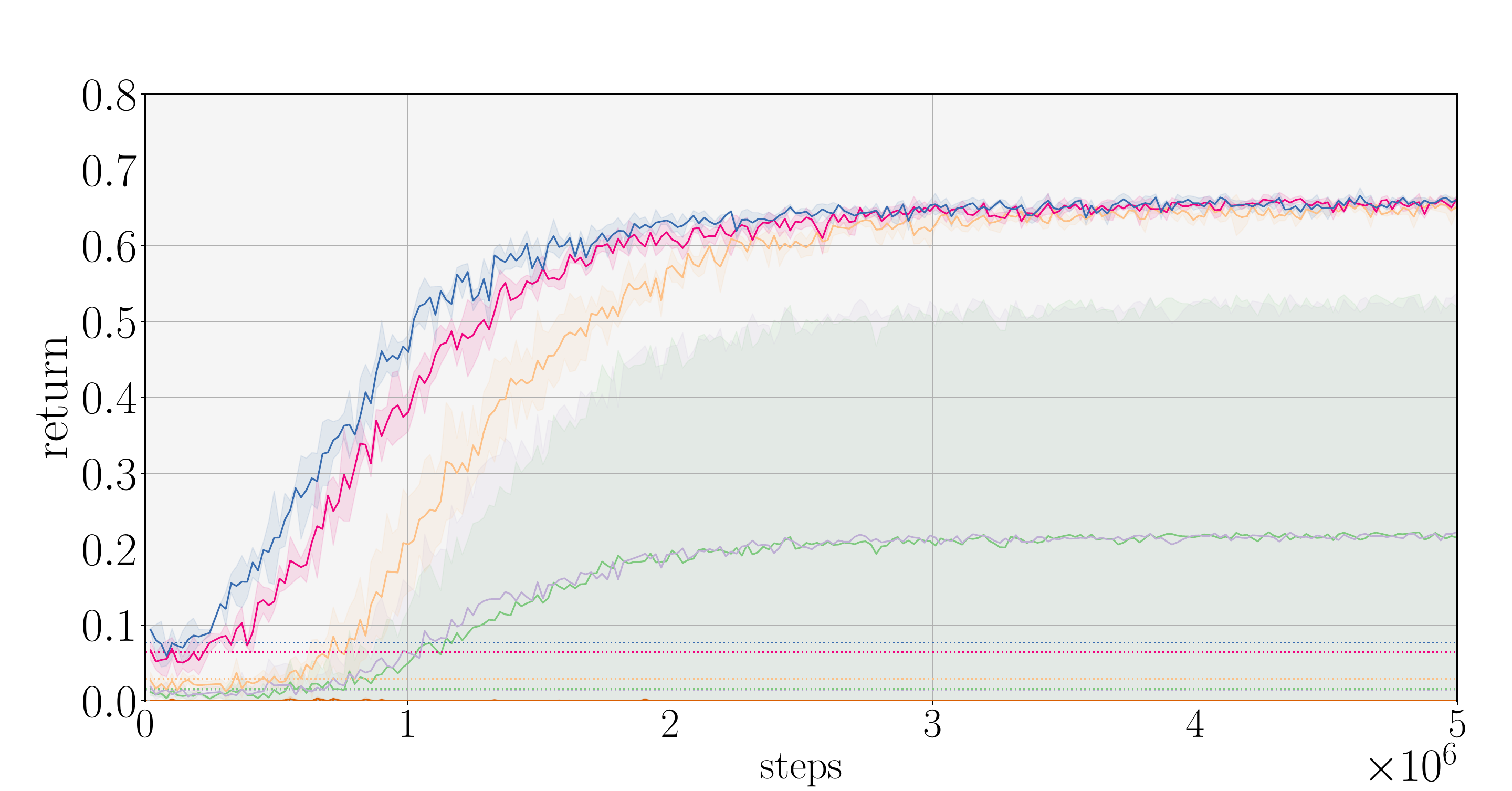} &
    \includegraphics[width=0.475\columnwidth]{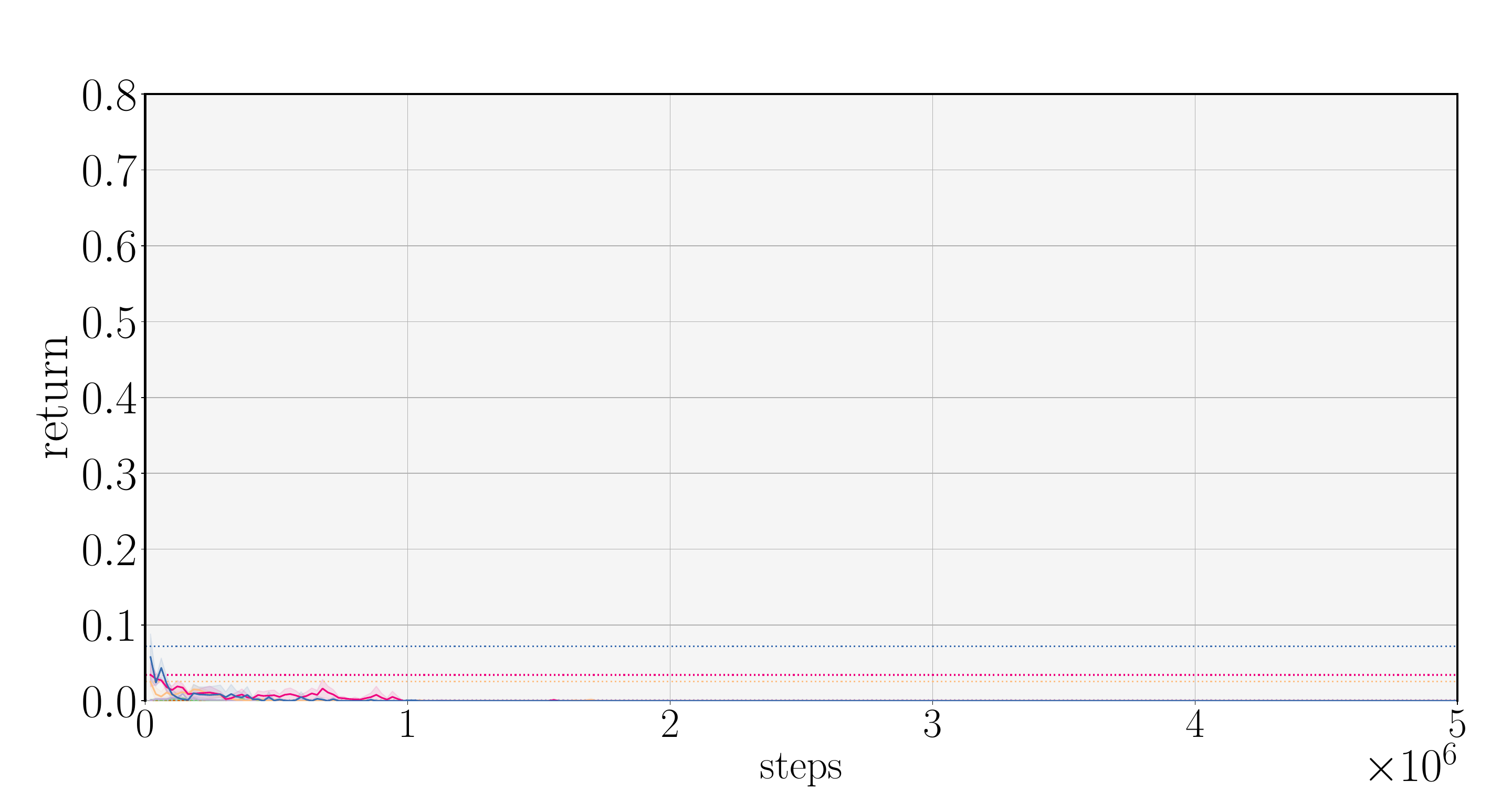}
    \\
    \end{tabular}}
    \caption{Agent performance when initializing the agent policy after pre-training it with IL with different fixed number of trajectories (one per level) that are considered of high-quality (top), medium-quality (middle) or low-quality (bottom). We provide the results, from left to right, for: \texttt{O1Dlhb}, \texttt{O2Dlh}, \texttt{MN7S8} and \texttt{MN12S10}. As in Figure \ref{fig:pre_training}, the dashed lines represent the BC pre-trained policies' evaluation score.}
    \label{fig:diversity}
\end{figure*}

\begin{figure*}[ht]
    \centering
    \resizebox{2\columnwidth}{!}{\begin{tabular}{c|ccccc}
    \rotatebox[origin=l]{90}{90\% Buffer}&
    \includegraphics[width=0.175\textwidth]{29_crop.pdf} &
    \includegraphics[width=0.175\textwidth]{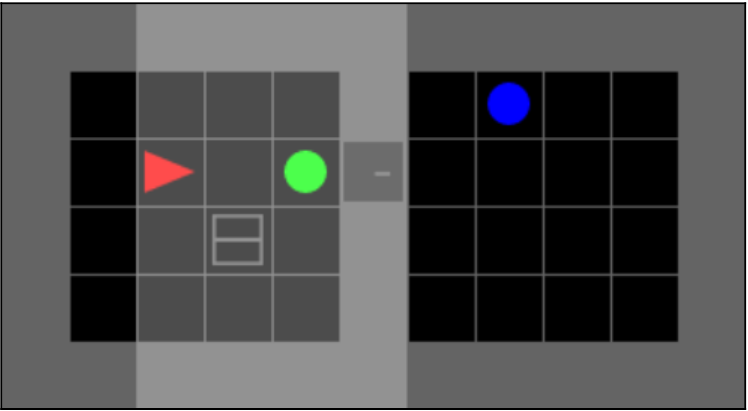} &
    \includegraphics[width=0.175\textwidth]{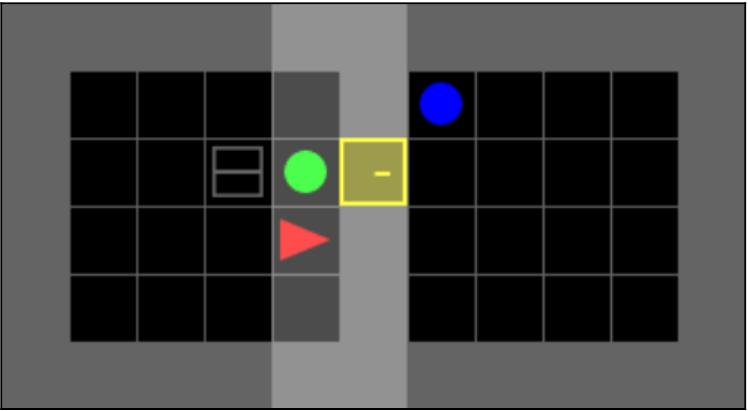} &
    \includegraphics[width=0.175\textwidth]{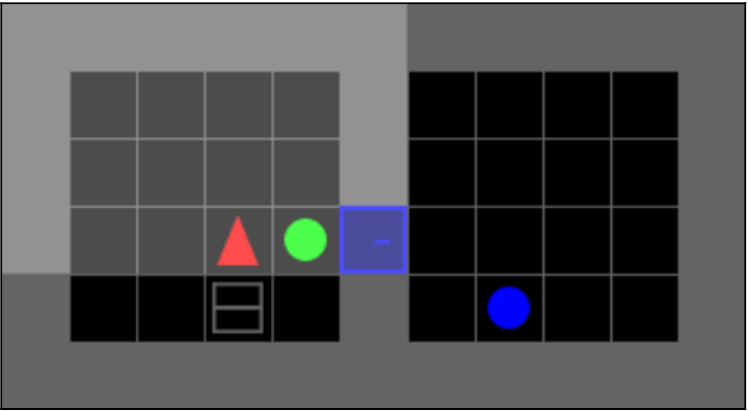} &
    \includegraphics[width=0.175\textwidth]{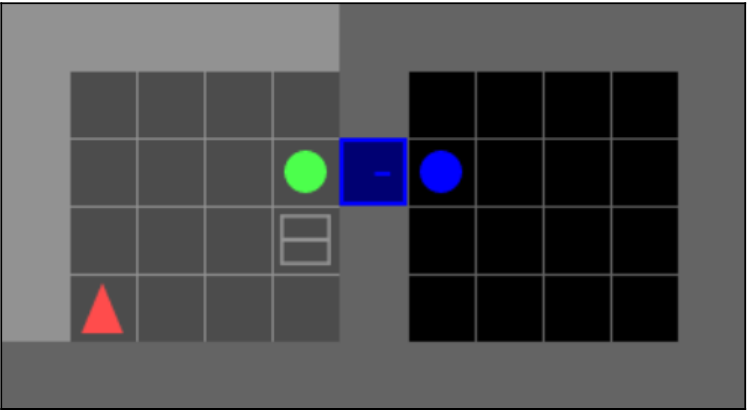} 
    \\
    &0.95 (16) & 0.941 (19) & 0.941 (19) & 0.941 (19) & 0.938 (20)
    \\
    \rotatebox[origin=l]{90}{60\% Buffer}&
    \includegraphics[width=0.175\textwidth]{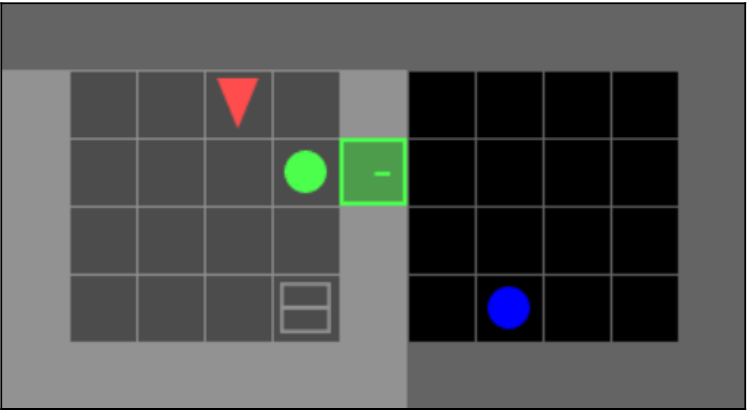} &
    \includegraphics[width=0.175\textwidth]{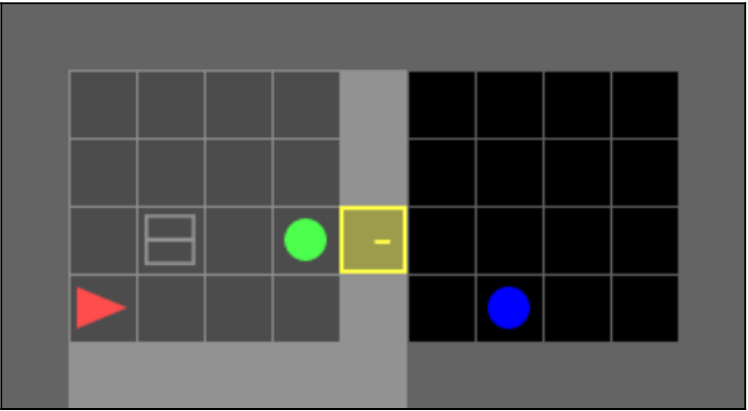} &
    \includegraphics[width=0.175\textwidth]{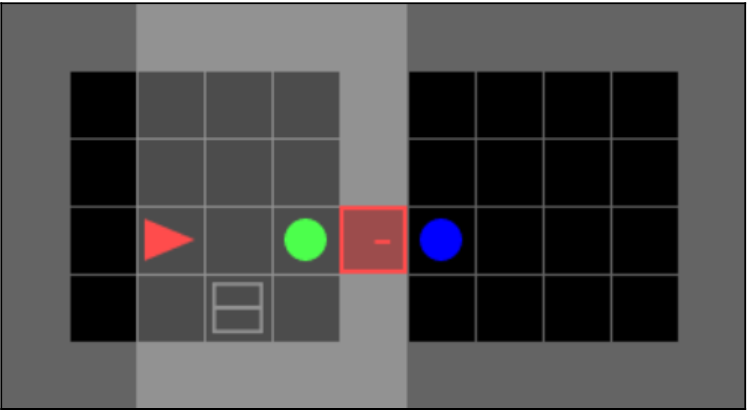} &
    \includegraphics[width=0.175\textwidth]{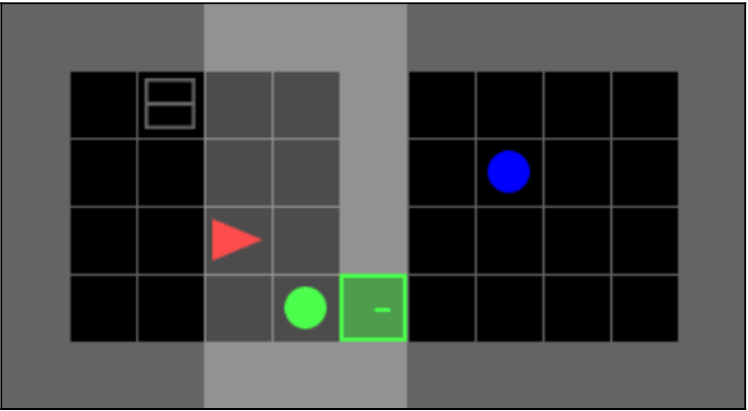} &
    \includegraphics[width=0.175\textwidth]{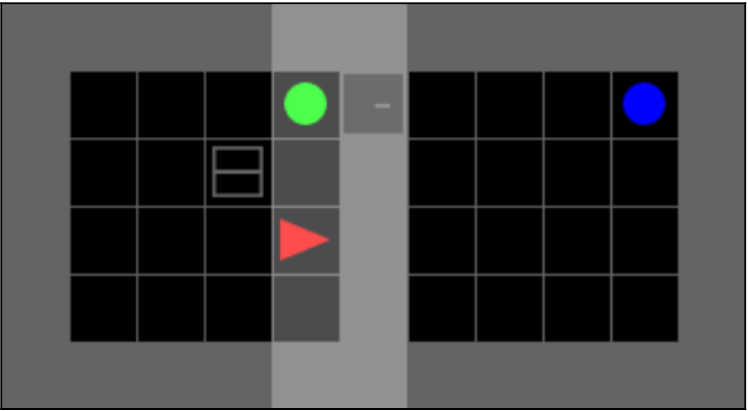}
    \\
    & 0.875 (40) & 0.909 (29) & 0.897 (33) & 0.856 (46) & 0.872 (41)
    \end{tabular}}
    \caption{Illustration of 5 levels present in the 90\% and 60\% buffers collected from \texttt{O1Dlhb}. Below each level's maze figure, the associated return (and corresponding steps) of the collected trajectory/demonstration to be mimic are shown.}
    \label{fig:o1dlhb_5levels_opt_vs_subopt}
\end{figure*}

\subsubsection{Sensitivity: Quantity and Diversity of Demonstrations} \label{subsubsec:results_div_minigrid}

In this section we analyze the sensitivity of using IL in a low-data regime: training the agent with IL using a low number of different levels, each characterized by a single trajectory. Our goal is to expose that limited data diversity and quantity influence the efficacy of IL across various tasks. To this end, we investigate the impact of imitation learning for pre-training or concurrent training.

\paragraph{Pre-training}\label{subsec:results_div_pretraining}
We delve into how trajectories from different solution qualities, categorized as 90\% buffer (high quality), 60\% buffer (medium quality), and 10\% buffer (low quality), significantly influence the benefits derived from IL. Figure ~\ref{fig:diversity} shows that \textbf{the agent manages to effectively solve \texttt{O1Dlhb}, \texttt{O2Dlh}, \texttt{MN7S8} and \texttt{MN12S10} tasks when only provided with as low as 2 and up to 20 different trajectories}. Moreover, the quality of those demonstrations can positively decrease the number of agent-environments interactions. However, the value of employing high quality demonstrations can be hampered if the diversity of the levels they represent is biased.

While pre-training with an increased number of trajectories does improve sample-efficiency, a small amount of trajectories was already sufficient to achieve these benefits in most environments. For instance, in \texttt{O1Dlhb} only incremental benefits can be observed beyond just training on 5 trajectories. Similarly, in both \texttt{MN7S8} and \texttt{MN12S10}, the agent achieves success when using 5 or more levels; however, with fewer than 5 levels, the likelihood of the agent's failure escalates significantly.
In addition, when adopting low-quality demonstrations, the agent either struggles regardless of the number of selected demonstrations (as seen in \texttt{O2Dlh} and \texttt{MN12S10}), or requires more demonstrations (and more time steps) to be able to solve the task (evident in \texttt{O1Dlhb} and \texttt{MN7S8}).

\begin{figure}[t]
    \centering
    \includegraphics[width=0.9\columnwidth]{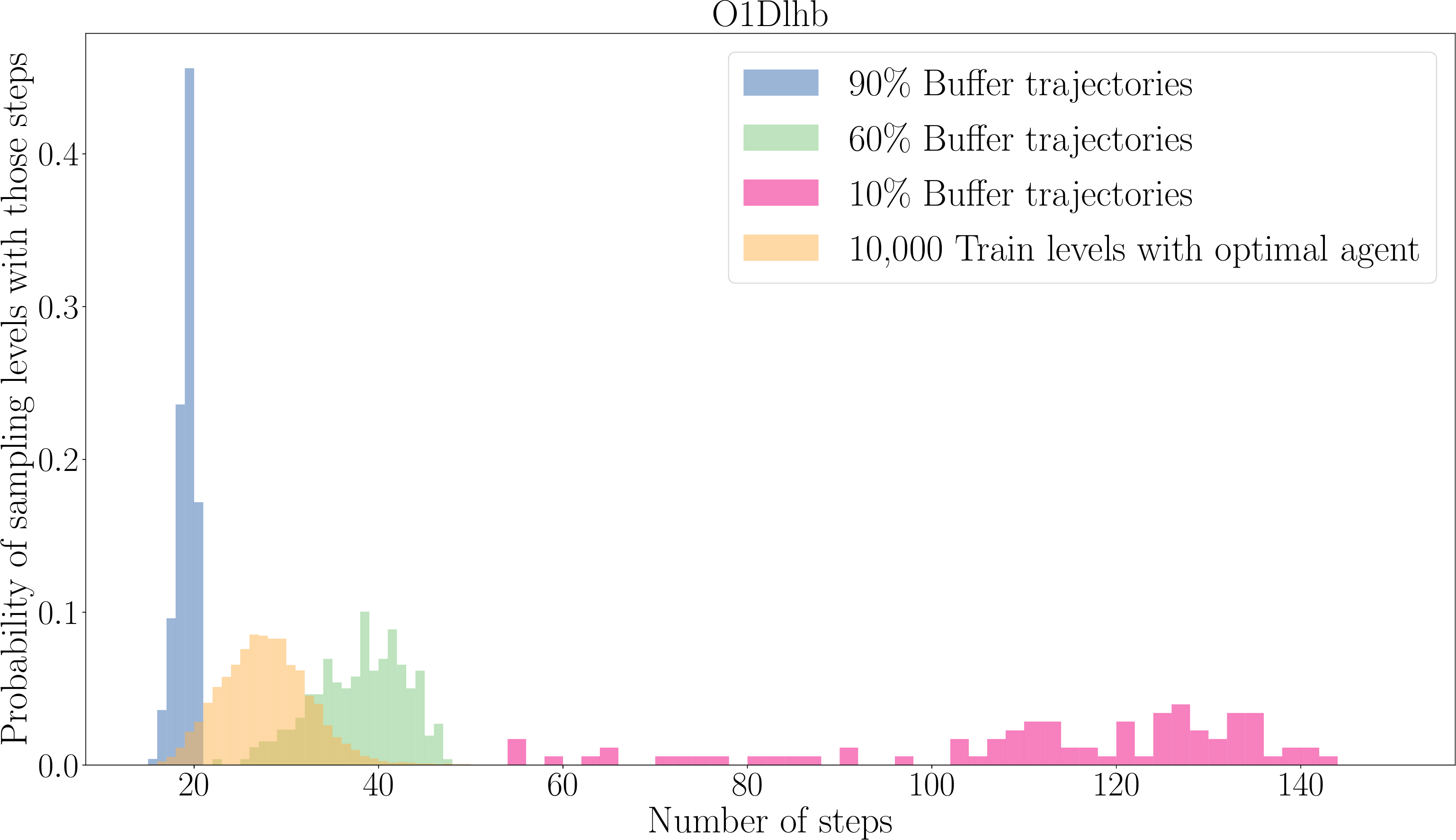}
    \caption{Probability distribution of sampling trajectories with variable number of steps from the 10\% (pink), 60\% (green) and 90\% (blue) buffers. The same distribution is provided when doing it across the 10,000 train levels with an optimal agent (orange).}
    \label{fig:level_step_distributions}
\end{figure}

By analyzing Figure \ref{fig:diversity} for each scenario, we can note that training on medium-quality demonstrations (60\% buffer) exhibits greater robustness and performance than training on low- and even high-quality demonstrations contained in the 10\% and 90\% buffers, respectively. An example supporting this statement can be noticed in \texttt{MN12S10}, where the agent with the 90\% buffer only learns when using 10 or 20 levels (pink and blue curves, respectively), whereas with the 60\% buffer the agent can learn with as few as 5 levels (orange curve). 
We hypothesize that this occurs because of the specific levels stored (and consequently sampled) from each buffer. In order to verify this, in Figure~\ref{fig:o1dlhb_5levels_opt_vs_subopt} we show the specific sampled levels in \texttt{O1Dlhb}, together with the return and number of steps of the associated trajectory. The first 2 levels beginning from the left are used for the reported '2 levels' results in Figure~\ref{fig:diversity}; in the same way, the first 3 levels beginning from the left in Figure~\ref{fig:o1dlhb_5levels_opt_vs_subopt} are used for the reported '3 levels' in Figure ~\ref{fig:diversity}; and all the provided 5 levels in Figure~\ref{fig:o1dlhb_5levels_opt_vs_subopt} for the '5 levels' results in Figure~\ref{fig:diversity}. 

By inspecting these levels, we can see that the trajectories within the 60\% buffer are of notably lower quality compared to those in the 90\% buffer (the expected optimal number of steps required to solve levels in this task are $\sim$26, see Table ~\ref{tab:buffer_metada}), whereas the levels within the 90\% buffer contain trajectories with as few as 16-20 steps. The distribution of levels and thereby trajectories contained within the 90\% buffer is therefore skewed towards easier levels, which require shorter trajectories than those expected for levels of this environment.

Therefore, there are two main possible reasons that might explain why the agent pre-trained with few trajectories from the 90\% buffer exhibits worse results: 
\begin{enumerate}
    \item The stored distribution of levels: each trajectory contained in the buffer belongs to a specific level which, at the same time, requires a different number of steps to be solved optimally~\cite{raileanu_decoupling_2021}. Thus, some levels can be considered easier due to them requiring fewer steps, leading to trajectories with higher returns. The RAPID prioritization makes trajectories belonging to such easier levels prevail over trajectories of other levels~\cite{andres_towards_2022}, causing a shift in the distribution of stored levels.
    \item The coverage and interactions represented by the trajectories within these levels: The length of medium-quality trajectories in MiniGrid is longer than those of high-quality, thereby covering a larger part of the state space. Possible interactions with the environment might be beneficial for learning skills required in the task.
\end{enumerate}

Regarding the first hypothesis, we visualize the probability distribution of sampling trajectories depending on their number of steps in Figure~\ref{fig:level_step_distributions}. The distribution related to the steps needed to complete each task by an optimal agent across 10,000 train levels (orange) is not covered by any of the buffers. For the 90\% buffer, the overlap with this distribution is fairly small, clearly indicating that the levels covered within this buffer are skewed towards levels with shorter optimal solutions between 15 and 21 steps. In contrast, the 10\% buffer only contains low-quality trajectories encompassing trajectories between 50 and 150 steps. Only the 60\% buffer contains a notable number of levels which are representative of the data distribution generated by an optimal agent (with levels requiring between 20 and 40 steps to be solved), which might explain the results shown in Figure~\ref{fig:diversity}.

For our second hypothesis, we raise the following question: \textit{How would the agent's learning be influenced if we were to train it using trajectories from identical levels selected from both the 90\% and 60\% buffers, where the only difference is the quality of the demonstrations and not their diversity?}

\begin{figure}[t]
    \centering
    \includegraphics[width=\columnwidth]{diversity_legend_crop.pdf}
    \\
    \includegraphics[width=0.475\columnwidth]{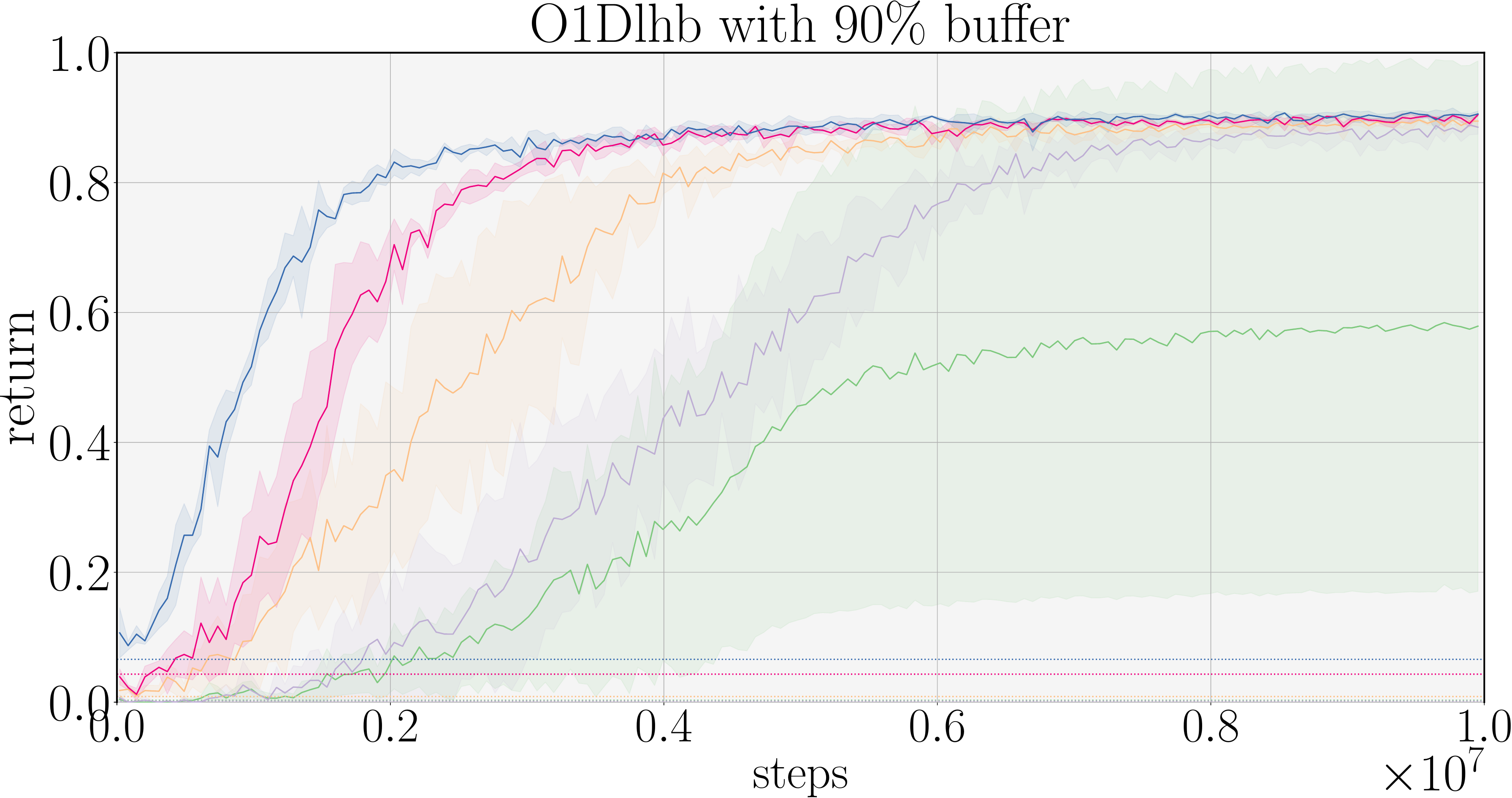}
    \includegraphics[width=0.475\columnwidth]{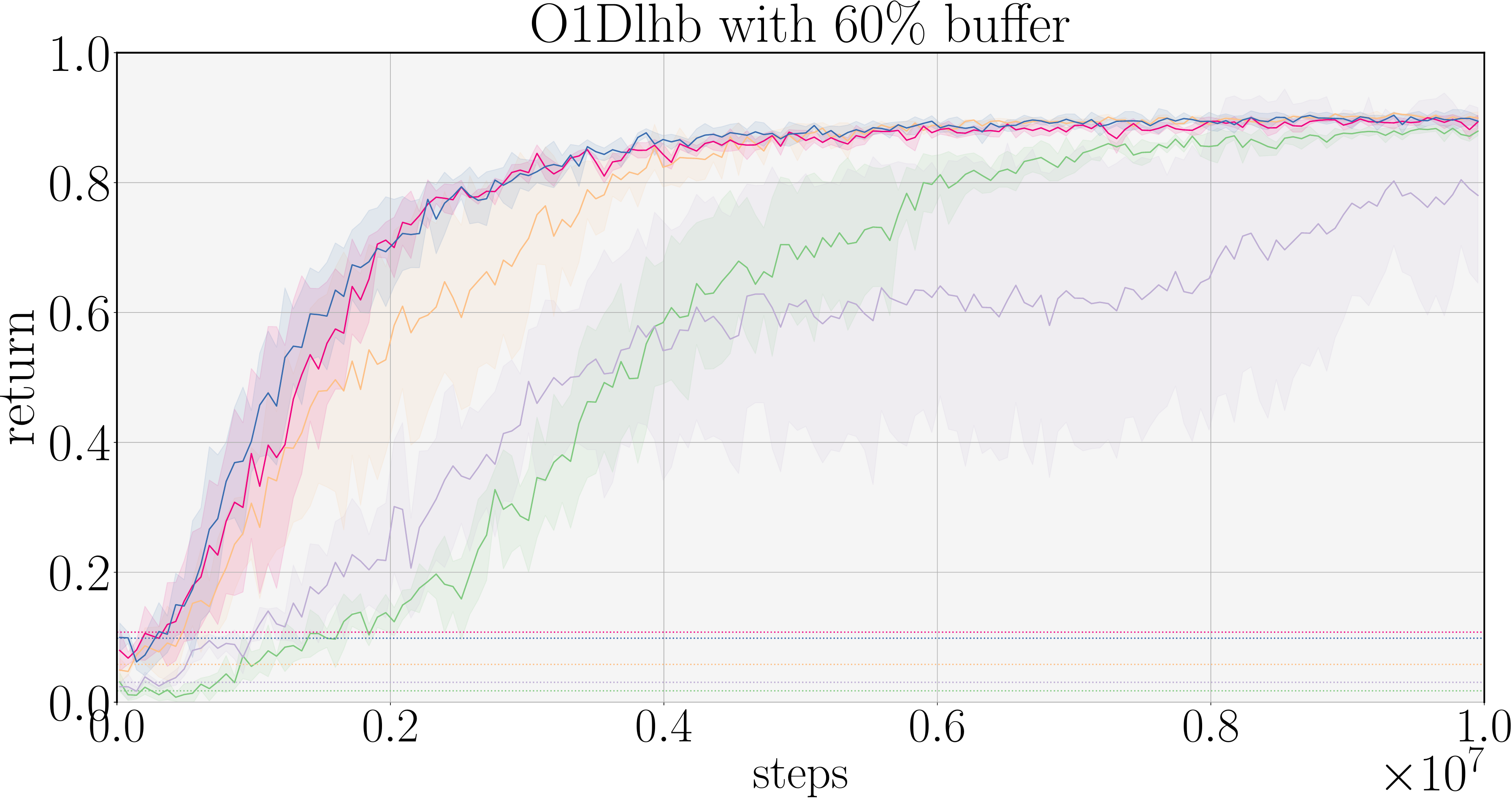} 
    \caption{
    Agent's performance in \texttt{O1Dlhb} when initializing the agent policy after pre-training it with IL with a fixed set of trajectories. Only a single demonstration per level is considered. The same 20 levels are used for both the 90\% and 60\% buffers, ensuring consistent level diversity across demonstrations that differ in quality.
    }
    \label{fig:o1dlhb_diversity_common_levels}
\end{figure}

To investigate this, we select 20 levels and acquire two trajectories for each level: one of medium-quality (present at the 60\% buffer), and another of high-quality (from the 90\% buffer). In this way, we ensure that the variation in the learning process is attributable to the quality of demonstrations, independent of diversity. Considering this controlled setup, in Figure~\ref{fig:o1dlhb_diversity_common_levels} we further analyze the influence of IL at pre-training for the \texttt{O1Dlhb} task. When considering the same levels --yet different quality of trajectories-- the results are very similar: agents trained from the 60\% and 90\% buffer exhibit robustness issues when using only 2 or 3 levels with instabilities being more severe for the 90\% buffer. However, the agents pre-trained from the 90\% buffer seem to converge slightly faster when having a larger amount of levels available. 

In light of these results, we can state that the selection of levels (distribution shift of levels) used for pre-training is perhaps surprisingly more important than the quality and quantity of the trajectories. This explains why in Figure~\ref{fig:diversity} the 90\% buffer reports worse results compared to the 60\% buffer: the trajectories contained in the 90\% buffer belong to levels that do not represent the whole level distribution, which can be seen in the mismatch between $\mu_{G(\tau)}$ and $\mathbb{E}^*[G(\tau)]$ in Table ~\ref{tab:buffer_metada} for all the considered environments and also in the mismatch of probability distributions shown in Figure~\ref{fig:level_step_distributions}.

In summary, using IL to pre-train RL agents with only a handful of demonstrations can significantly speed-up the learning. Moreover, when using few demonstrations, it is more important to select trajectories belonging to the whole spectre of the level distribution (i.e., maximize the diversity of the levels) rather than providing optimal examples. 
\begin{figure*}[h]
    \centering
    \begin{tabular}{ccc}
    \multicolumn{3}{c}{\includegraphics[width=\columnwidth]{diversity_legend_crop.pdf}} 
    \\
    \texttt{O1Dlhb 10\% Buffer} & \texttt{O1Dlhb 60\% Buffer} & \texttt{O1Dlhb 90\% Buffer} 
    \\
    \includegraphics[width=0.65\columnwidth]{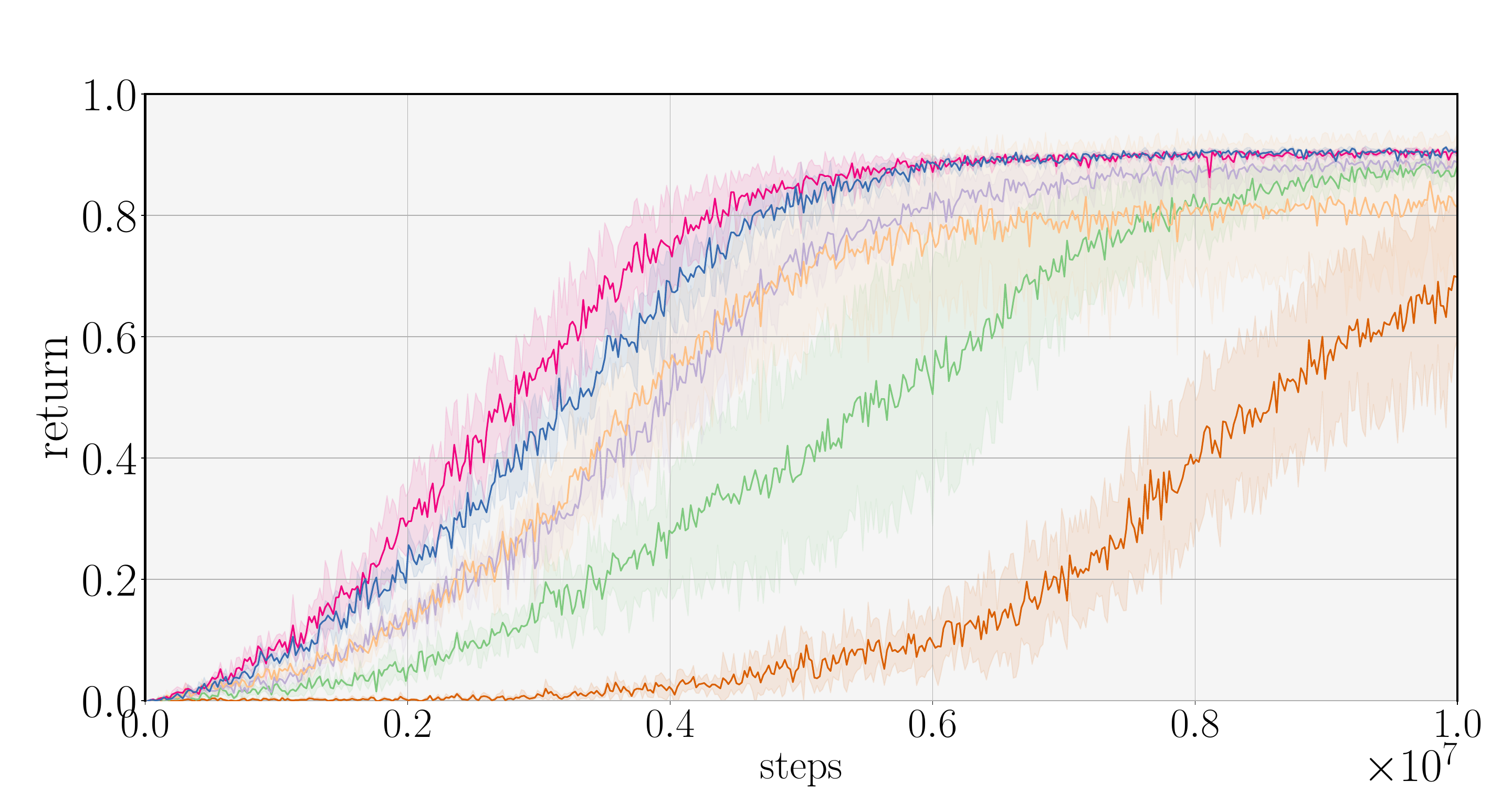} &
    \includegraphics[width=0.65\columnwidth]{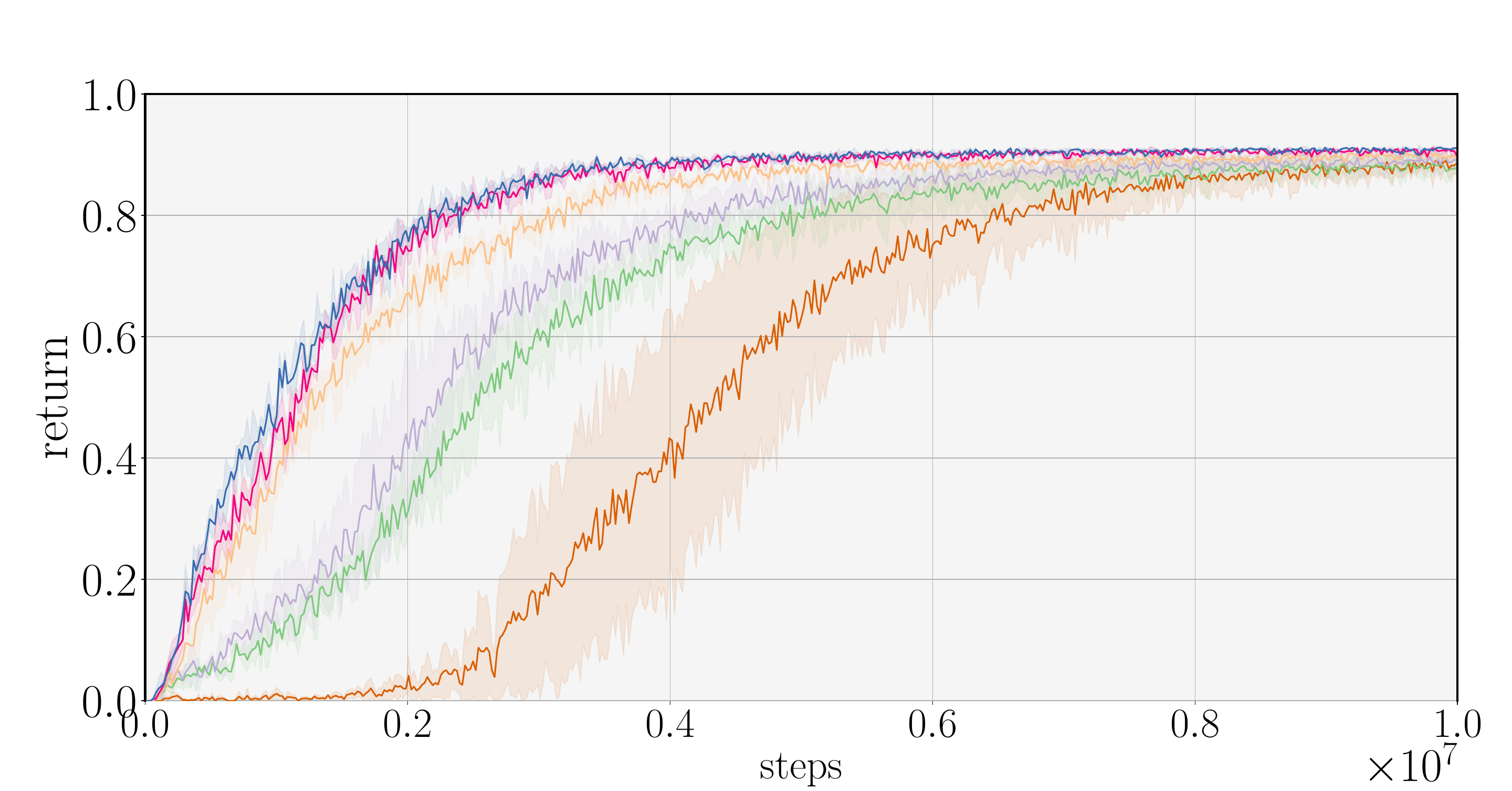} &
    \includegraphics[width=0.65\columnwidth]{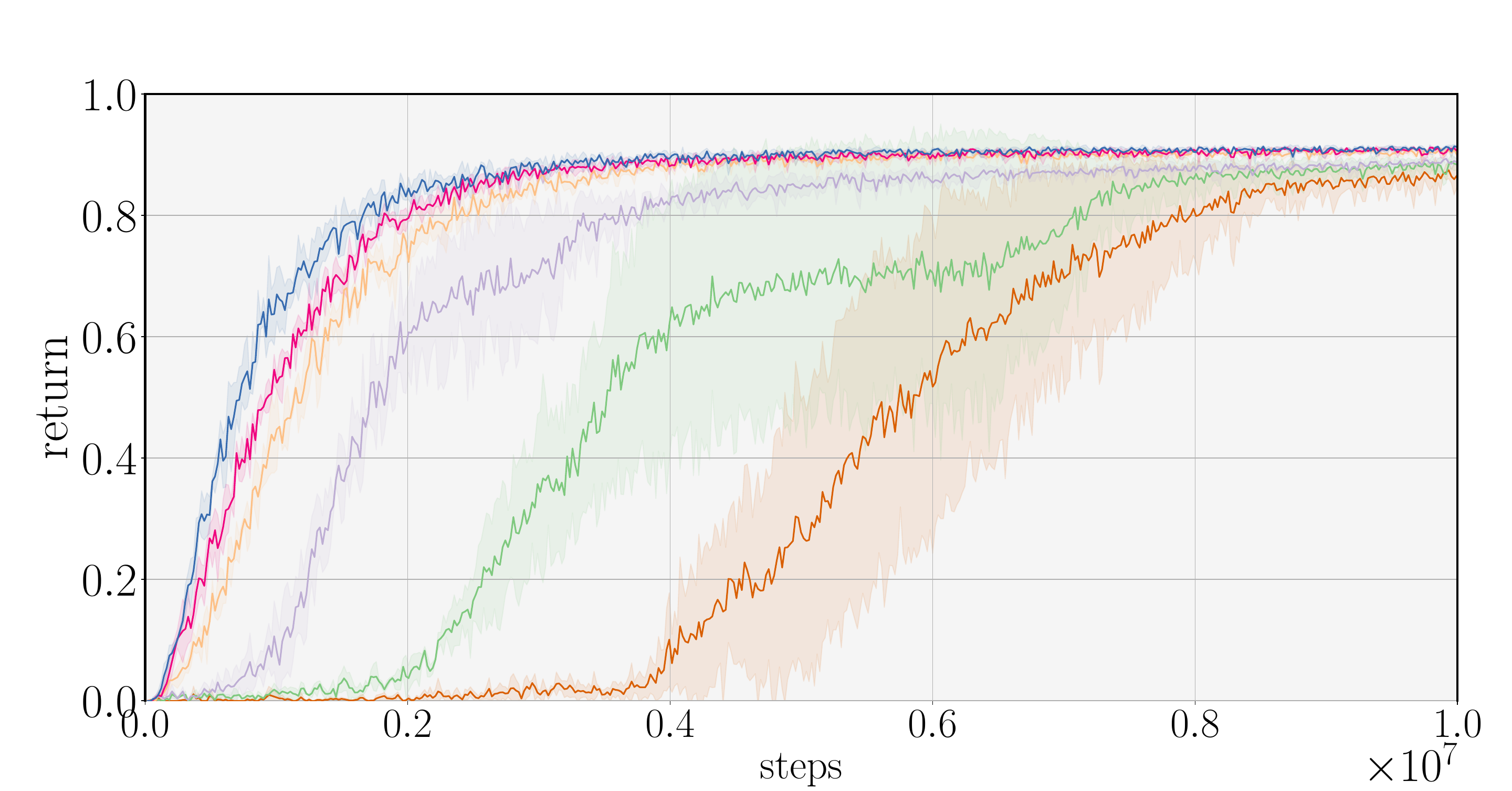}
    \\
    
    \texttt{MN12S10 10\% Buffer} & \texttt{MN12S10 60\% Buffer} & \texttt{MN12S10 90\% Buffer} 
    \\
    \includegraphics[width=0.65\columnwidth]{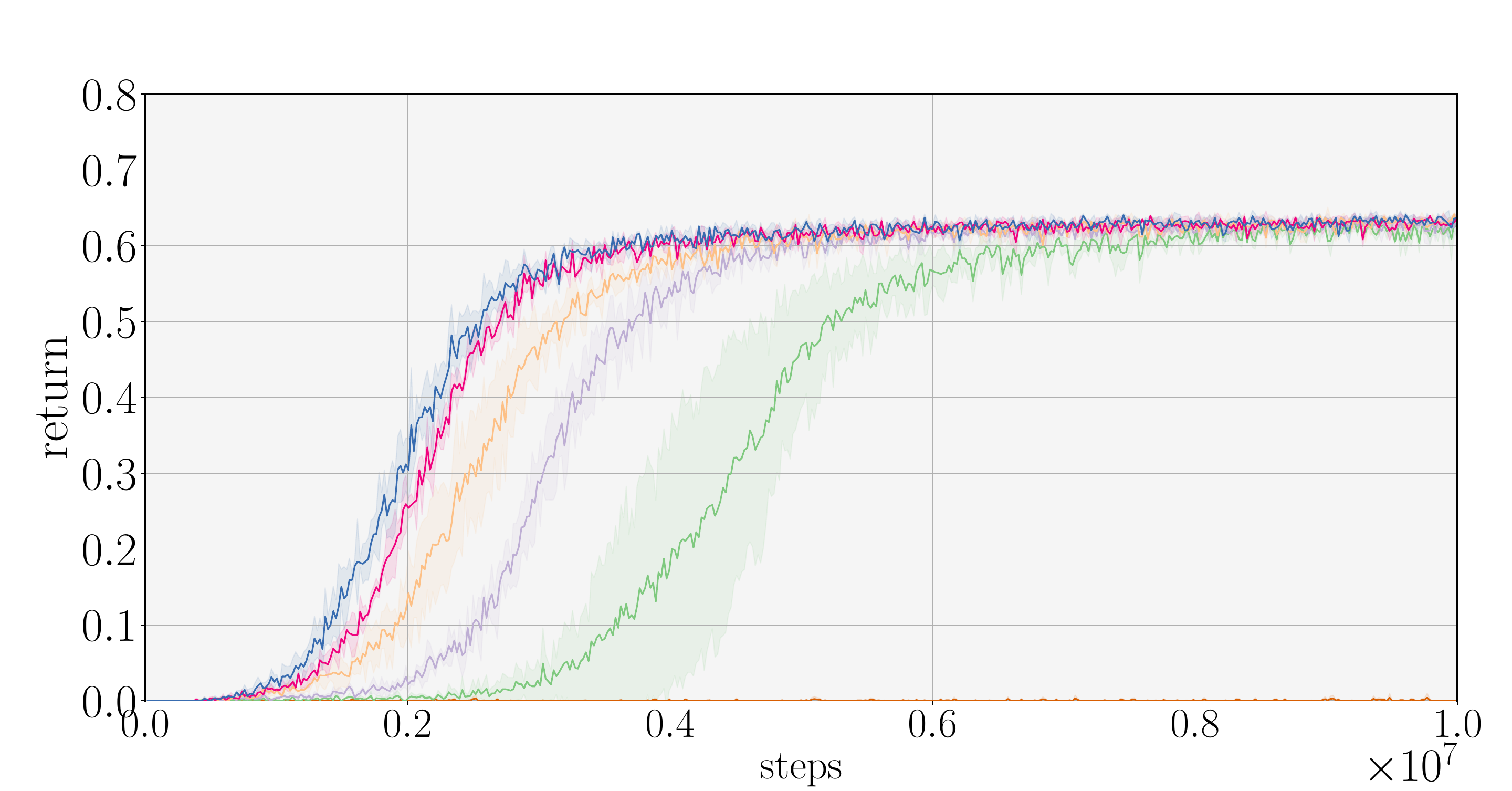} &
    \includegraphics[width=0.65\columnwidth]{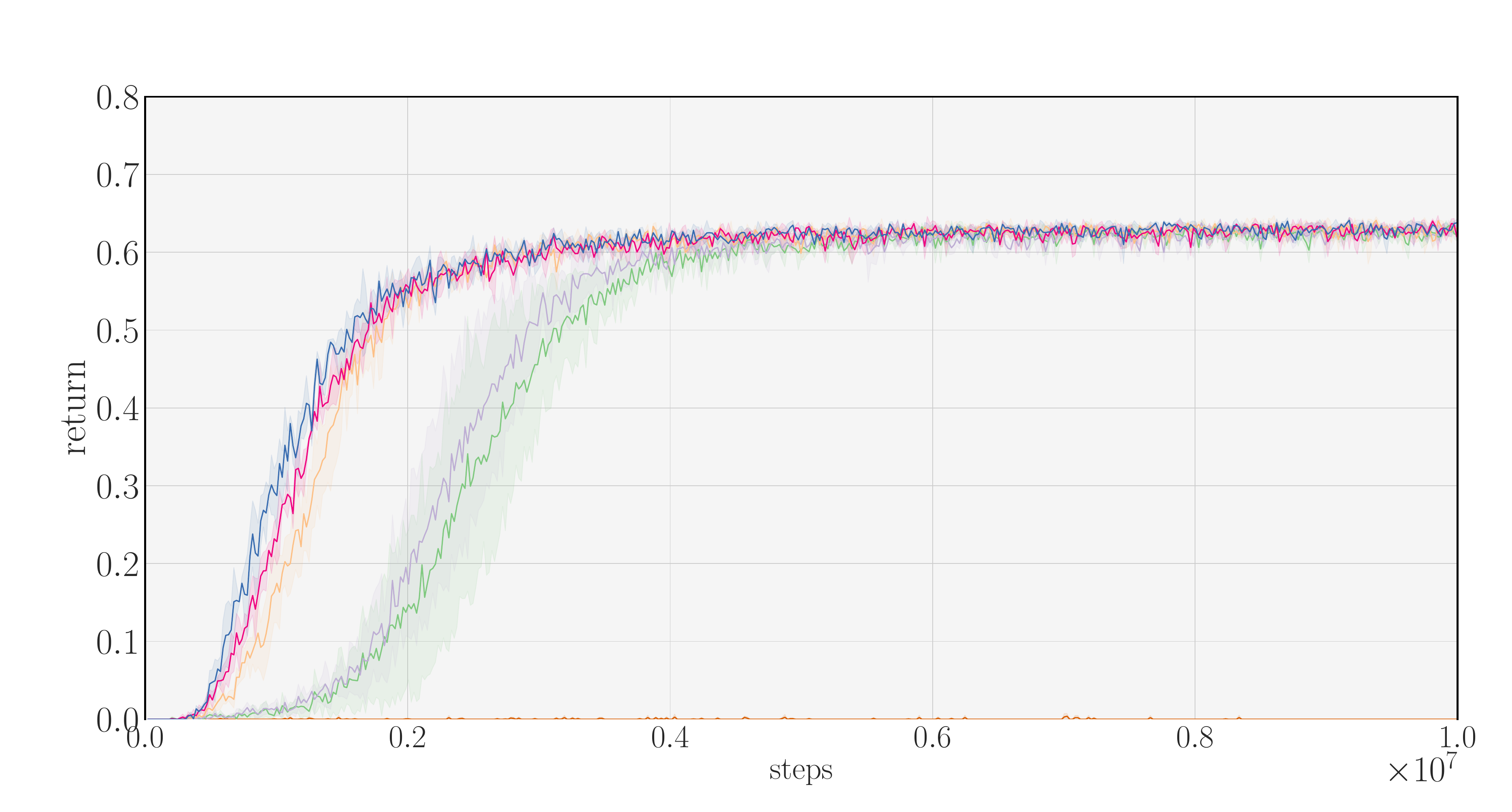} &
    \includegraphics[width=0.65\columnwidth]{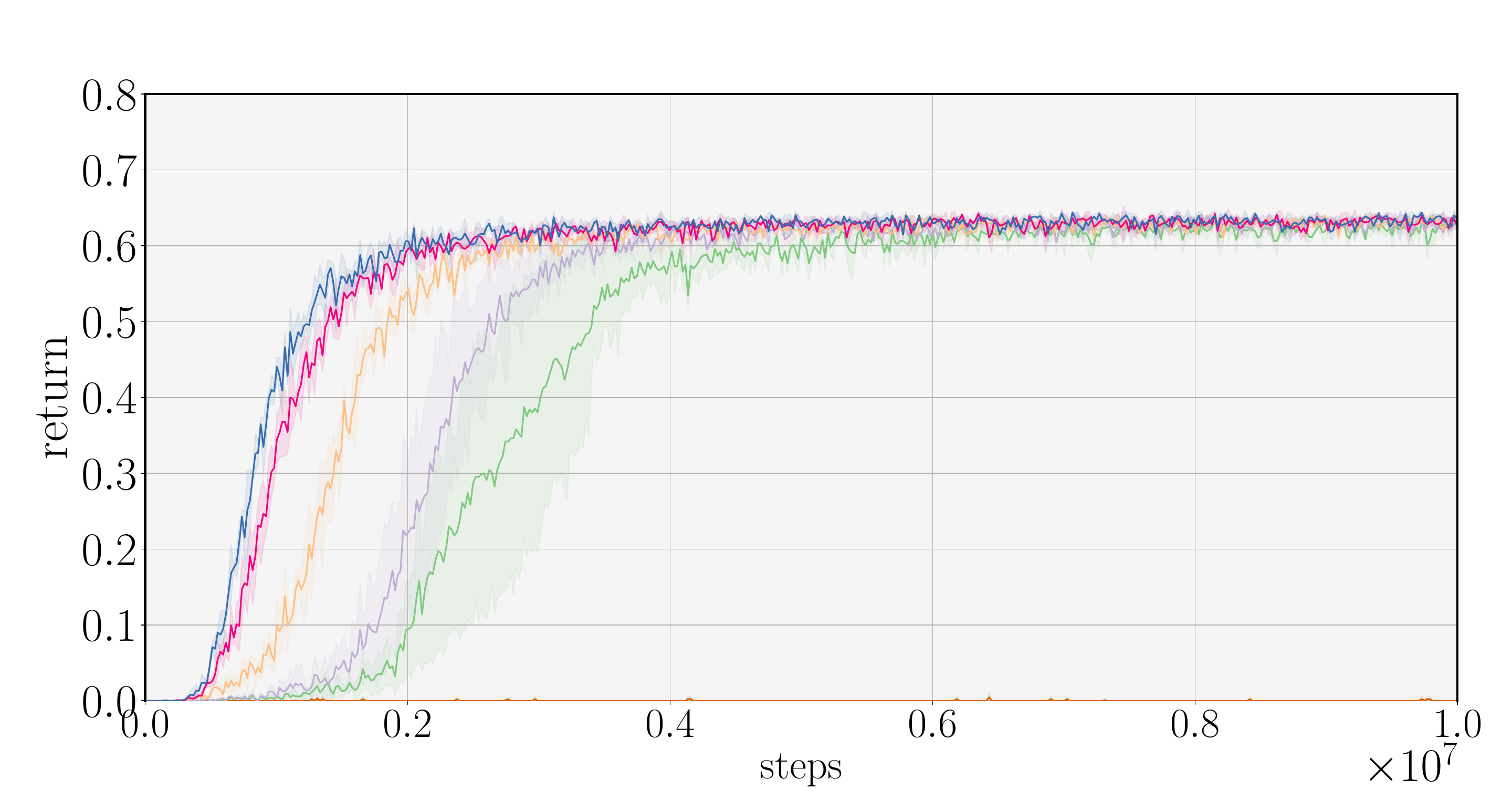}
    \\
    \end{tabular}
    \caption{Agent's performance when randomly initializing the policy and using concurrently IL and RL losses during the online training with different fixed number of trajectories (one per level) at \texttt{O1Dlhb} and \texttt{MN12S10}. The buffer is initialized with the provided demonstrations and they are repeated until filling the whole buffer capacity (i.e., 10,000 experiences). At \ref{app:populate_buffer_online_diversity_minigrid} the impact of populating the buffer with the demonstrations (without repetition) is further analyzed.}
    \label{fig:diversity_online_minigrid}
\end{figure*}
\paragraph{Concurrent Training} \label{subsec:results_div_conctraining}
In line with the previous evaluation, we further explore the impact of using as few as 2 to 20 trajectories of varying solution qualities, namely 90\% buffer(high quality), 60\% buffer (medium quality), and 10\% buffer(low quality), when concurrently using RL and IL (without pre-training). For this purpose, we load the buffer with the available offline demonstrations, and subsequently replicate the experiences uniformly to fill the buffer to its maximum capacity. Additionally, we considered the alternative of populating the buffer exclusively with the demonstrations, while reserving space for new experiences. The outcomes of such comparison are detailed in \ref{app:populate_buffer_online_diversity_minigrid}.

As shown in Figure~\ref{fig:diversity_online_minigrid}, for \texttt{O1Dlhb} and \texttt{MN12S10} the general trend indicates a direct relationship between the quality of demonstrations and sample efficiency: higher quality demonstrations result in enhanced efficiency. Similarly, an increase in the quantity of demonstrations (which inherently increases diversity) accelerates the learning process. 

\textbf{With the adoption of concurrent learning, the agent is now capable of learning the optimal policy from as few as 2 trajectories}. This advancement starkly contrasts with prior outcomes using IL only in the pre-training phase, as detailed in Figure \ref{fig:diversity}. 
For instance, in \texttt{MN12S10}, the agent previously needed a minimum of 5 or 10 high-quality trajectories during pre-training to learn effectively. In contrast, using concurrent IL, the agent is able to successfully learn to solve the task with 2 or more trajectories irrespective of the quality of the data. This improvement is even more significant in \texttt{O1Dlhb}, where the agent learns the expected optimal behavior with just a single trajectory.

In conclusion, when dealing with a limited amount of demonstrations, the most stable and robust learning is achieved by concurrently using IL with RL during online interactions, where as few as 1 or 2 demonstrations can be sufficient for effective learning.

\subsection{Procgen} \label{subsec:procgen_results}

\subsubsection{Pre-training with Imitation Learning}

Analogously to the experimental setup executed in Mini-Grid, we evaluate the impact of using IL in pre-training for \texttt{Ninja} and \texttt{Climber} tasks available in the Procgen benchmark. 

\begin{figure}[t]
    \centering
    \includegraphics[width=0.5\columnwidth]{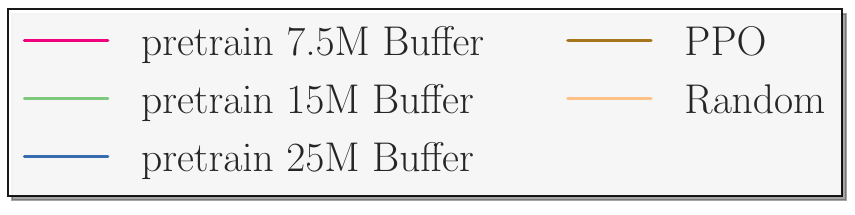}
    \\
    \includegraphics[width=0.9\columnwidth]{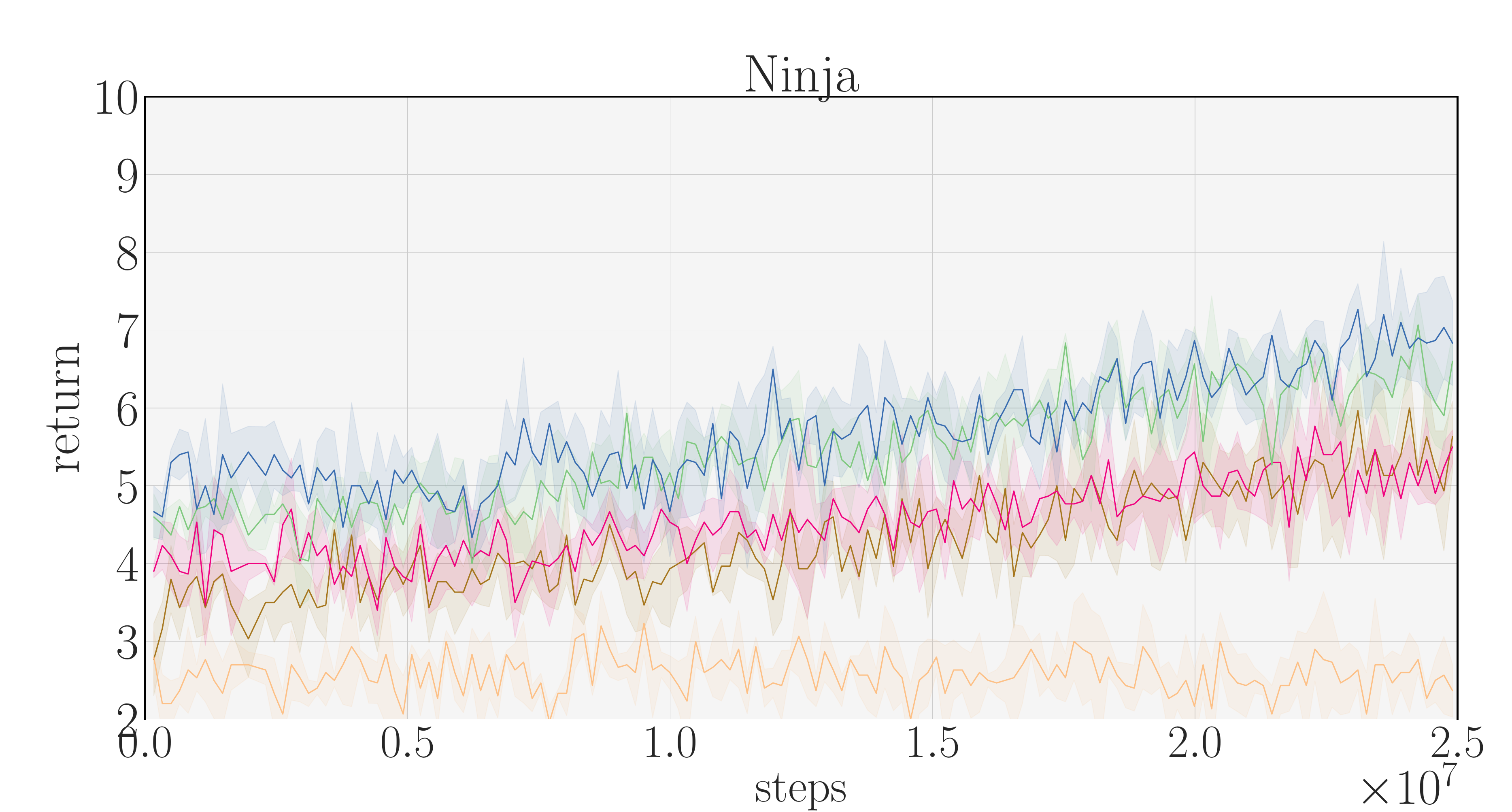}
    \includegraphics[width=0.9\columnwidth]{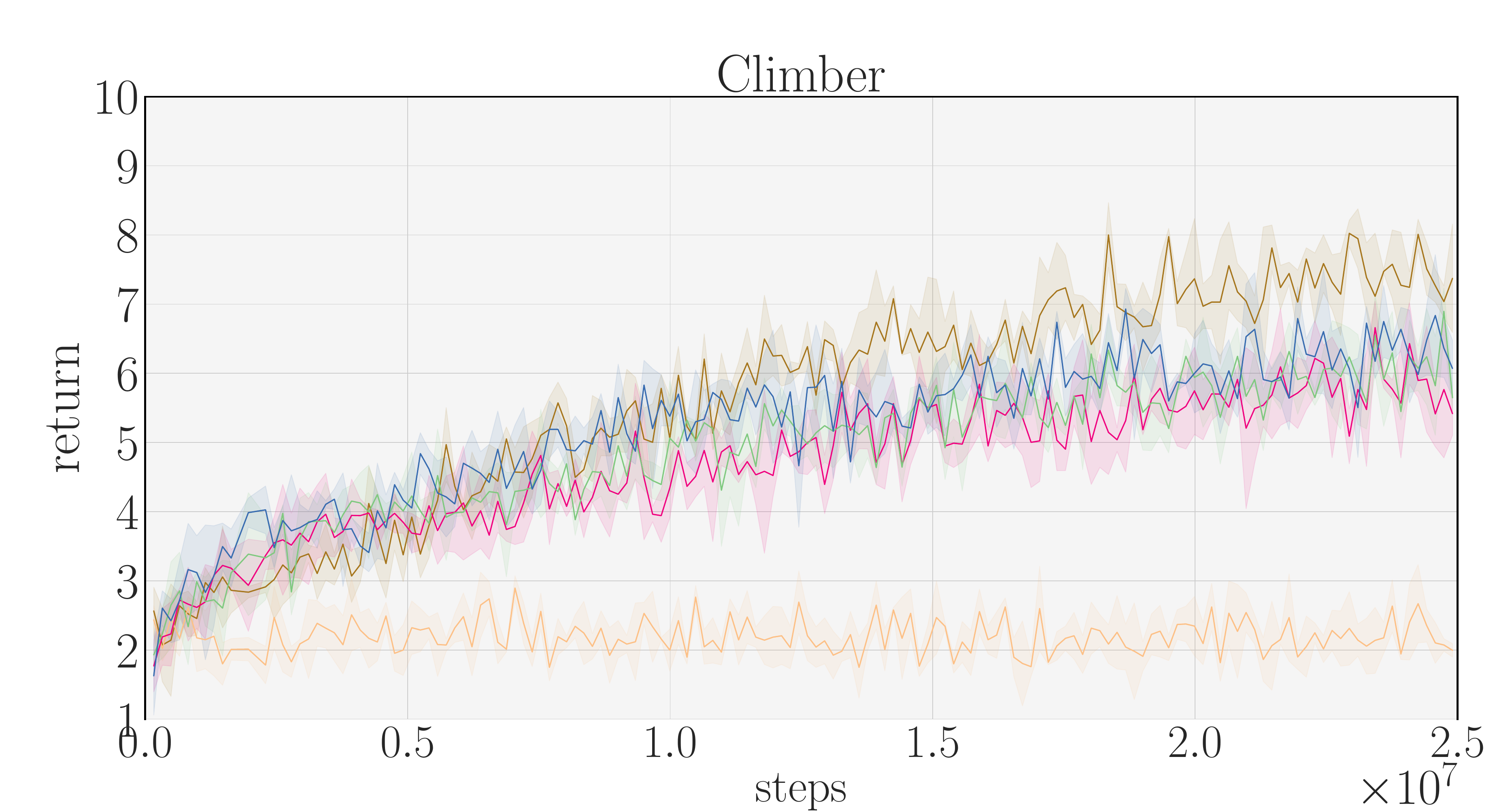}
    \\
    \caption{Performance of the agent when pre-training with IL before the RL training phase in \texttt{Ninja} and \texttt{Climber}.}
    \label{fig:pre_training_procgen}
\end{figure}

As illustrated in Figure \ref{fig:pre_training_procgen}, the results show variable outcomes depending on the task at hand. On the one hand, in \texttt{Ninja} pre-training with IL reflects a positive influence in performance when using data coming from either the 15M or 25M Buffers, outperforming in both cases the return of the PPO baseline. Conversely, the \texttt{Climber} task does not echo these advantages, as the performance in all the cases falls short of the baseline PPO. This suggests that pre-training with IL may have detrimental effects. This underwhelming performance might be attributed to the buffer's limited diversity, particularly in the \texttt{Climber} task, where, as Table \ref{tab:buffer_metada_procgen} reveals, the content is heavily biased with demonstrations belonging to just 2-4 levels.

\subsubsection{Concurrent Online RL and IL}
\begin{figure}[t]
    \centering
    \includegraphics[width=0.5\columnwidth]{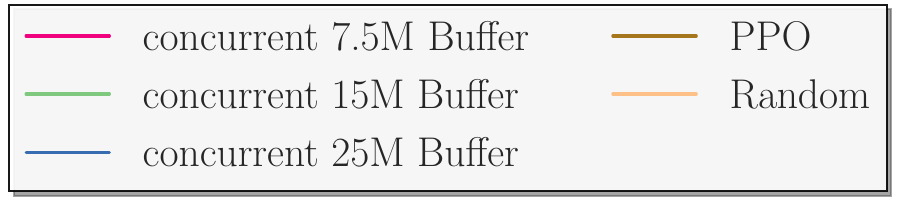}
    \\
    \includegraphics[width=0.9\columnwidth]{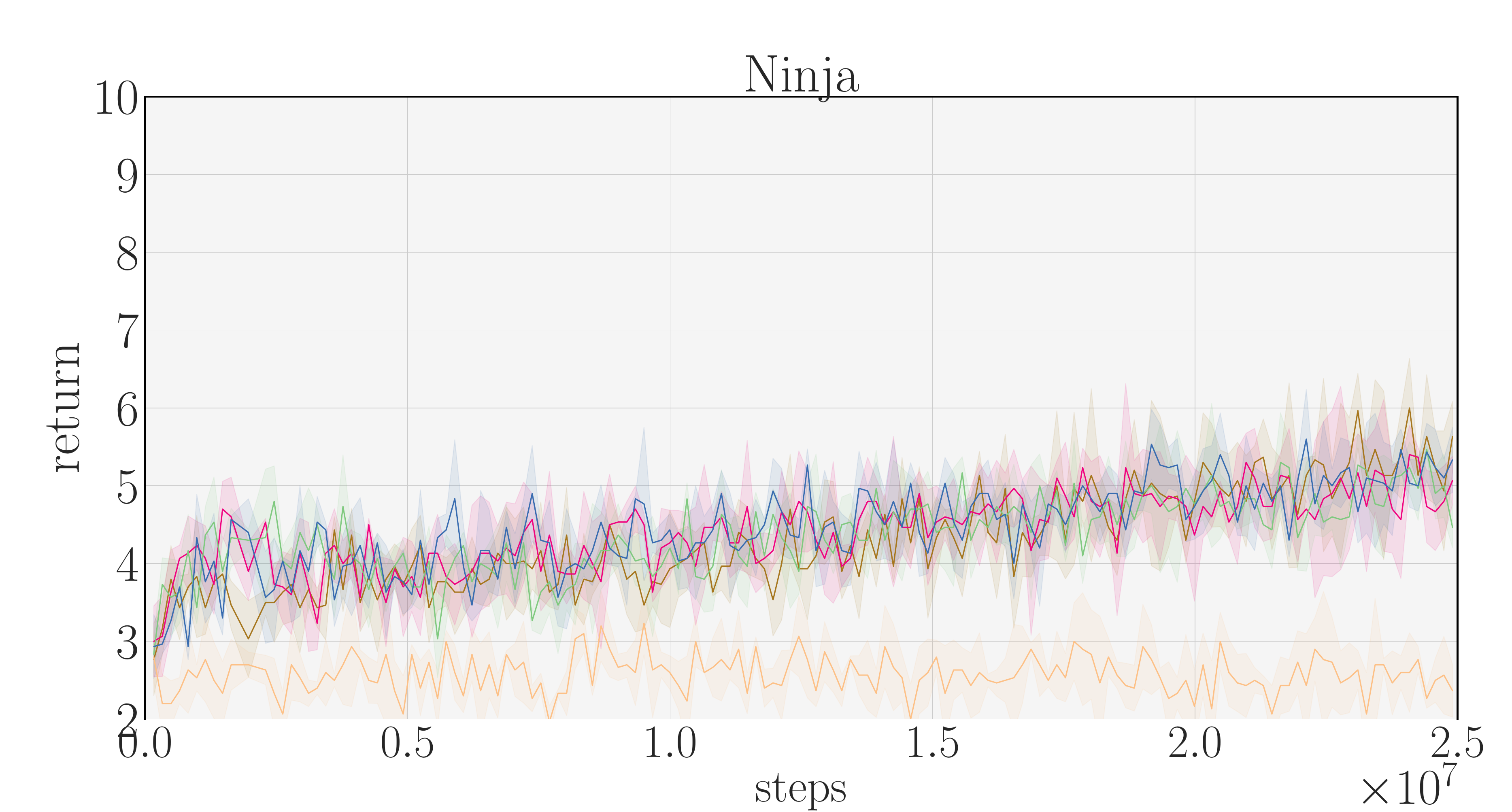}
    \includegraphics[width=0.9\columnwidth]{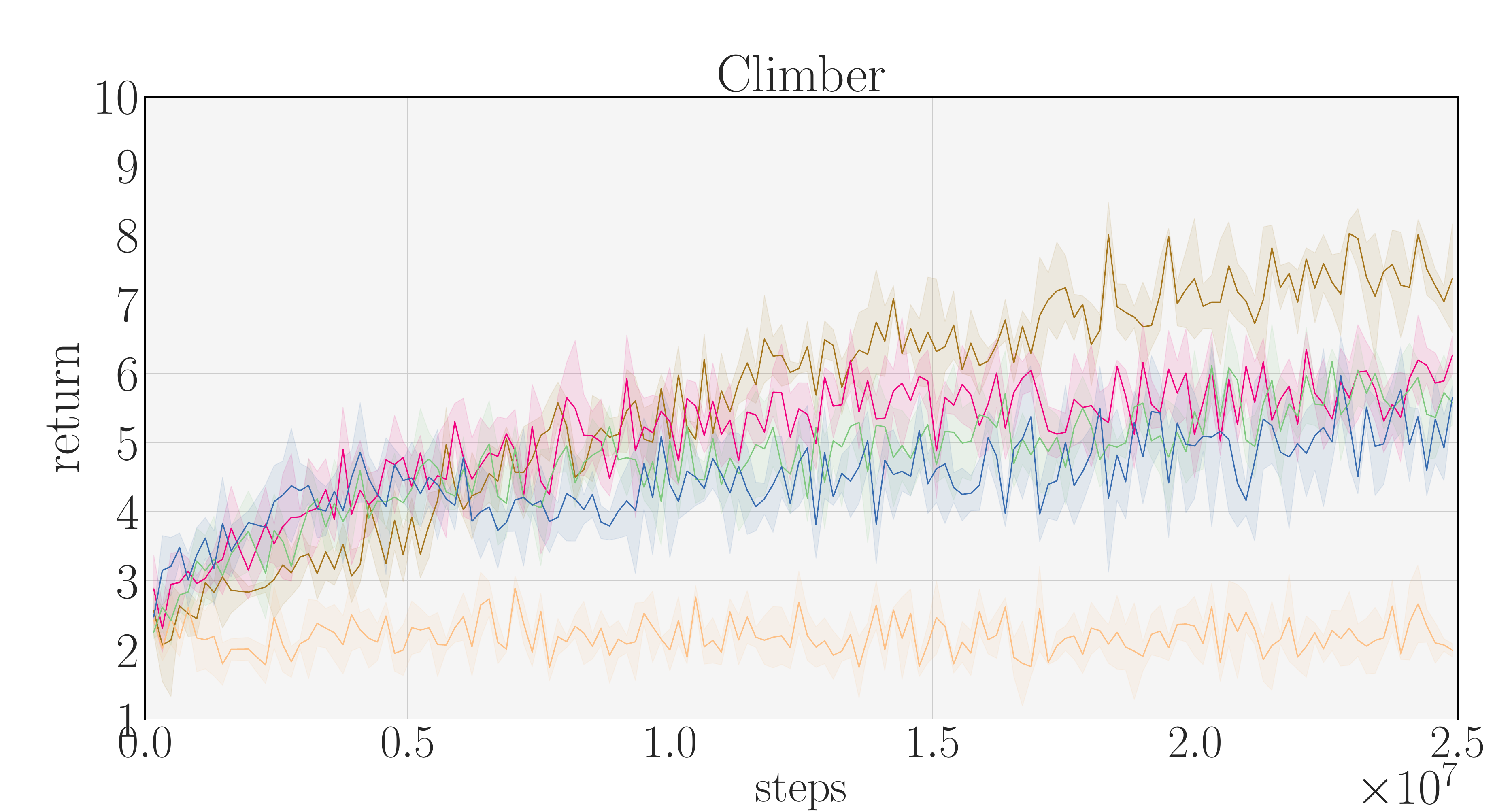}
    \\
    \caption{Performance of the agent when concurrently training the agent with RL and IL in \texttt{Ninja} and \texttt{Climber}.}
    \label{fig:concurrent_training_procgen}
\end{figure}

The insights obtained from Figure \ref{fig:concurrent_training_procgen}
suggest that using IL concurrently with RL does not yield any meaningful differences. Interestingly, at least during pre-training, IL facilitated a performance boost in \texttt{Ninja} when certain buffers were utilized. However, this advantage disappears when IL is applied in concurrent learning, leading to a performance similar to that of the PPO baseline.

\subsubsection{Sensitivity: Diversity of Demonstrations}
\label{subsubsec:procgen_buffer_diversity_1ep}
\begin{figure*}[t]
    \centering
    \resizebox{2\columnwidth}{!}{\begin{tabular}{c|c|c}
        & Ninja & Climber 
        \\
        \rotatebox[origin=l]{90}{Agent selecting random actions} &
        \includegraphics[width=\columnwidth]{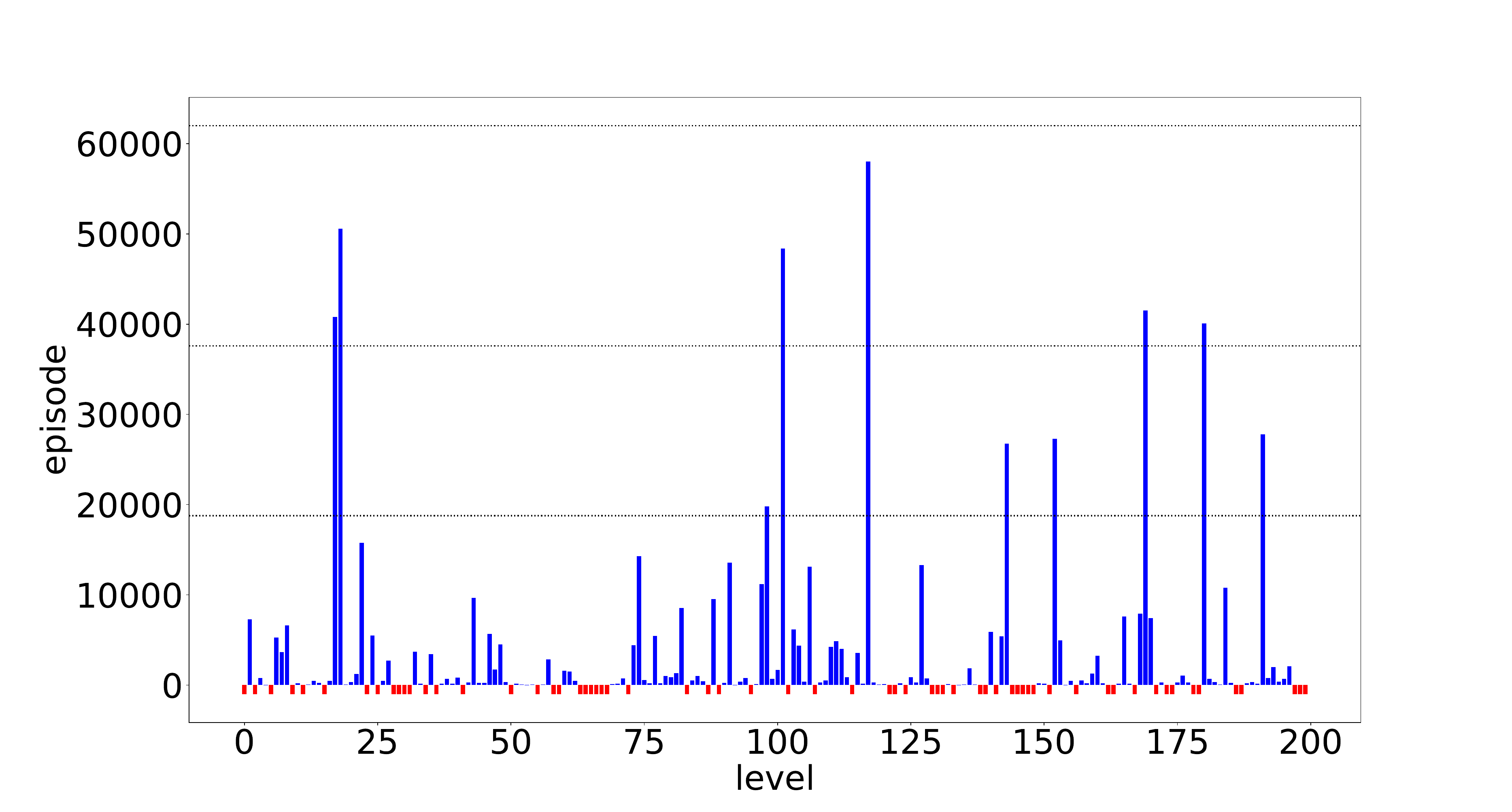} &
        \includegraphics[width=\columnwidth]{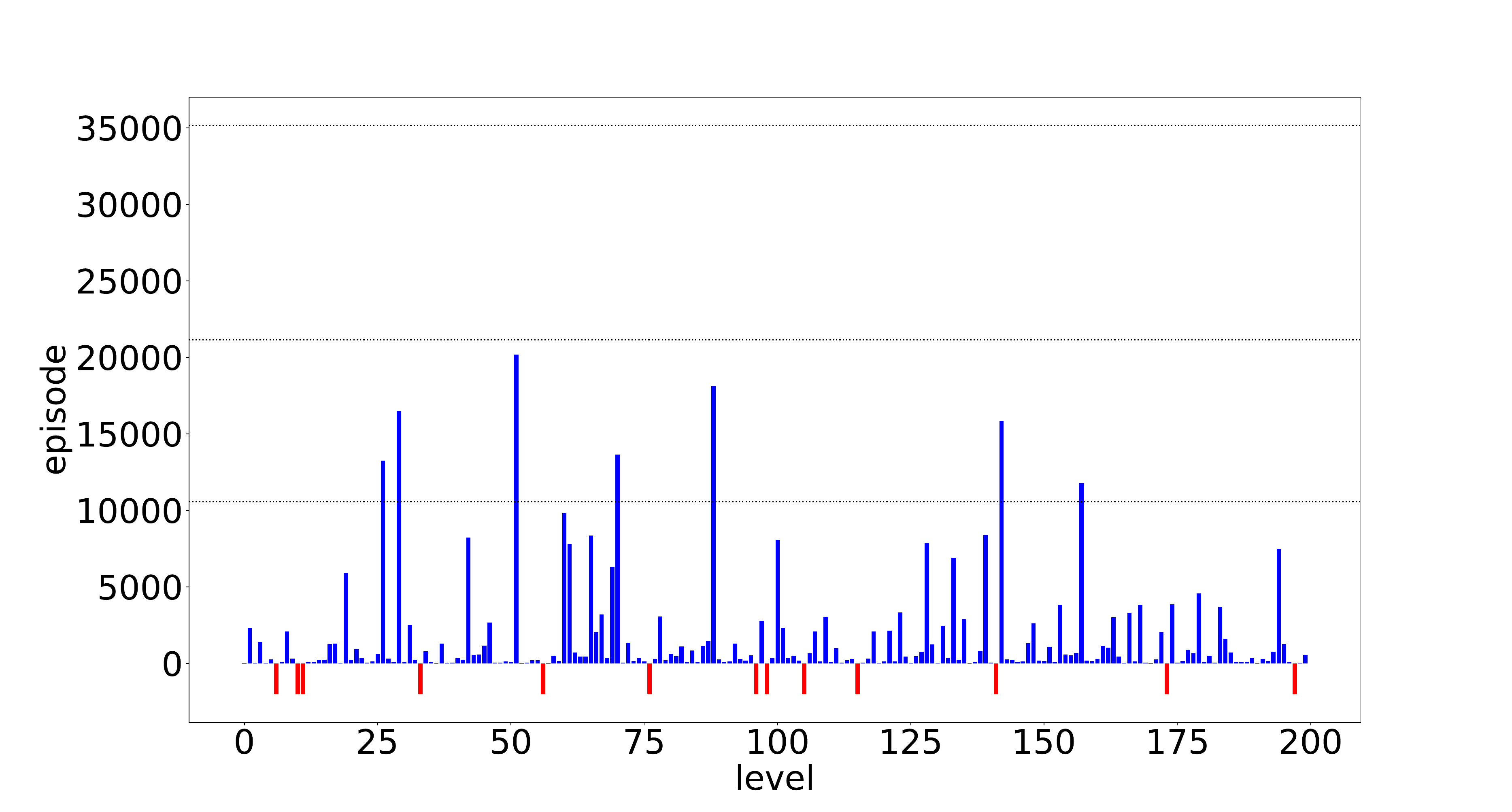}
        \\

        \hline
        \rotatebox[origin=l]{90}{Agent trained solely with PPO} &
        \includegraphics[width=\columnwidth]{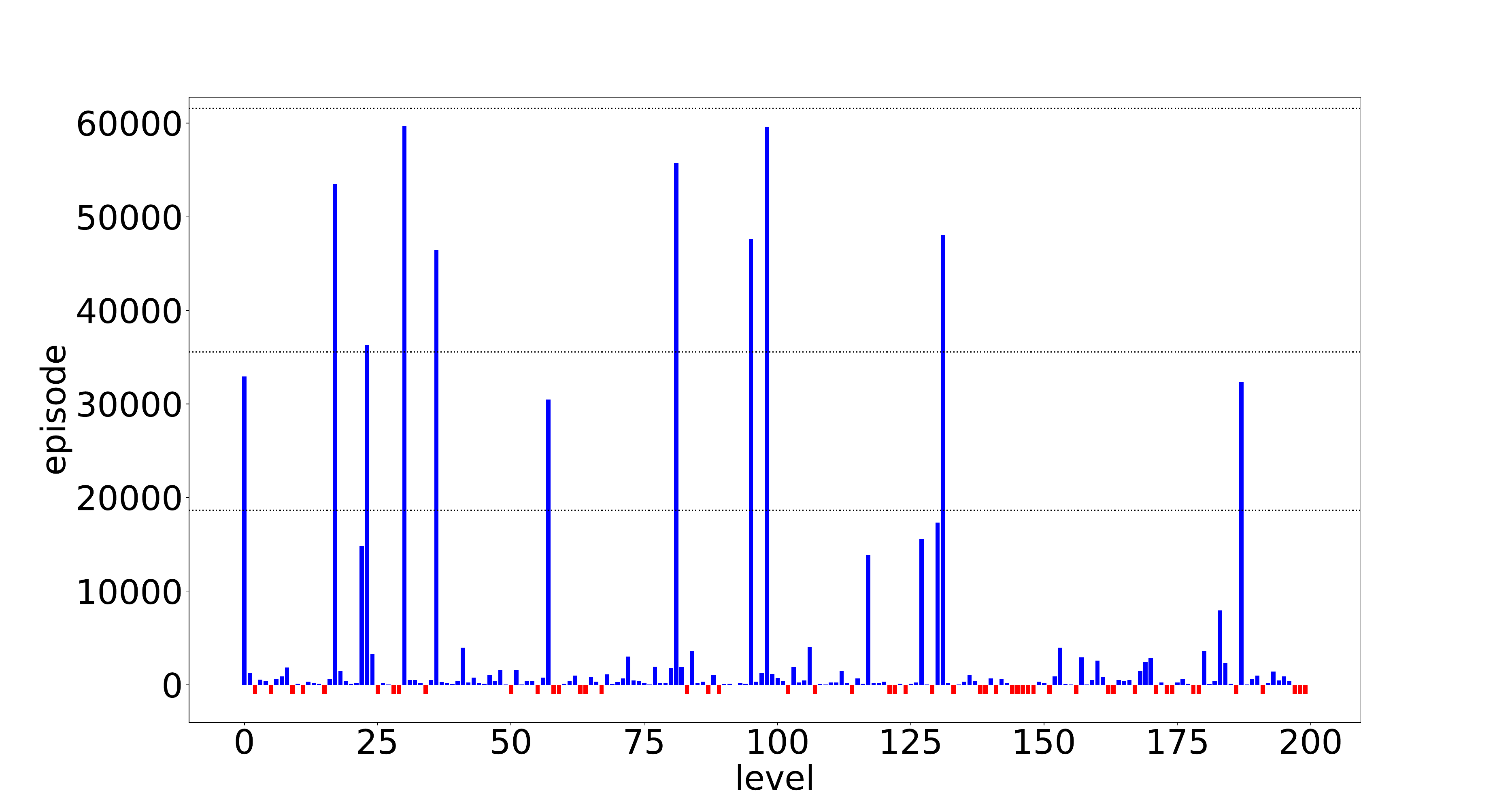} &
        \includegraphics[width=\columnwidth]{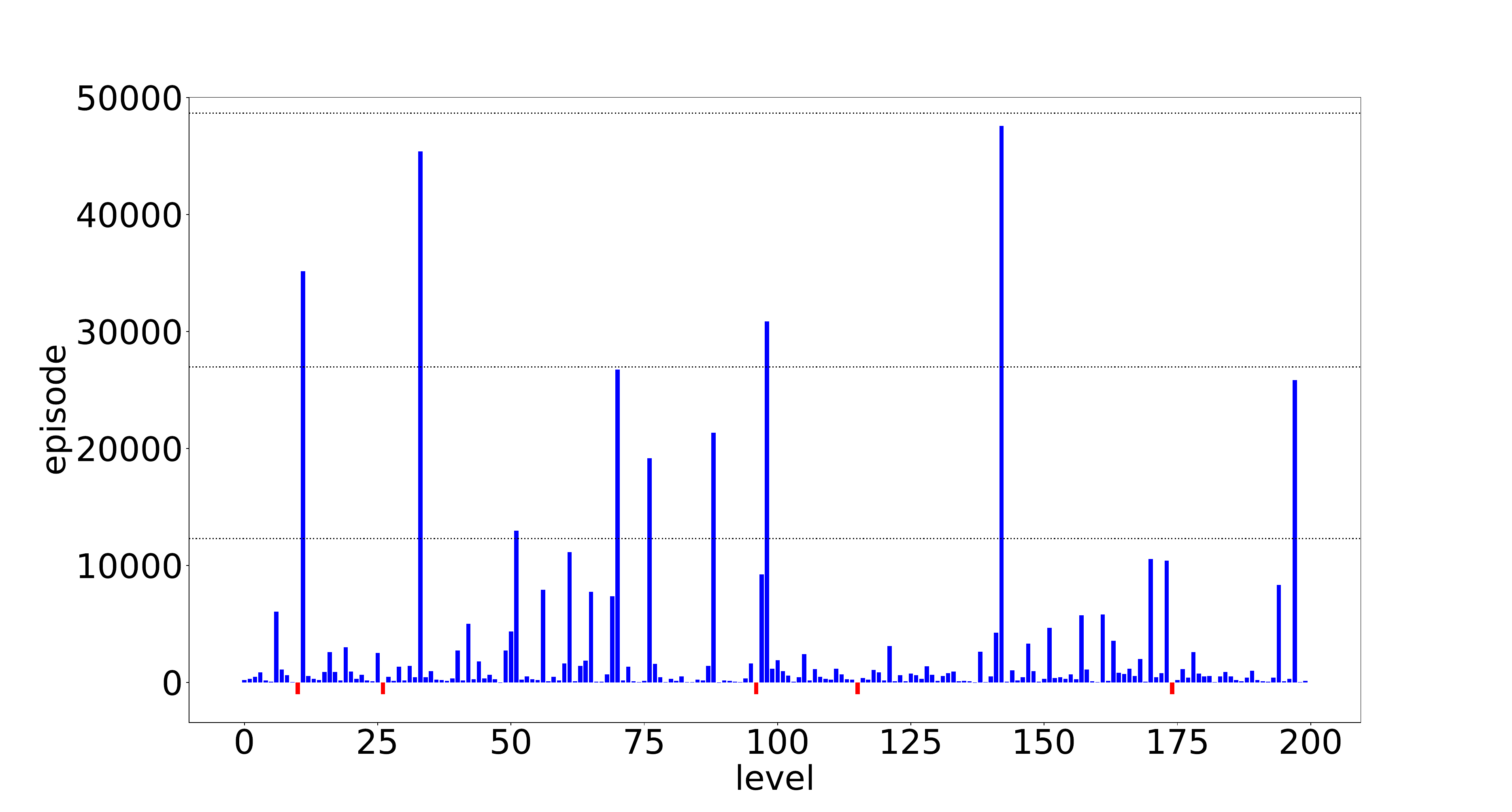}
        \\
        
        \hline
        \rotatebox[origin=l]{90}{Agent trained with Buffer\_1ep} &
        \includegraphics[width=\columnwidth]{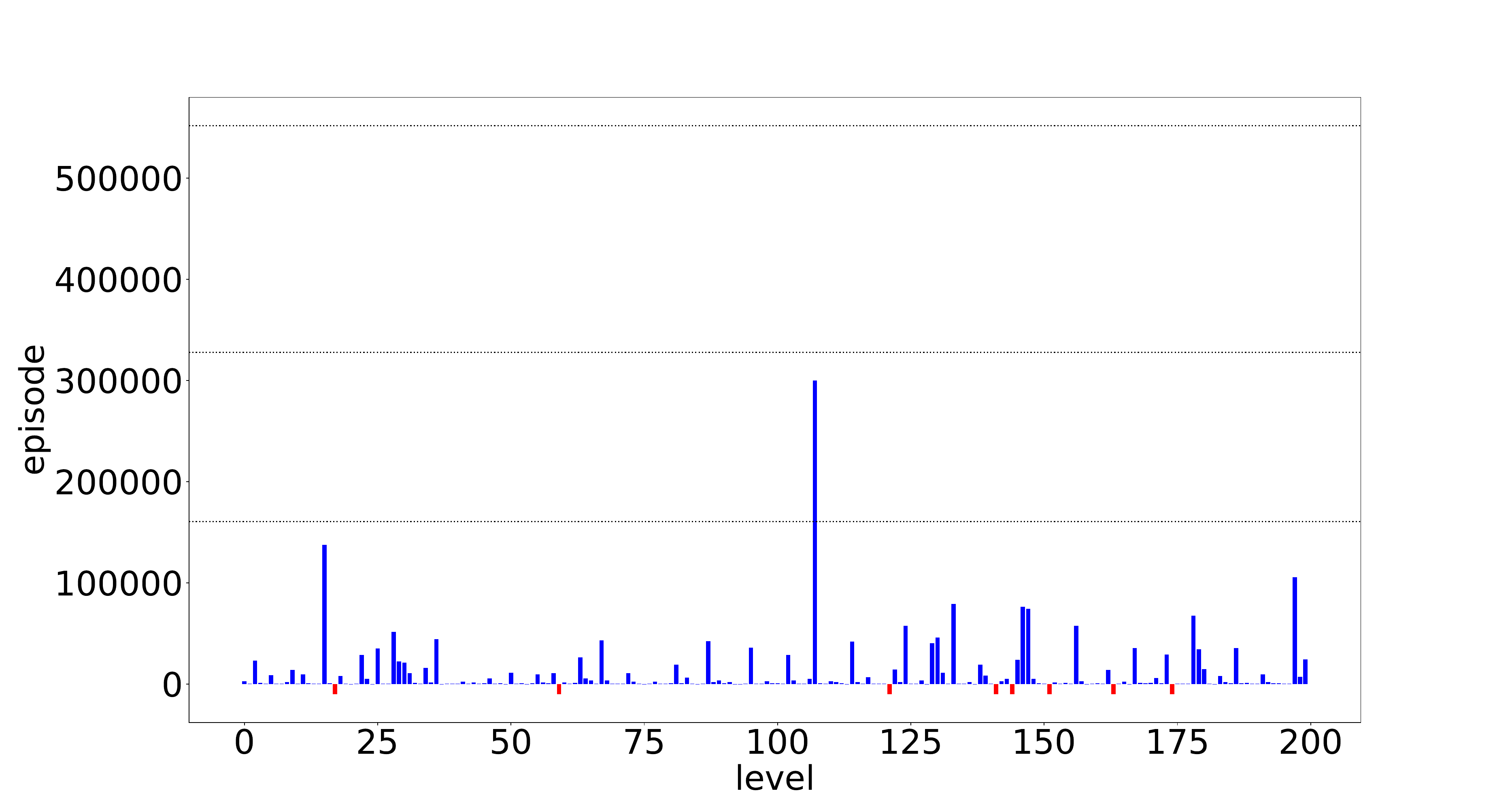} &
        \includegraphics[width=\columnwidth]{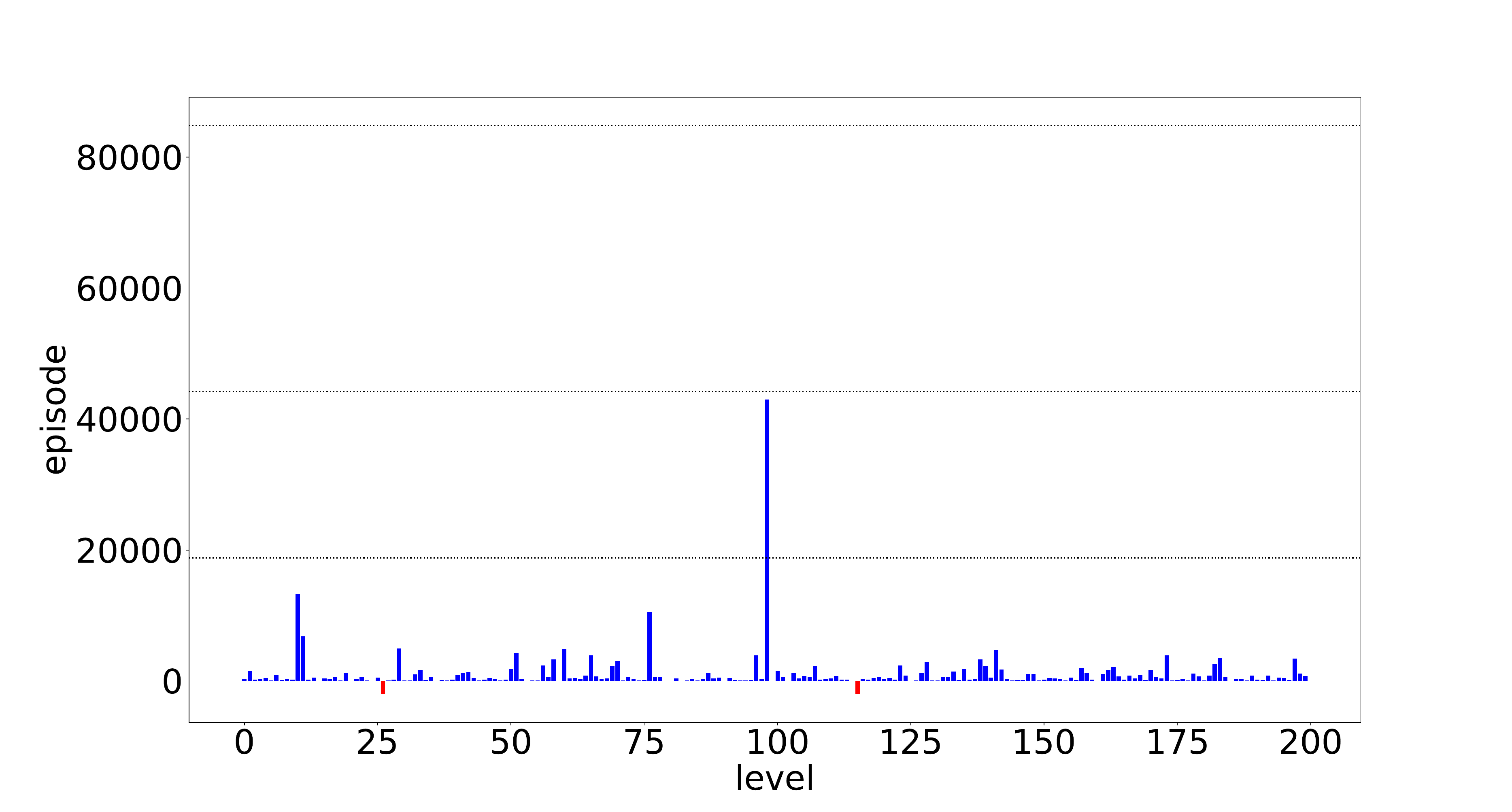}
    \end{tabular}}
    \caption{Graphical representation of the first time a non-zero return is obtained at each of the 200 training levels in \texttt{Ninja} and \texttt{Climber} after 25M steps/interactions. Dashed lines denote milestones at 7.5M, 15M, and 25M time steps. Levels unsolved throughout the training are highlighted in red, indicating the persistent challenge in mastering these levels.
    Randomly selecting actions (top) shows that it is possible to solve 137 levels (68\%) in \texttt{Ninja} and 187 levels (93.5\%) in \texttt{Climber} at least once after 25M steps. In the case of training an agent with PPO during 25M steps (middle), it increases those values to 150 levels (75\%) in \texttt{Ninja} and to 195 (97.5\%) in \texttt{Climber}. 
    In contrast, when training the agent concurrently with RL and IL considering \textit{Buffer\_1ep} (bottom), the number of solved levels after 25M steps is further increased to 192 (96\%) and 198 (99\%) levels for \texttt{Ninja} and \texttt{Climber}, respectively.  
    }
    \label{fig:procgen_succesful_levels}
\end{figure*}

So far we have seen that the potential benefits of IL in Procgen are not as significant as the ones observed in MiniGrid. The effectiveness of IL is closely tied to the demonstrations in use and their relevance in learning the task at hand. We hypothesize that these benefits are diluted due to the inherent properties of Procgen levels, which inadvertently shape the demonstrations. We have identified three key aspects of Procgen levels that may hinder the agent's performance, thereby reducing the expected efficacy of IL in \texttt{Ninja} and \texttt{Climber} tasks:
\begin{itemize}[leftmargin=*]
    \item \textbf{Success Ratio}: The variable difficulty across Procgen levels is significantly greater when compared to MiniGrid. In Procgen, some levels are almost trivial to solve, while others are extremely challenging. This results in a relatively high overall success ratio, as a number of levels are easily solvable
    Such example can be seen in Figure \ref{fig:procgen_succesful_levels}, where a random policy --policy with uniform probability distribution over all actions-- can eventually solve approximately 68\% (\texttt{Ninja}) and 93.5\% (\texttt{Climber}) of the 200 training levels during 25M steps. In contrast, the likelihood of a random policy to solve any level in MiniGrid is close to 0\%. We refer to \ref{app:success_levels_training} for more evidence supporting these observations.

    \item \textbf{Goodness}: The quality of an executed trajectory is evaluated using the reward function $\mathcal{R}$. However, if the reward function does not capture differences between trajectories belonging to different levels, then it is not suitable for the purpose \cite{andres_towards_2022}. This actually happens in both MiniGrid and Procgen tasks. In the Procgen tasks of \texttt{Ninja} and \texttt{Climber}, however, the challenge is even more significant. Here, regardless of the number of steps taken (i.e., the length of the trajectory), the same reward is given. This uniformity in the reward assignment hinders the evaluation of the quality of individual trajectories, a problem that is not present in MiniGrid. As a result, the difficulty in distinguishing between trajectories is higher in Procgen.

    \item \textbf{Similarity}: Measuring the similarity between levels is complex, distinct from simply assessing the success ratio of solving them. This issue, often encountered in multi-task problems\footnote{We can imagine each level at the selected Procgen task as a subtask. For instance, considering \texttt{Ninja}, we could consider the 200 levels as subtasks, where our goal would be to obtain a \textit{multi-task} policy that is able to generalize across that entire distribution of levels (subtasks).}, is crucial for understanding the transferability of learning from one level to others. While a certain level might be easy to solve, indicating a high success ratio, this does not automatically imply that the strategies learned there will be effective for different levels. 
    In other words, similarity between tasks allows discerning the utility of learning how to complete a level and its impact in the learning of the rest of the levels. Although we have not computed a specific metric, our empirical observations indicate a notable variance in the similarity of levels. Some levels show more prevalent patterns or strategies for solving them than others. 
\end{itemize}

Independently from the data collection process, even if human experts where in charge of gathering demonstrations, the issue of \textbf{similarity} highlighted earlier will still be present. This situation leads to an imbalance in the learning process because different levels exhibit varying degrees of similarity, necessitating the development of distinct strategies for each level. Thus, even with expertly gathered demonstrations, the challenge of adapting to this diversity in similarity and designing a strategy accordingly remains.

The problem is further exacerbated when considering our \textit{data collection} methodology (Section \ref{sec:data_collection}), which prioritizes storing levels with high return scores. Consequently, the buffer content is filled with demonstrations that have high \textbf{success ratio} and high return values (associated with \textbf{goodness}), which do not necessarily represent demonstrations that can accelerate the learning process of the agent. This is clearly exposed in Table \ref{tab:buffer_metada_procgen}, where the content for \texttt{Climber} is skewed towards just 2-4 levels that range with return scores between 15 and 17. Similarly, in \texttt{Ninja}, all successful trajectories --regardless of their complexity or length-- yield the same +10 return. Therefore, the buffer content is dominated by demonstrations of levels that are easily solved (i.e., with high success ratio).

As a result, the collected buffers become dominated by demonstrations that are either biased towards spurious features represented in small subset of levels, or towards levels that are easily solvable. \textbf{These skewed demonstrations significantly hamper the learning of more complex levels, thus undermining the potential benefits of using IL}.

\subsubsection{Ensuring Diversity in the Replay Buffer}

In order to address the aforementioned issues, we draw inspiration from the findings of a concurrent study~\cite{andres_enhanced_2023} and actively enforce the diversity of trajectories in the buffer. Specifically, we modify the data collection strategy so that the buffer stores one unique trajectory per level, rather than selectively storing only the best trajectories regardless of the level they belong to.
By implementing this modification, referred as \textit{Buffer\_1ep},  we aim to eliminate the bias towards levels that are either too easily solved (like those observed in \texttt{Ninja}) or that have outlier return scores (as in \texttt{Climber}). This ensures a more balanced and comprehensive representation of all levels in the training set. 

\paragraph{Concurrent Training}
\begin{figure}[t]
    \centering
    \includegraphics[width=\columnwidth]{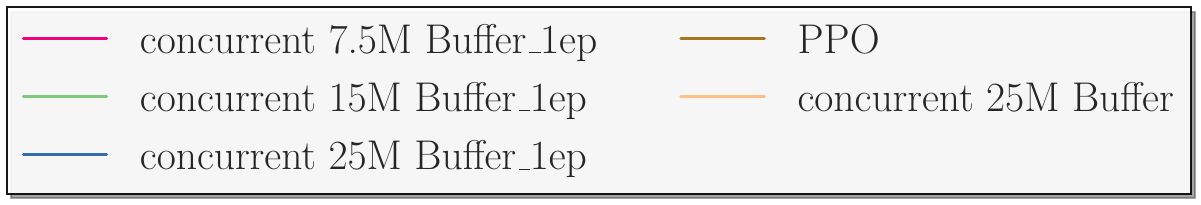}
    \\
    \includegraphics[width=0.9\columnwidth]{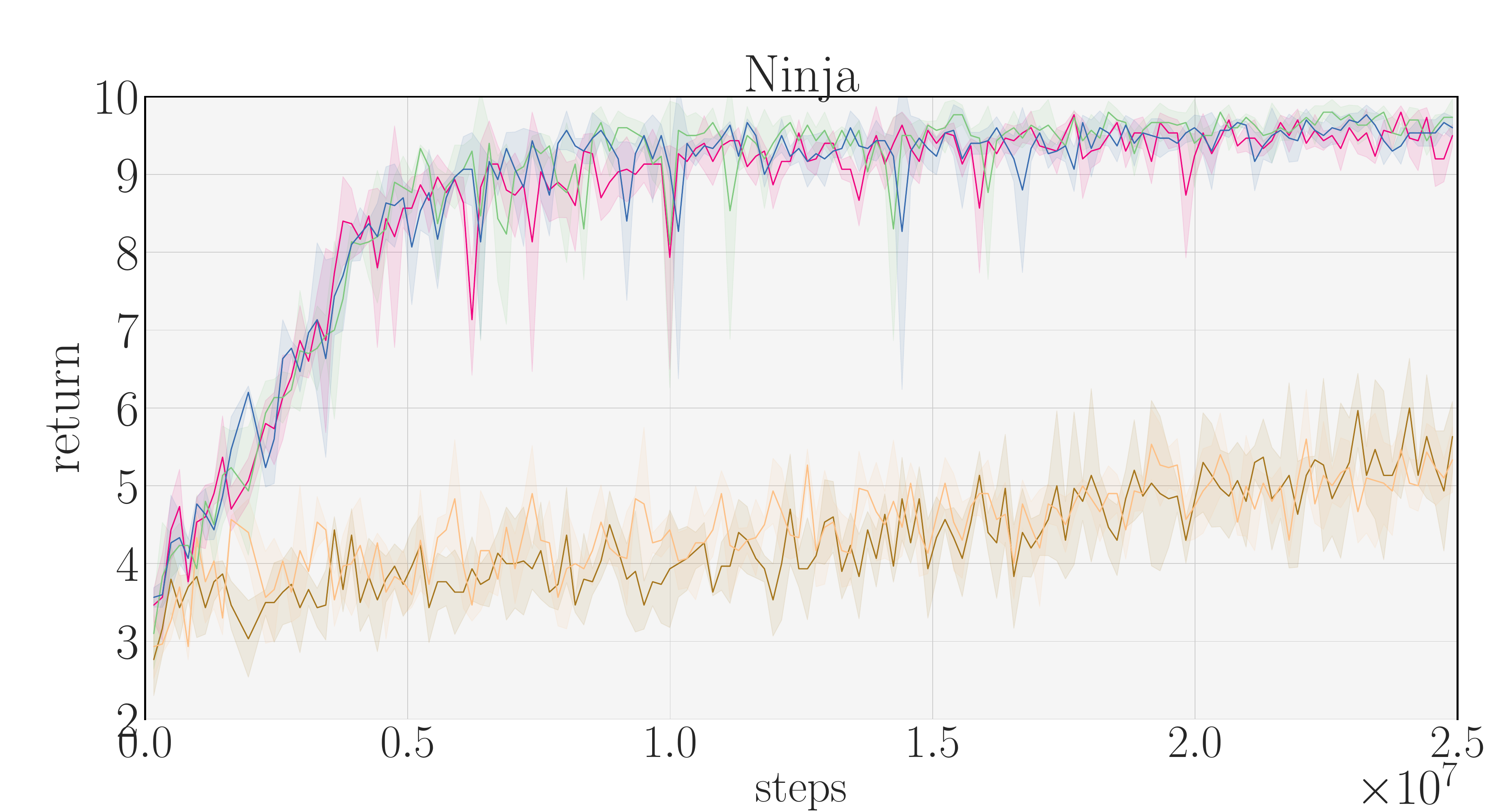}  
    \\
    \includegraphics[width=0.9\columnwidth]{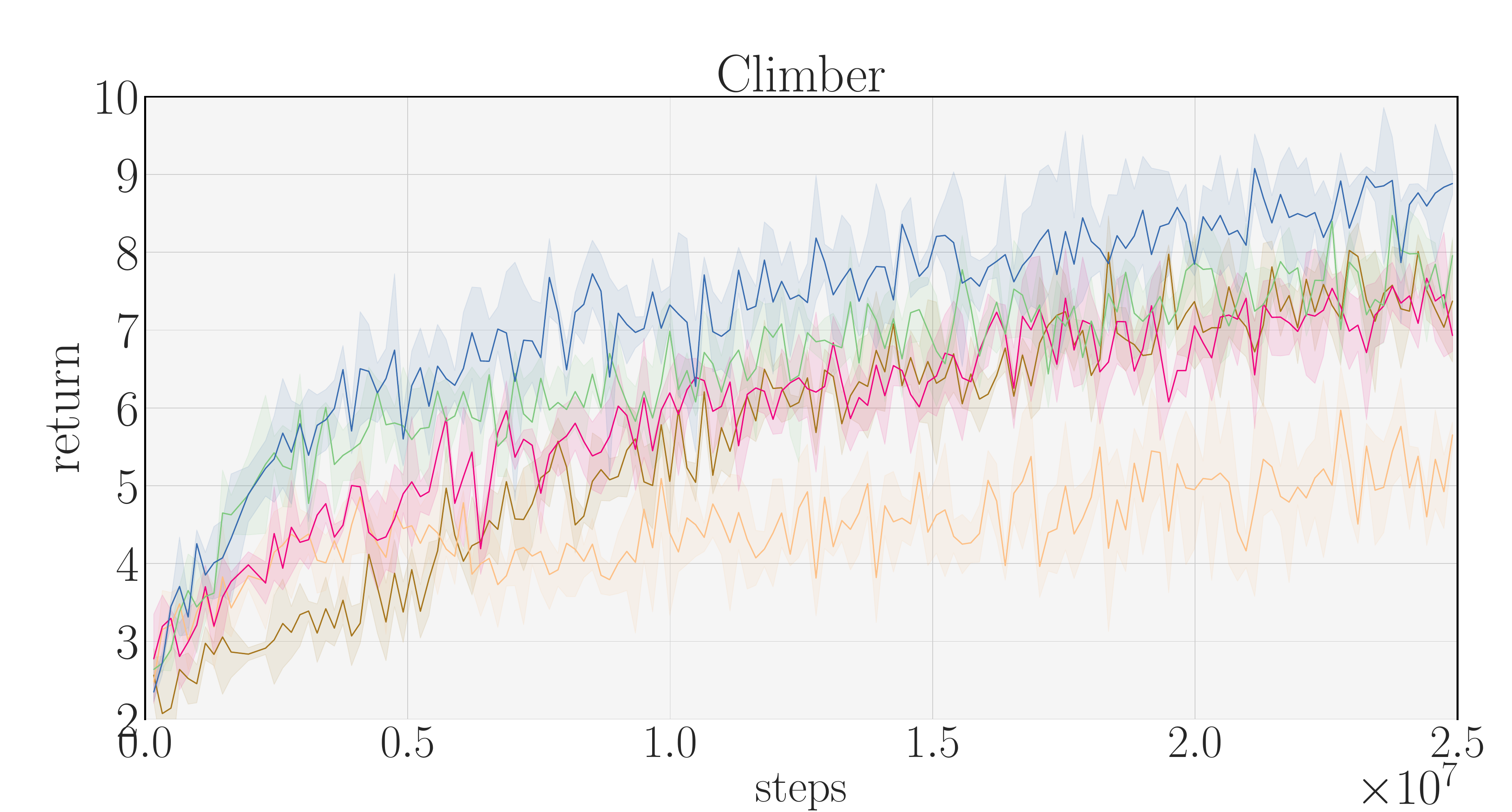}
    \caption{Train performance of the agent when concurrently training the agent with RL and IL with demonstrations stored at a buffer with one trajectory per level (i.e., \textit{Buffer\_1ep})
    in Procgen's \texttt{Ninja} and \texttt{Climber} tasks.}
    \label{fig:procgen_1ep}
\end{figure}

Figure \ref{fig:procgen_succesful_levels} shows that enforcing the diversity throughout training with \textit{Buffer\_1ep} enables the agent to solve levels that were previously intractable. Consequently, the agent increases its sample efficiency and converged return in both \texttt{Ninja} and \texttt{Climber} tasks, as reflected by the training curves plotted in Figure \ref{fig:procgen_1ep}. The key to this enhanced performance lies in the uniform diversity promoted across training levels. We hypothesize that \textit{Buffer\_1ep} enables the agent to learn a more robust representation that better matches the generalization requirements exhibited in the entire level distribution. In turn, this allows the agent to have higher success probability on solving other levels that were not solved before. For instance, an agent trained solely with PPO was incapable of solving 50 levels in \texttt{Ninja}, and 5 levels in \texttt{Climber}. However, when training the agent concurrently with RL and IL with the \textit{25M Buffer\_1ep}, the resulting policies manage to reduce the number of unsolved levels to 8 and 2 levels for \texttt{Ninja} and \texttt{Climber}, respectively\footnote{\ref{app:success_levels_training} provides further information regarding the levels that the agent successfully solved at least once during its training.}.

A closer inspection of the results in Figure \ref{fig:procgen_1ep} reveals a more pronounced performance improvement in \texttt{Ninja} compared to \texttt{Climber}. Interestingly, the only configuration that outperforms the baseline in \texttt{Climber} is the \textit{25M Buffer\_1ep}. This suggests that the magnitude of improvement in each task correlates with the reduction in the number of unsolved levels. The substantial difference in terms of unsolved levels in \texttt{Ninja} compared to \texttt{Climber} highlights a greater potential for improvement in the former.

\paragraph{Pre-training + Concurrent RL and IL}

\begin{figure}[h!]
    \centering
    \includegraphics[width=\columnwidth]{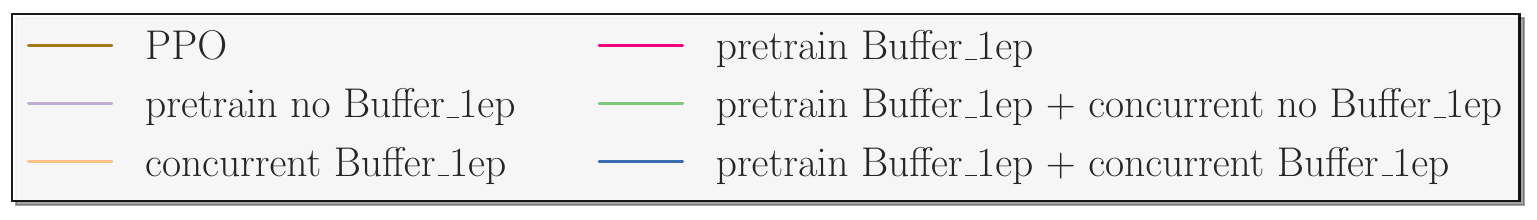}
    \\
    \includegraphics[width=0.9\columnwidth]{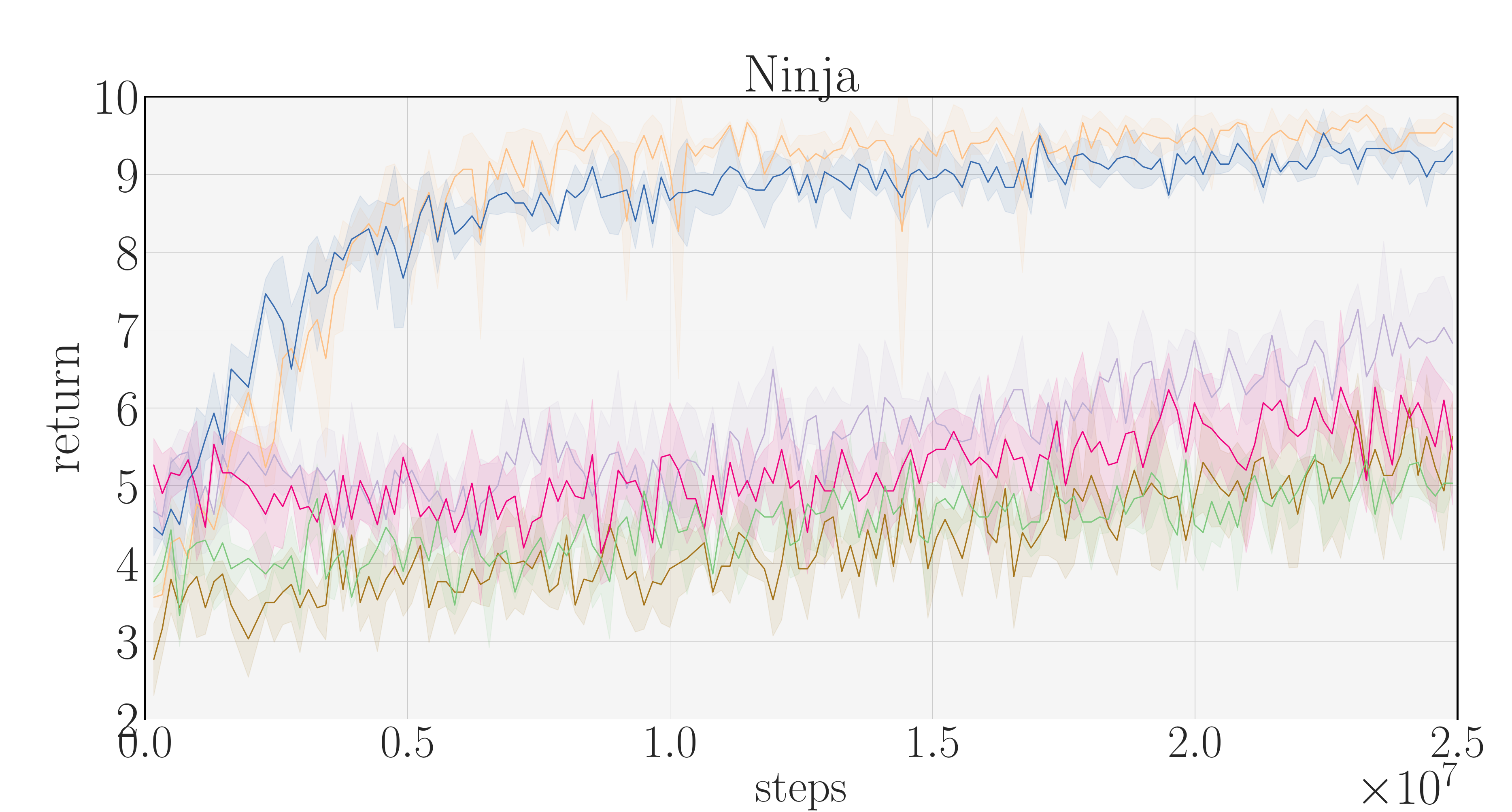} 
    \\
    \includegraphics[width=0.9\columnwidth]{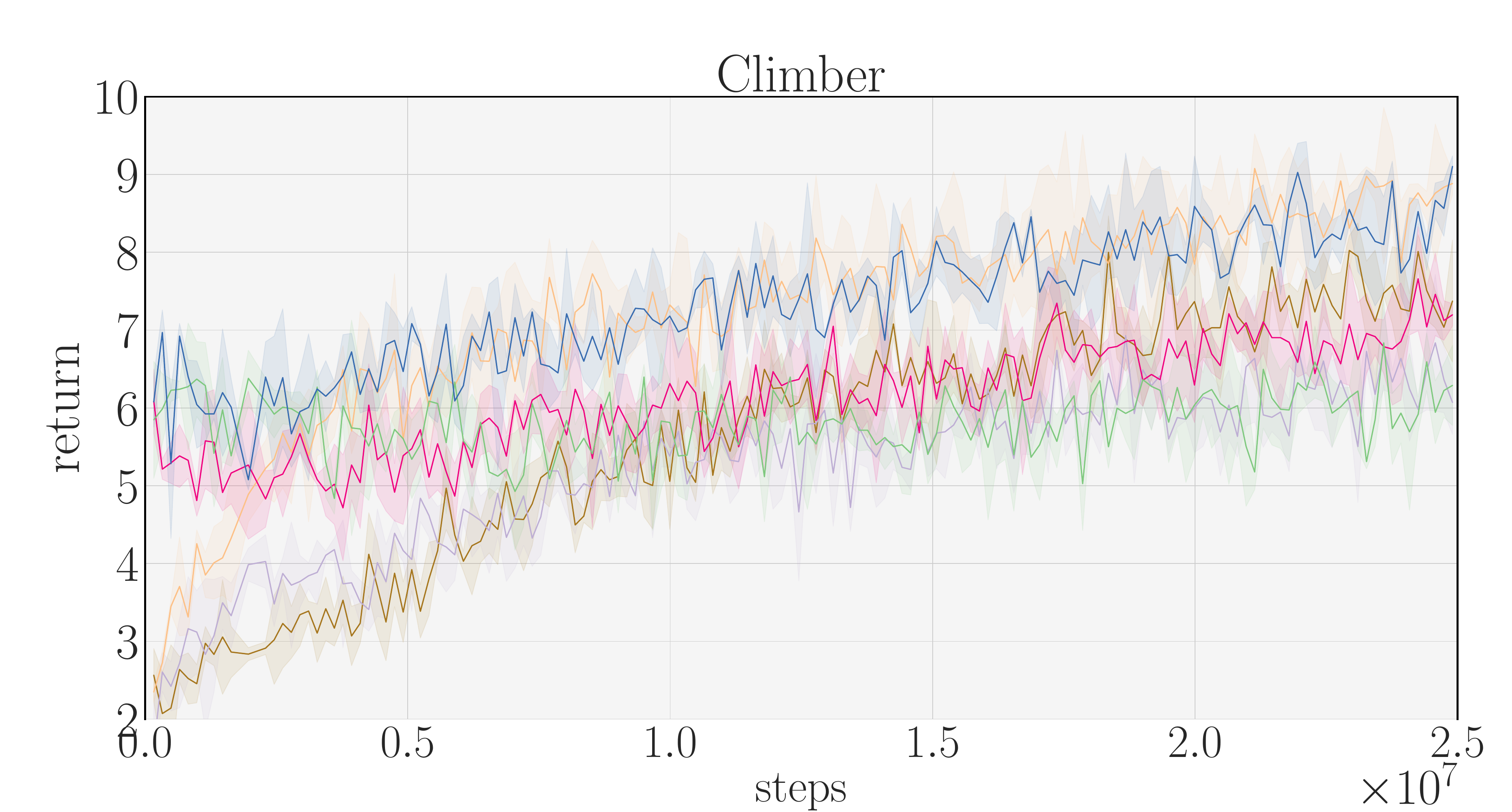} 
    \caption{Impact of using IL at pre-training in Procgen's \texttt{Ninja} and \texttt{Climber} tasks with different buffers when considering at least one trajectory per level. We also plot what happens if we keep using IL during the online training phase concurrently with RL.}
    \label{fig:procgen_pretrain_online_1ep}
\end{figure}

Figure \ref{fig:procgen_pretrain_online_1ep} reveals a discernible jumpstart advantage in the initial stages of learning when pre-training is employed. Indeed, the agent experiences a tangible jumpstart if it undergoes pre-training with \textit{Buffer\_1ep} (as indicated by the pink, green, and blue curves). However, this initial advantage tends to reach a plateau in the absence of further IL updates during concurrent training, underscoring the limitations of pre-training as an isolated method. More crucially, the benefits of concurrent IL are diminished if the buffer does not maintain a trajectory-per-level approach during online interactions (see difference between green and blue curves). Therefore, the most effective results are achieved when using \textit{Buffer\_1ep} in pre-training and concurrent learning phases. Nevertheless, it is worth noting that the concurrent learning strategy, even without pre-training, manages to achieve high returns and sample efficiency (orange curve), matching the initial jumpstart of pre-training. Thus, employing just concurrent learning with \textit{Buffer\_1ep} proves to be a sufficient and efficient training approach. 

These outcomes differ from those obtained in MiniGrid, where pre-training enable agents to learn levels they would have never solved otherwise. Our hypothesis centers on two key factors: the \textbf{success ratio} and the \textbf{similarity} between levels. 
In Procgen, even using \textit{Buffer\_1ep}, the buffer exhibits an imbalance; levels with a \textbf{high success ratio}, which are easier to solve, are disproportionately represented compared to more challenging levels. This leads to the agent mastering levels it would likely have learned quickly anyway during initial online interactions. Furthermore, owing to the \textbf{low similarity} between these easily solvable levels and others, the benefit of this mastery in terms of transferability to different levels is minimal. Consequently, due to these factors, the overall benefits of pre-training are significantly reduced.

\paragraph{Trajectory Quality Assessment} While the \textit{Buffer\_1ep} strategy leads to improved results by enhancing the diversity of experiences in the buffer, the quality of these demonstrations remains unclear\footnote{The reward function $\mathcal{R}$ in \texttt{Ninja} and \texttt{Climber} does not effectively reflect if one trajectory is better than another. We recall the \textbf{goodness} property previously explained.} 
To determine whether the observed gains due to the induced diversity can also be increased (or decreased) with higher (or lower) quality data, we introduce two additional experimental conditions under the concurrent RL and IL paradigm:
\begin{itemize}[leftmargin=*]
    \item \textit{Buffer\_1ep\_random}: In this approach we maintain the constraint of one trajectory per level in the buffer. However, instead of collecting data during the training of an agent with RAPID, we gather trajectories using randomly selected actions over a total of 25M steps. This approach will help ascertain the value of diversity when the data quality is not necessarily high.
    \item \textit{Buffer\_1ep\_higherQuality}: After finishing the training of an agent concurrently with RL and IL using data belonging to \textit{25M Buffer\_1ep}, this should result in a new buffer with potentially higher-quality data from the outset. This upgraded buffer, referred to as \textit{Buffer\_1ep\_higherQuality}, is used to assess whether starting with higher-quality data can translate to further improvements in the agent's performance.
\end{itemize}

\begin{figure}[t]
    \centering
    \includegraphics[width=\columnwidth]{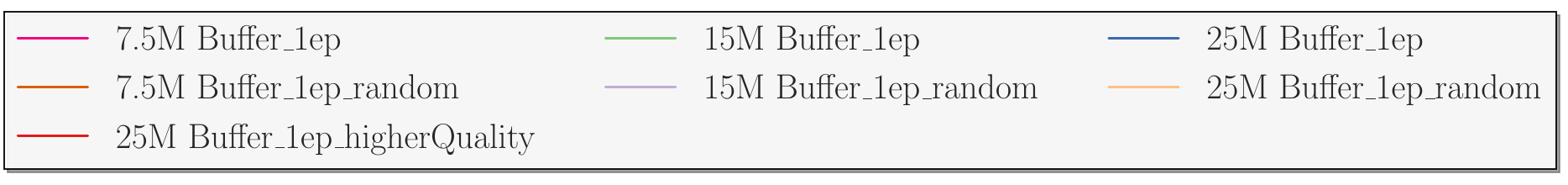}
    \\
    \includegraphics[width=0.9\columnwidth]{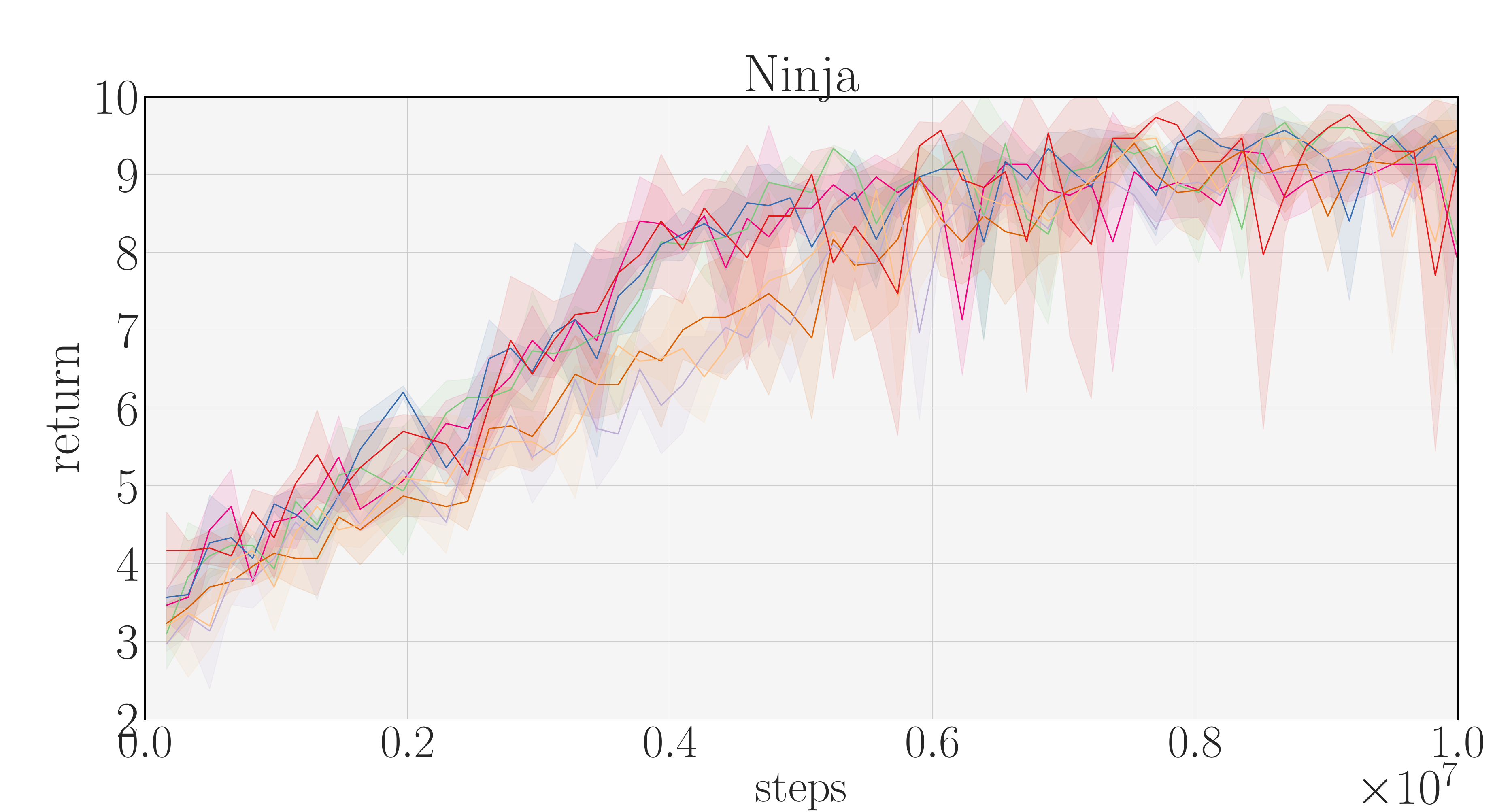} 
    \includegraphics[width=0.9\columnwidth]{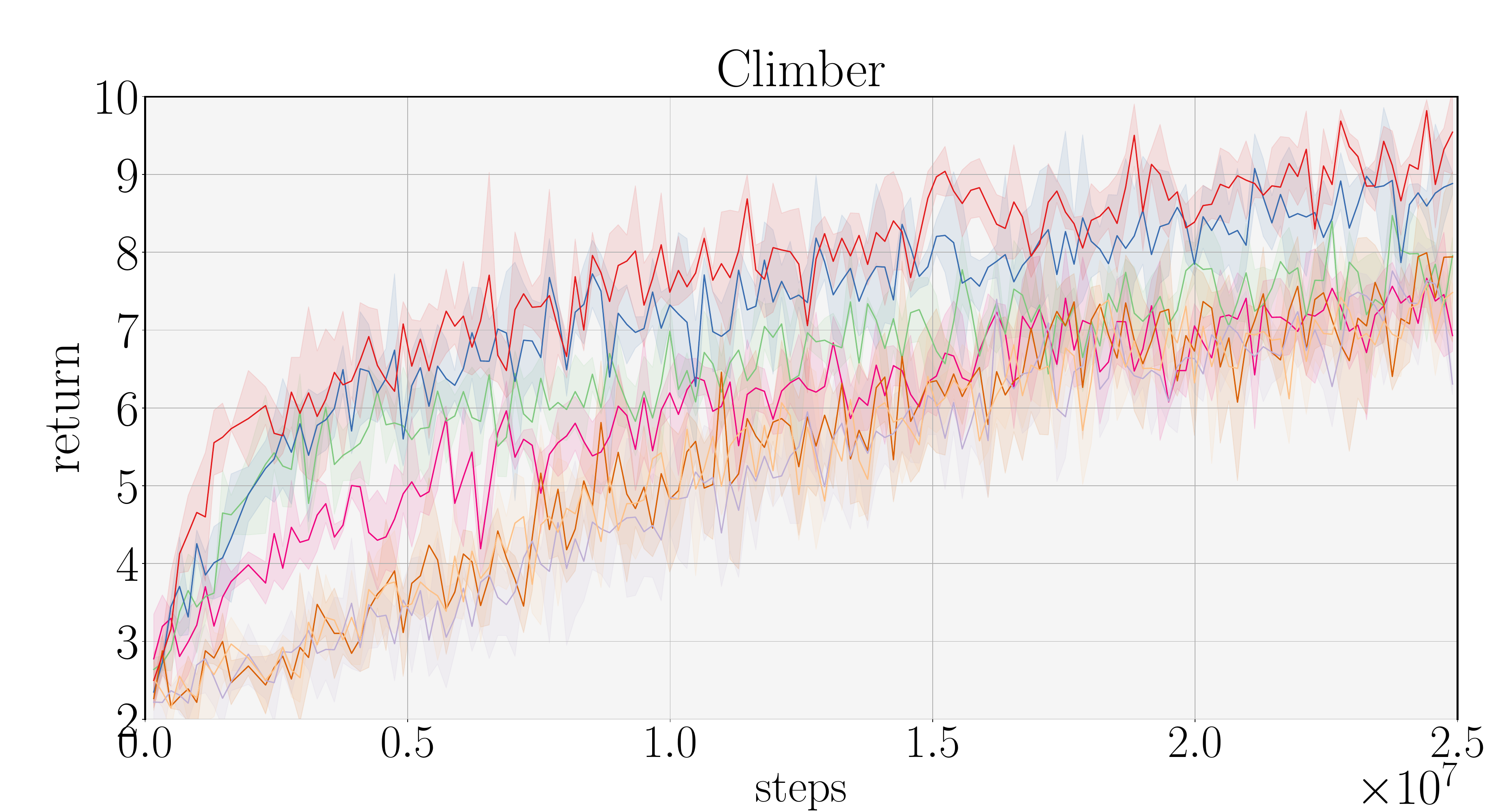} 
    
    \caption{
    Comparative performance analysis of agents trained concurrently with RL and IL using demonstrations from buffers collected via distinct methods, which presumptuously reflect different quality of data: \textit{Buffer\_1ep\_random} with the lowest expected quality, \textit{Buffer\_1ep} with moderate quality, and \textit{Buffer\_1ep\_higherQuality} representing the highest quality demonstrations.
    }
    \label{fig:procgen_1ep_normal_random_expert}
\end{figure}

Surprisingly, Figure \ref{fig:procgen_1ep_normal_random_expert} shows that even lower-quali\-ty data, as reflected in \textit{Buffer\_1ep\_random}, can lead to significantly better sample efficiency in \texttt{Ninja} compared to the PPO baseline. However, this advantage is less pronounced in \texttt{Climber}. In contrast, using higher-quality data from \textit{25M Buffer\_1ep\_higherQuality} yields little to none improvement in \texttt{Ninja} with respect to any solution using \textit{Buffer\_1ep}. Conversely, in \texttt{Climber}, the best results --actually the only ones surpass the PPO baseline-- are reported with \textit{25M Buffer\_1ep} and \textit{25M Buffer\_1ep\_higherQuality}. Nevertheless, we believe that such big differences between \texttt{Ninja} and \texttt{Climber} are more related to an imbalance derived by the \textbf{similarity} between their levels. In \texttt{Climber}, after finishing the training with high-quality demonstrations, the agent still has an $\sim$ 85\% success ratio in all the levels even if it has 99\% of successful demonstration examples in the buffer (see \ref{app:success_levels_training}, Figure \ref{fig:procgen_succesful_levels_extended}). On the contrary, in \texttt{Ninja} the agent achieves a $\sim 90\%$ of success with such high-quality data, although it has a lower 96-97\% of successful demonstration examples. However, that $\sim 90\%$ is consistent disregarding the quality of data we collect the demonstrations from.

This reflects that mastering certain levels in \texttt{Climber} is more difficult than in \texttt{Ninja}, even if valid demonstrations are available in both cases. Moreover, if those levels that are complex to master have little representation in the buffer (i.e., an imbalanced buffer content), then learning the required strategy from them becomes even more complicated.

\vspace{1mm}
In order to strengthen our hypothesis, we bring the reader's attention to the y-axis in Figure \ref{fig:procgen_succesful_levels}, which shows the number of episodes experienced by the agent during 25M training steps. In the case of training the agent with \textit{Buffer\_1ep}, we can see that in \texttt{Ninja} almost 500,000 episodes are used, in contrast to the 80,000 in \texttt{Climber}. This suggests that the agent has learned to master a large proportion of the levels faster in \texttt{Ninja}, which can be due to the \textbf{similarity} between levels to be higher than in \texttt{Climber}. However, this may also occur because, on average, \texttt{Climber} levels require more episodes (i.e., interactions to discover a valid strategy) or demonstrations of higher quality to be learned.

In summary, in some cases learning with IL from diverse data of low quality, such as collected by an agent selecting actions randomly, can significantly improve sample efficiency. However, in other tasks, using higher-quality of data can be necessary. Moreover, evaluating the similarity between tasks can be helpful to see if there exists any imbalance in the strategies to be mastered by the agent.

\subsubsection{Generalization: Train-Test Distributions}

In Section \ref{subsec:generalization} we demonstrated that, given a set number of training levels, the return achieved during training aligns well with the expected evaluation return on levels that have never been seen before. In MiniGrid, we utilized 10,000 training levels, which were proven to be sufficient to ensure generalization on unseen episodes. However, this claim did not hold in Procgen, where training the agent on only 200 levels resulted in limited generalization capabilities. To further investigate this, we analyzed the agent's performance on both training (solid lines) and testing (dashed lines) levels for Procgen's \texttt{Ninja} and \texttt{Climber} tasks for 1,000 training levels.

\begin{figure}[t]
    \centering
    \includegraphics[width=\columnwidth]{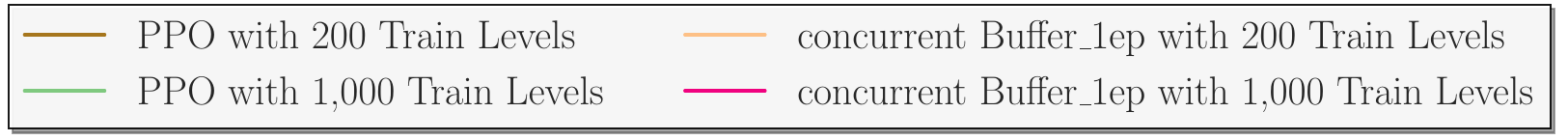}
    \\
    \includegraphics[width=0.9\columnwidth]{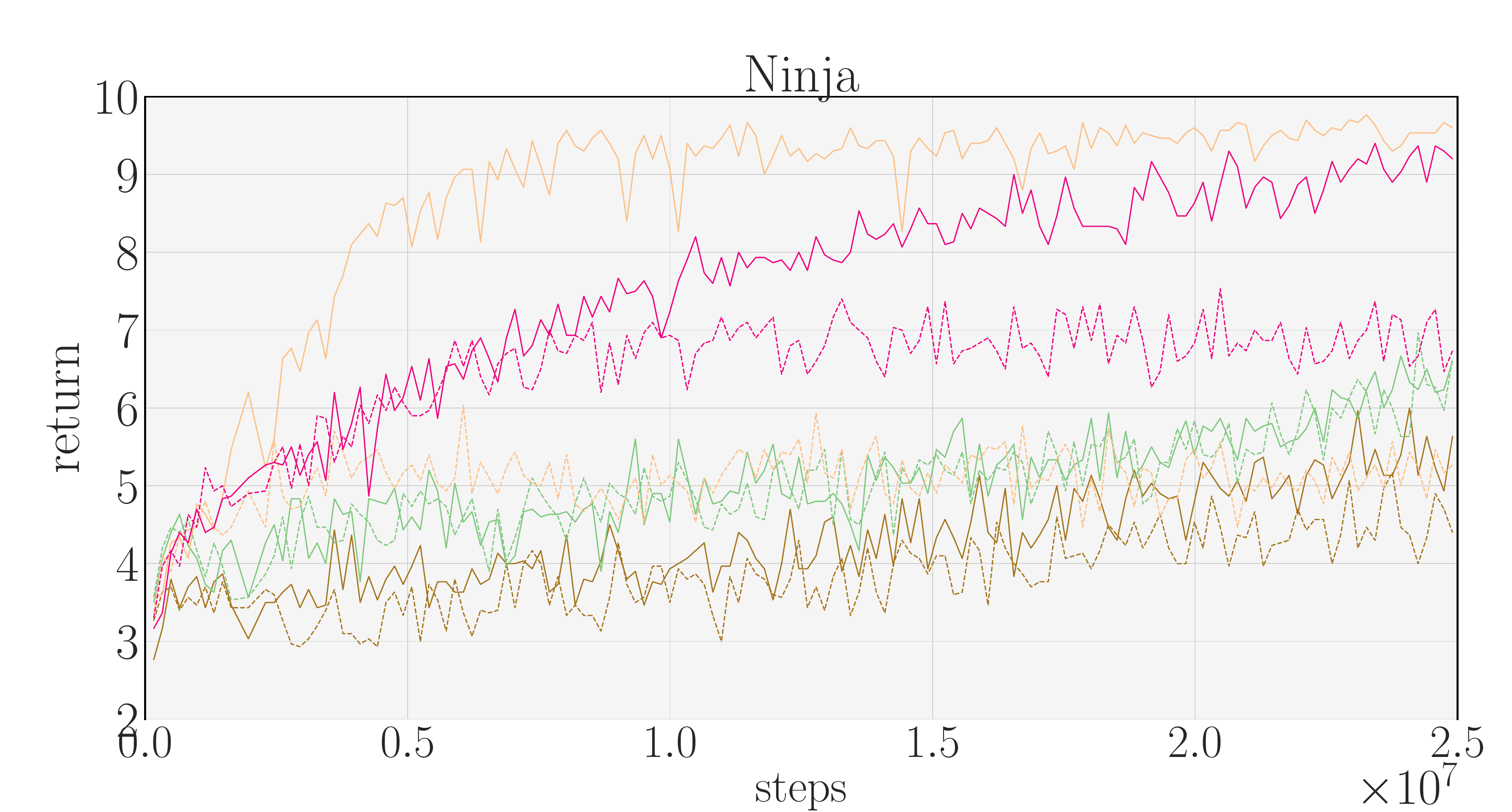} 
    \\
    \includegraphics[width=0.9\columnwidth]{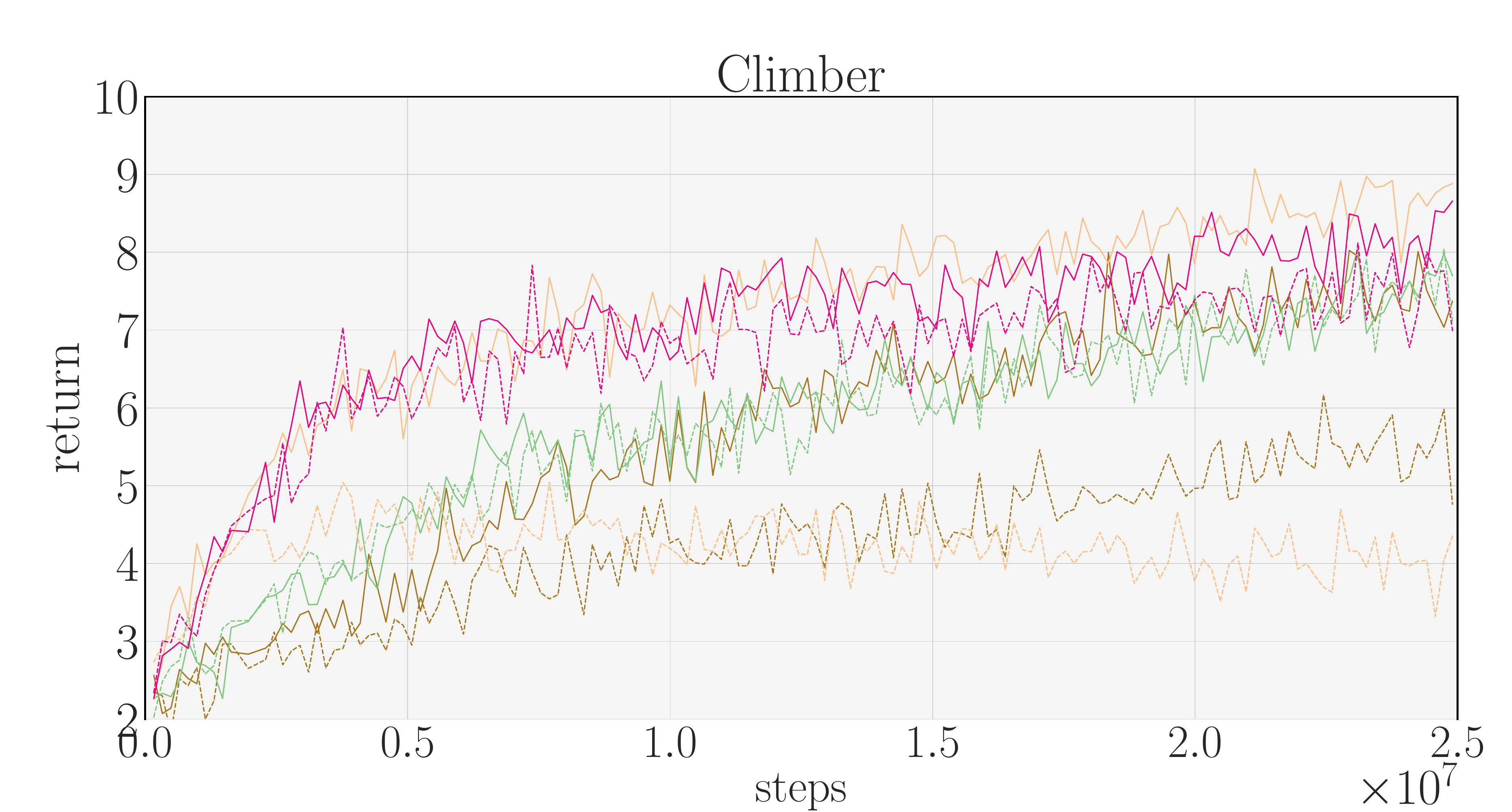} 
    
    \caption{Train (solid) and test (dashed) performance curves when the agent is taught with 200 and 1,000 train levels. Despite IL helps on learning faster with 200 levels when being applied concurrently with RL, the agent struggles to generalize to unseen levels. Increasing the amount of training levels to 1,000 mitigates this gap. However, doing that increases the required number of interactions to achieve an optimal policy. IL helps improve the sample efficiency.
    }
    \label{fig:procgen_1000levels_train_test}
\end{figure}

Figure \ref{fig:procgen_1000levels_train_test} clearly illustrates how using IL enhances sample efficiency, either by achieving the same return with fewer steps or by attaining a better return with the same number of interactions. For example, in the \texttt{Ninja} task with 200 training levels, PPO (brown curve) requires approximately 20M steps to reach a training return of 5, while PPO+IL (orange curve) achieves the same return in just 2–3M steps, demonstrating a 7–10x improvement in sample efficiency. With 1,000 training levels, PPO (green curve) takes 10M steps to achieve a return of 5, whereas PPO+IL (pink curve) achieves a return of 7 within the same number of steps. Additionally, Figure \ref{fig:procgen_1000levels_train_test}  reveals that although a generalization gap exists between training and testing performance, this gap narrows as the number of training levels increases. Nonetheless, training the agent on a larger number of levels requires more steps, thus reducing overall sample efficiency -- a limitation that can be effectively addressed by leveraging IL. These findings underscore the actual potential of IL: \textbf{regardless of the number of training levels, IL enhances sample efficiency, enabling agents to achieve optimal behavior with fewer online interactions.}

\section{Conclusions} \label{sec:conclus}

\paragraph{Summary}In this paper we have studied the potential of IL from offline data to improve the sample-efficiency and overall performance of on-policy RL algorithms in challenging PCG environments. We have considered the setting of pre-training a policy using IL, as well as concurrently optimizing the policy with IL during online RL training. For this purpose, we collect demonstrations (buffers) belonging to different levels, with variable quality, quantity and diversity. We have shown that pre-training on offline demonstrations leads to a significant jumpstart in the performance, consequently improving sample-efficiency in many tasks, even when the provided demonstrations are far from optimal. Concurrently training the agent with IL and RL during the online training exhibits robust performance, being consistent for demonstrations of various quality. Overall, the best strategy is to combine and use IL for both pre-training and during the online training concurrently with RL. 

\paragraph{Experimental Observations}Our empirical results show that for all MiniGrid tasks, training with just 2 to 5 trajectories enables the agent to learn an optimal policy, unlike RL without demonstrations, which fails to solve these tasks. This indicates that pre-training or concurrent training with IL on a few trajectories significantly enhances agent performance, with the diversity of levels in the pre-training dataset proving more crucial than the quality of demonstrations. In contrast, in Procgen, offline data from a random policy can be nearly as effective as data from a trained policy if it maintains uniform diversity. However, in tasks like \texttt{Ninja} and \texttt{Climber}, a skewed distribution toward easier levels in the offline data can lead to an imbalanced buffer, limiting the agent’s ability to devise strategies for complex levels not represented in the data.

\paragraph{Limitations and Future Work}Our study has exposed several limitations in the use of offline data to learn RL agents for PCG environments. Among them, the management of diversity inside the replay buffer has been identified as a limiting factor, particularly in environments where the similarity between levels is large. Several research directions can be pursued to tackle this issue. For instance, diversity metrics can be devised to prioritize the sampling of certain levels \cite{jiang_prioritized_2021} or the selection of trajectories that guarantee such a diversity \cite{andres_towards_2022}. Moreover, techniques for unsupervised environment design could also be interesting to generate and cover consequently the whole level distribution ~\cite{dennis_emergent_2020,parker-holder_evolving_2022}. Last but not least, we believe that more advanced IL techniques such as adversarial IL~\cite{ho_generative_2016,orsini_what_2021} and curriculum learning approaches~\cite{liu_curriculum_2022} can be leveraged to further improve the contribution of the collected trajectories to the agent's learned policy. 

\section*{Acknowledgements}

A. Andres and J. Del Ser acknowledge funding support from the Spanish ``Centro para el Desarrollo Tecnológico Industrial'' (CDTI, AI4ES project, grant number CER-20211030) and from the Basque Government through the BIKAINTEK and ELKARTEK (KK-2023/00012) funding programs. J. Del Ser also receives support from the Basque Government through the consolidated research group MATHMODE (ref. IT1456-22).

\section*{Author contributions}

Conceptualization: A. Andres; Methodology: A. Andres, L. Schäfer; Formal analysis and investigation: A. Andres, L. Schäfer; Writing - original draft preparation: A. Andres, L. Schäfer; Writing - review and editing: S. Albrecht, J. Del Ser; Supervision: S. Albrecht, J. Del Ser.

\section*{Conflict of Interest}
The authors declare that they have no conflicts of interest regarding this work.

\section*{Code and Data Availability}
Code is available at: \url{https://github.com/uoe-agents/imitation-learning-pcg}. Additionally, the original buffers for some experiments are released. For experiments where the buffers are not available, the manuscript details the methodology used to collect the data, and the repository includes examples demonstrating how to replicate this data collection procedure. 


\bibliographystyle{elsarticle-num}      
\bibliography{biblos.bib}

\clearpage
\begin{appendix}
\section{Hyperparameters \& Neural Network Architectures} \label{app:hyperparams}
\begin{table}[h!]
    \centering
    \caption{PPO Hyperparameters}
        \begin{tabular}{rcc}
            \toprule
            Hyperparameter & MiniGrid & Procgen \\
            \midrule
            Optimiser & Adam & Adam\\
            Learning Rate & $10^{-4}$ & $5\cdot10^{-4}$\\
            Adam epsilon & $10^{-5}$ &  $10^{-5}$\\  
            Environment steps per update & 2048 & 16384\\
            Discount $\gamma$ & 0.99 & 0.999 \\
            GAE $\lambda$ & 0.95 & 0.95 \\
            Entropy coefficient & 0.01 & 0.01\\
            Value loss coefficient & 0.5 & 0.5 \\
            Number of epochs & 4 & 3\\
            Number of minibatches & 4 & 8\\
            PPO clipping constant & 0.2 & 0.2\\
            Max grad norm & 0.5 & 0.5\\
            \bottomrule
        \end{tabular}
    \label{tab:ppo_hyperparameters}
\end{table}

In our study we adopt the state-of-the-art PPO algorithm \cite{schulman_proximal_2017}. The selected hyperparameters can be found at Table \ref{tab:ppo_hyperparameters}. Moreover, when using IL concurrently with RL, we sample 5 batches, each containing 256 randomly sampled $\{s,a\}$ pairs from the buffer. Thus, we perform 5 optimization steps to the policy $\pi$ in each IL update.

\paragraph{MiniGrid} Unless otherwise stated, two independent actor and critic models are used. In line with other successful approaches \cite{andres_towards_2022,zha_rank_2021}, both networks consist of 2 fully connected layers of 64 neurons each, using tanh activation functions. It is important to note that the IL gradients are only applied to the actor network, compelling the agent to mimic the $\{s,a\}$ tuples provided in the demonstrations. This means the critic is not directly influenced by IL.

\paragraph{Procgen} 
The agent is parameterized by the \textit{large} ResNet architecture from \cite{espeholt_impala_2018}, which was used to achieve the best results in \cite{cobbe_leveraging_2020}. This structure constitutes a shared actor-critic. As a consequence, the IL 
optimization steps
would not only affect to the actor-head, but also to the critic-head. 

\begin{figure*}[t]
    \centering
    \resizebox{2\columnwidth}{!}{\begin{tabular}{c|ccc}
    \multicolumn{4}{c}{\includegraphics[width=\columnwidth]{diversity_legend_crop.pdf}} 
    \\
    & \texttt{O1Dlhb 10\% Buffer} & \texttt{O1Dlhb 60\% Buffer} & \texttt{O1Dlhb 90\% Buffer} 
    \\
    \rotatebox[origin=l]{90}{Filled to max capacity} &
    \includegraphics[width=0.65\columnwidth]{O1Dlhb_online_diversity_nonoptimal.pdf} &
    \includegraphics[width=0.65\columnwidth]{O1Dlhb_online_diversity_suboptimal.pdf} &
    \includegraphics[width=0.65\columnwidth]{O1Dlhb_online_diversity_optimal.pdf}
    \\
    \rotatebox[origin=l]{90}{Almost empty} &
    \includegraphics[width=0.65\columnwidth]{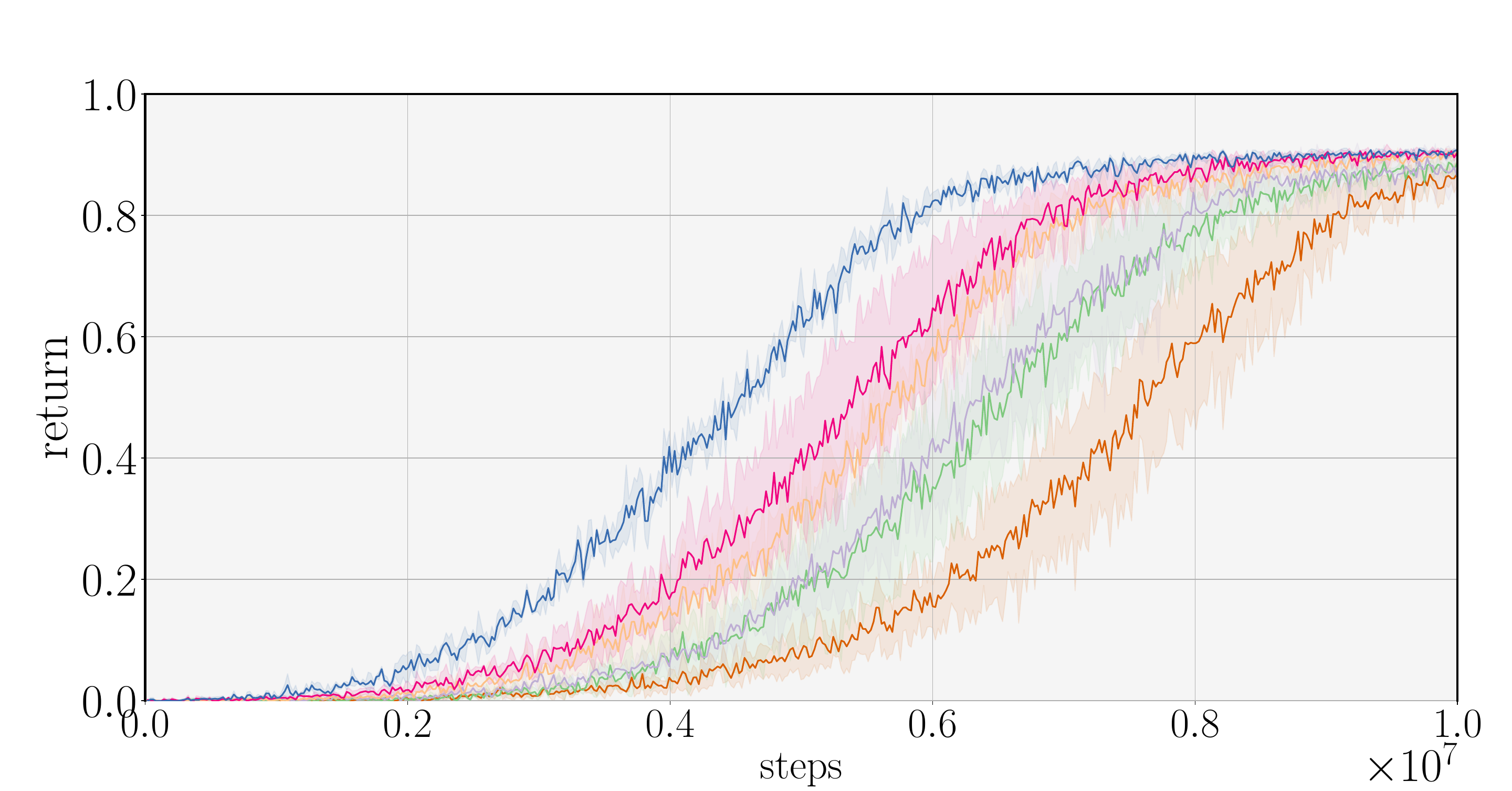} &
    \includegraphics[width=0.65\columnwidth]{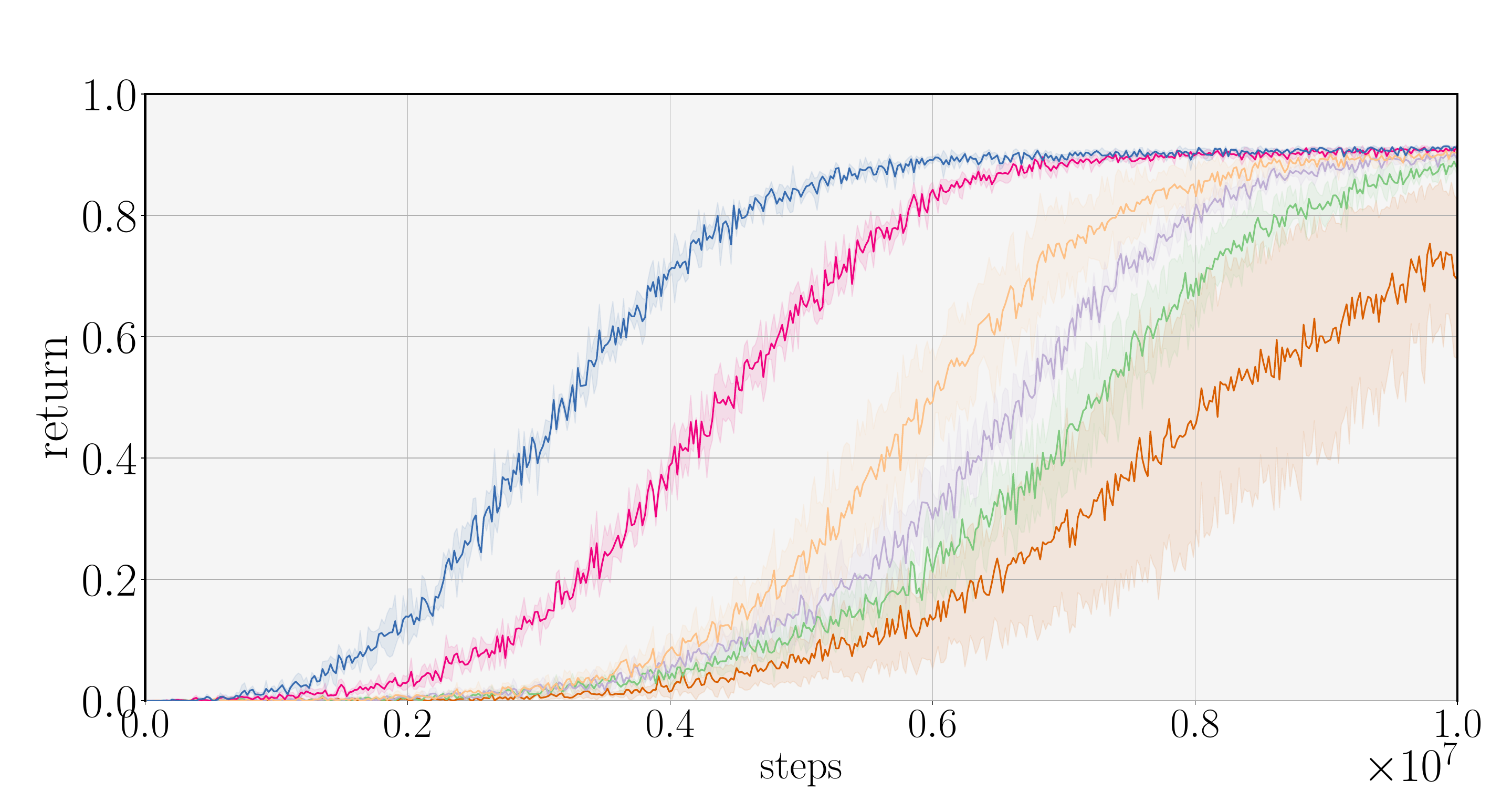} &
    \includegraphics[width=0.65\columnwidth]{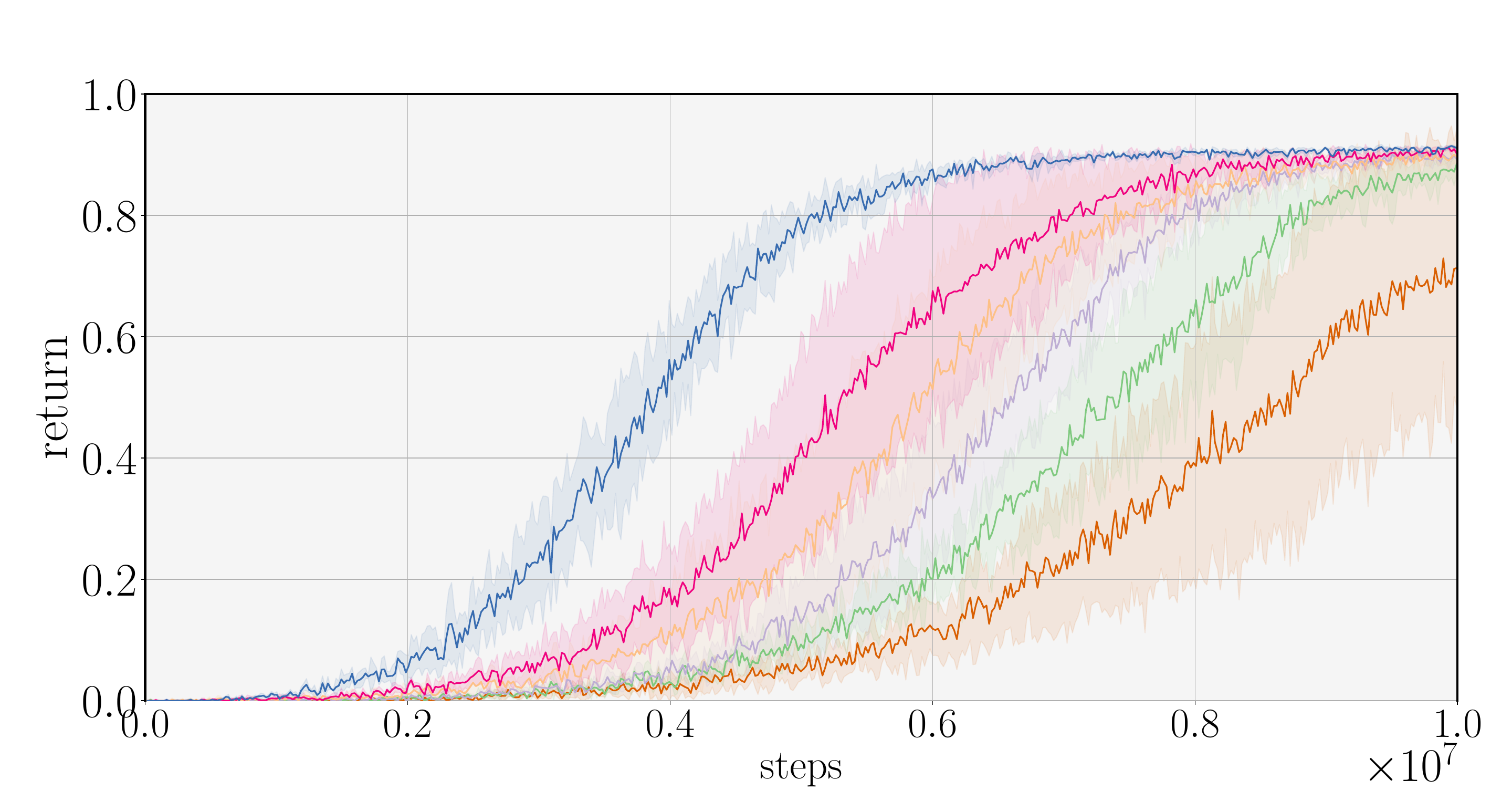}

    \\
    \hline
    
    & \texttt{MN12S10 10\% Buffer} & \texttt{MN12S10 60\% Buffer} & \texttt{MN12S10 90\% Buffer} 
    \\
    \rotatebox[origin=l]{90}{Filled to max capacity} &
    \includegraphics[width=0.65\columnwidth]{MN12S10_online_diversity_nonoptimal.pdf} &
    \includegraphics[width=0.65\columnwidth]{MN12S10_online_diversity_suboptimal.pdf} &
    \includegraphics[width=0.65\columnwidth]{MN12S10_online_diversity_optimal.pdf}
    \\
    \rotatebox[origin=l]{90}{Almost empty} &
    \includegraphics[width=0.65\columnwidth]{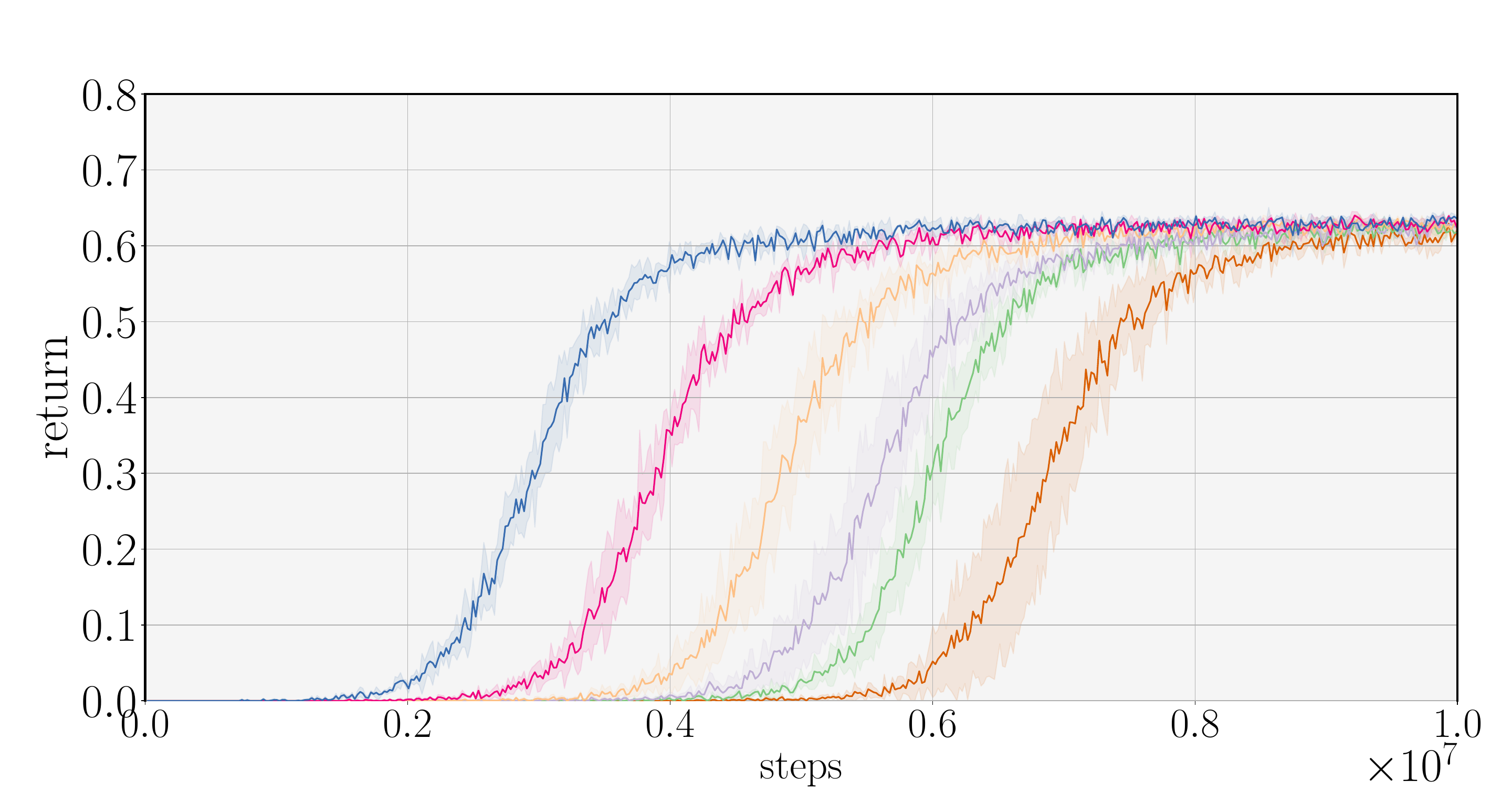} &
    \includegraphics[width=0.65\columnwidth]{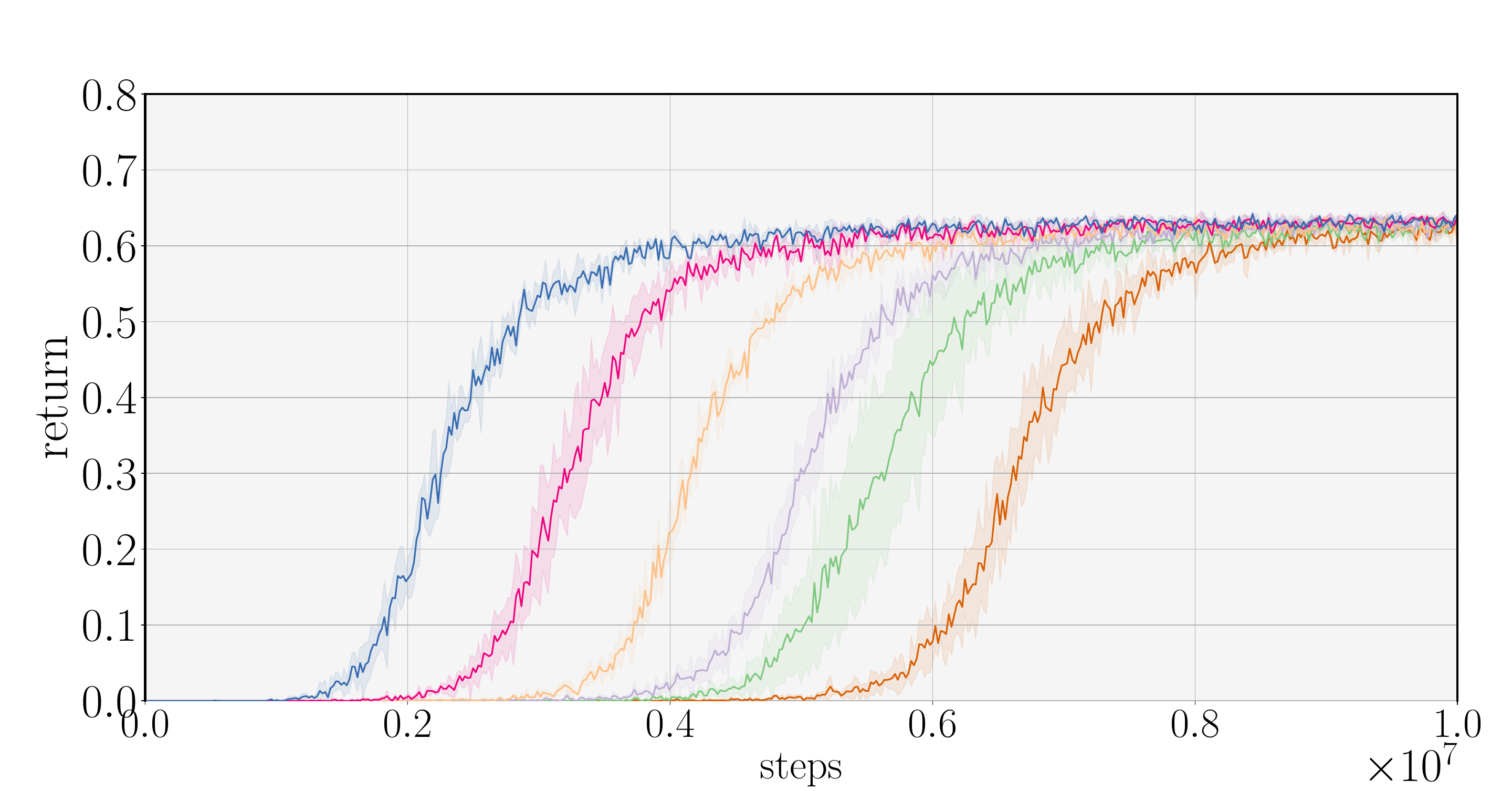} &
    \includegraphics[width=0.65\columnwidth]{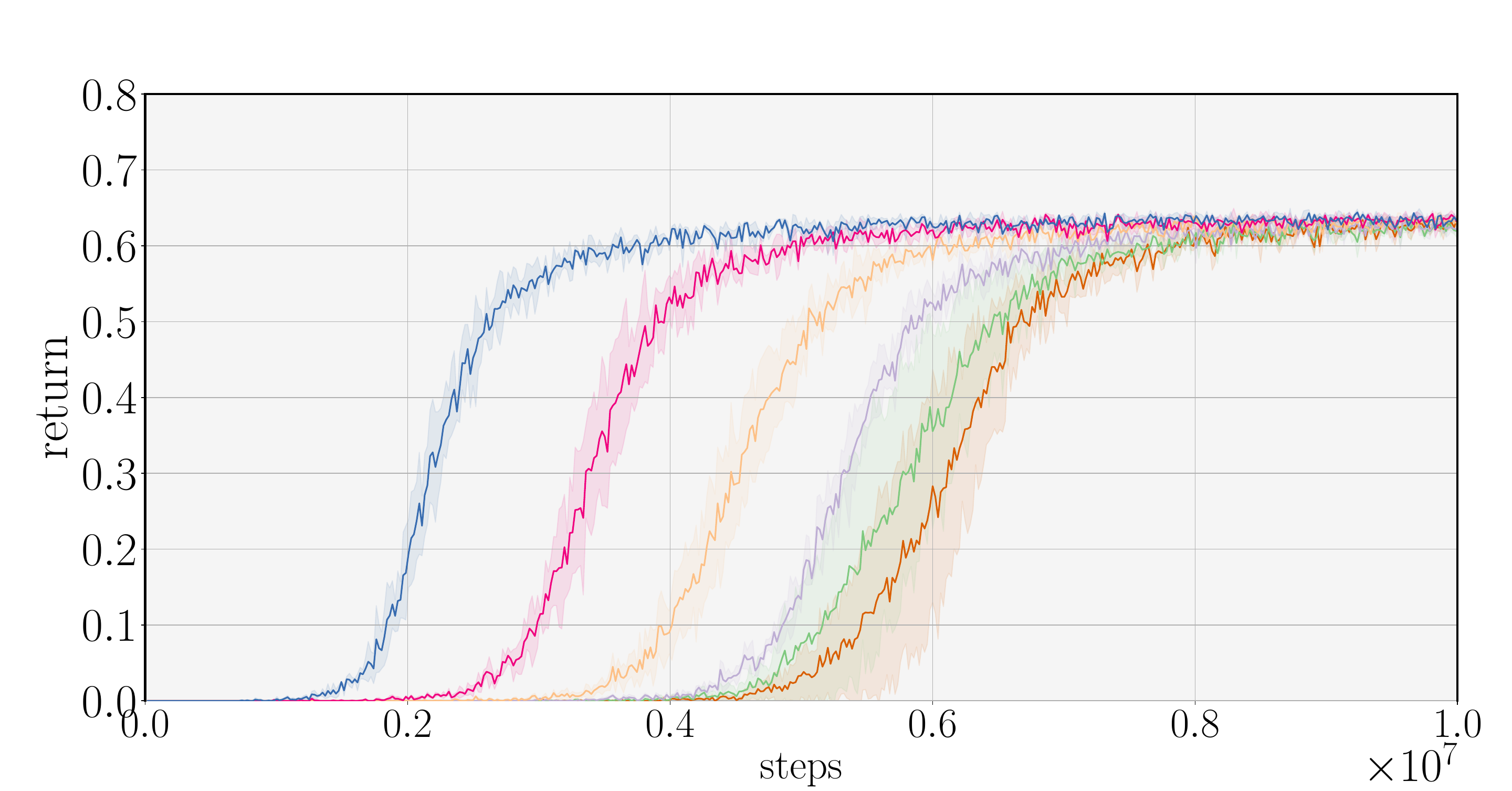}
    \\
    \\
    \end{tabular}}
    \caption{Agent performance when randomly initializing the policy and using concurrently IL and RL losses during the online training with different fixed number of trajectories (one per level) at \texttt{O1Dlhb} and \texttt{MN12S10}. On the top rows the buffer is initialized with the provided demonstrations and they are repeated until filling the whole buffer capacity (i.e., 10,000 experiences), whereas on the bottom, the same demonstrations are pre-loaded (without repetition) and led place to upcoming trajectories to be stored.}
    \label{fig:buffer_population_diversity_online_minigrid}
\end{figure*}

\section{Concurrent Reinforcement and Imitation Learning: Populating The Buffer}\label{app:populate_buffer_online_diversity_minigrid}

In traditional Imitation Learning, the learning process is given by the necessity to have a set of pre-collected demonstrations. Thus, the amount of these offline demonstrations is fixed. In contrast, Self-Imitation Learning operates under a different paradigm where such pre-collection is not typically required. As a consequence, the learning is not limited by a predefined quantity of demonstrations, but rather the capacity of a specific buffer designed to store a certain number of experiences, which can encompass a large (and undefined) number of trajectories \footnote{We note that these constraints regarding the number of demonstrations or buffer size are ultimately determined by the algorithm designer, and not governed by fixed rules.}. 

In our concurrent learning approach --Section \ref{sec:methodology}-- we set a maximum buffer size criteria. We load pre-collected demonstrations in that buffer, where we also consider the possibility of rearranging the content of it if trajectories of better quality are collected.
However, when dealing with a limited number of offline-collected demonstrations, the buffer that we are going to employ would be nearly empty. To address this, we explored the consequences of populating the buffer in two specific ways:
\begin{enumerate}
    \item \textbf{Filled to max capacity:} The agent's buffer is initially populated with available demonstrations. Subsequently, these experiences are replicated at random until the maximum capacity of the buffer is reached. This option is an analogous (and adapted) version of solely using the initially provided demonstrations when employing a buffer size limit criteria.
    \item \textbf{Almost empty:} The agent's buffer is stocked with the available demonstrations but leaves space for trajectories to be collected during the online phase. It is worth noting that the trajectories that are going to be acquired on the initial interactions of the online phase are likely to represent failures or non-optimal behaviors.    
\end{enumerate}

By examining Figure \ref{fig:buffer_population_diversity_online_minigrid}, it can be noticed that \textit{filling to the max capacity} consistently offers superior sample efficiency. Yet, this method might backfire and produce subpar results if the stored data is of poor quality or exhibits low diversity. For instance, the performance plummets in \texttt{O1Dlhb} when relying on a 10\% buffer with just one level. A similar decline in performance is observed in \texttt{MN12S10} with a singular level, irrespective of the quality of the selected demonstration. This downturn can be attributed to the reluctance to replace existing trajectories, unless the newly collected ones surpass those in the buffer in terms of overall score, as given in Equation \eqref{eq:rapid_scores}.

On the contrary, when using the \textit{almost empty} setup, we let space for new upcoming trajectories. While these trajectories might initially be suboptimal, their quality improves iteratively as the agent progresses during training. Although this approach might hinder sample efficiency, it fosters more stable and consistent learning outcomes, a trend observable across all buffer types for \texttt{MN12S10}.

\section{Successfully Solved Levels}\label{app:success_levels_training}

Besides the return, which is used to evaluate the performance of the agent's learned policy, the success ratio can also be an important metric to consider for this purpose. This is particularly relevant in Procgen, where the complexity exhibited by the levels can vary significantly. 

\begin{figure*}[t]
    \centering
    \begin{tabular}{cc}
        \multicolumn{2}{c}{\includegraphics[width=0.4\textwidth]{normal_random_rapid_legend_crop.pdf}}
        \\
        \includegraphics[width=0.45\textwidth]{ninja_concurrent.pdf} &
        \includegraphics[width=0.45\textwidth]{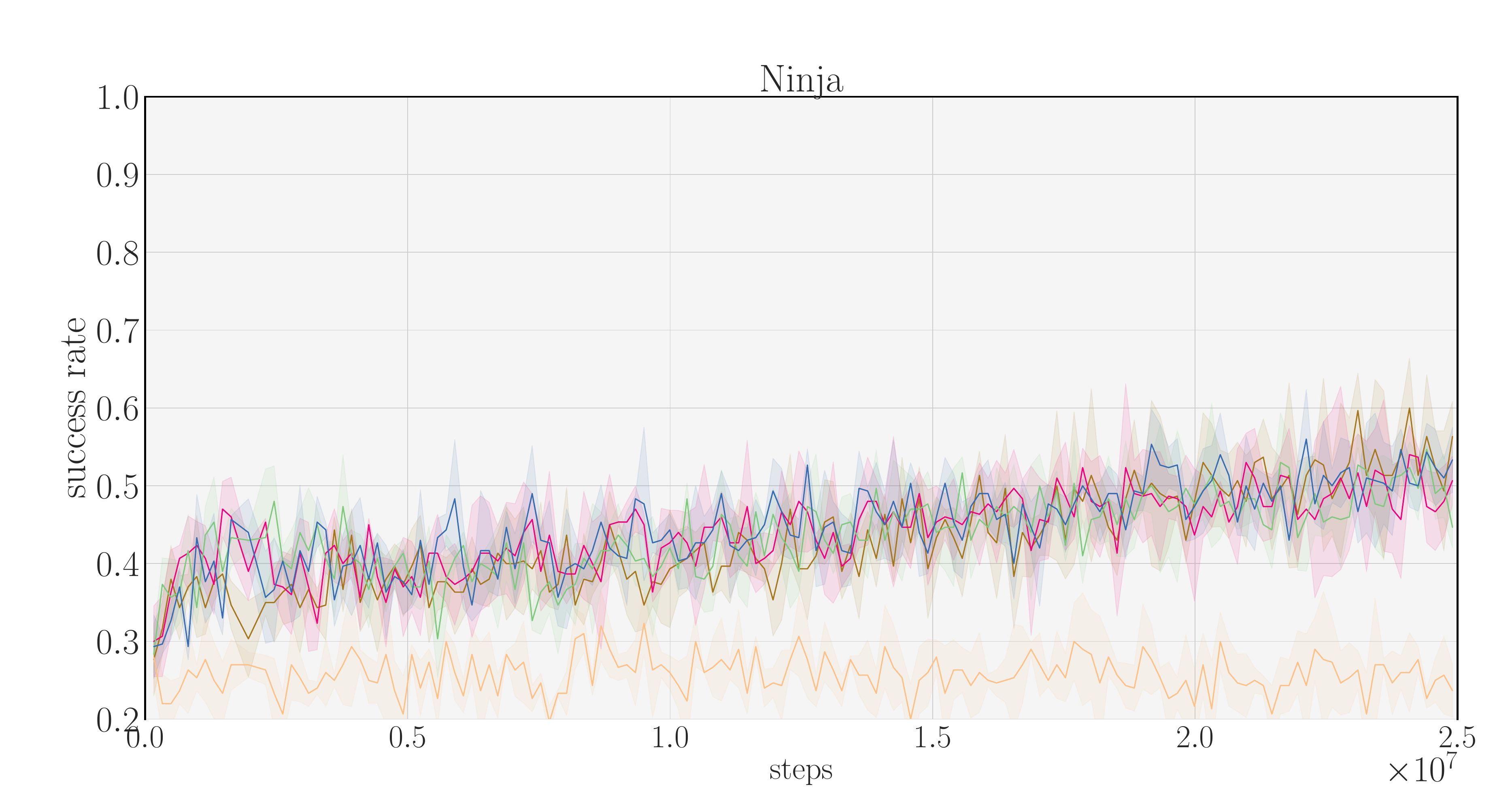}
        \\
        \includegraphics[width=0.45\textwidth]{climber_concurrent.pdf} &
        \includegraphics[width=0.45\textwidth]{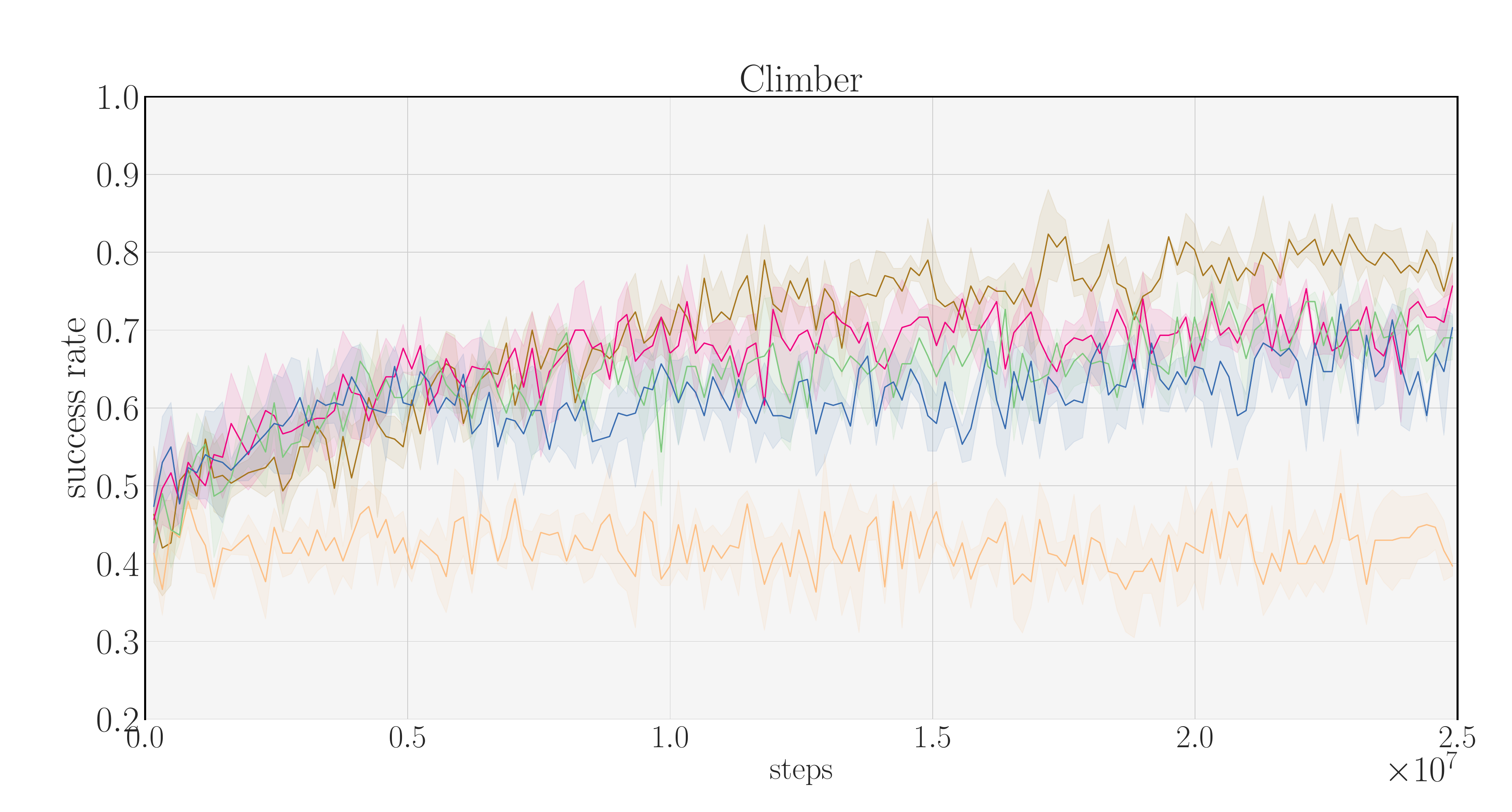}
        \\
        \toprule
        \\
        \multicolumn{2}{c}{\includegraphics[width=0.8\textwidth]{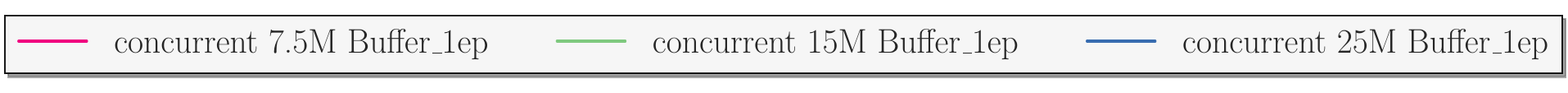}}
        \\
        \includegraphics[width=0.45\textwidth]{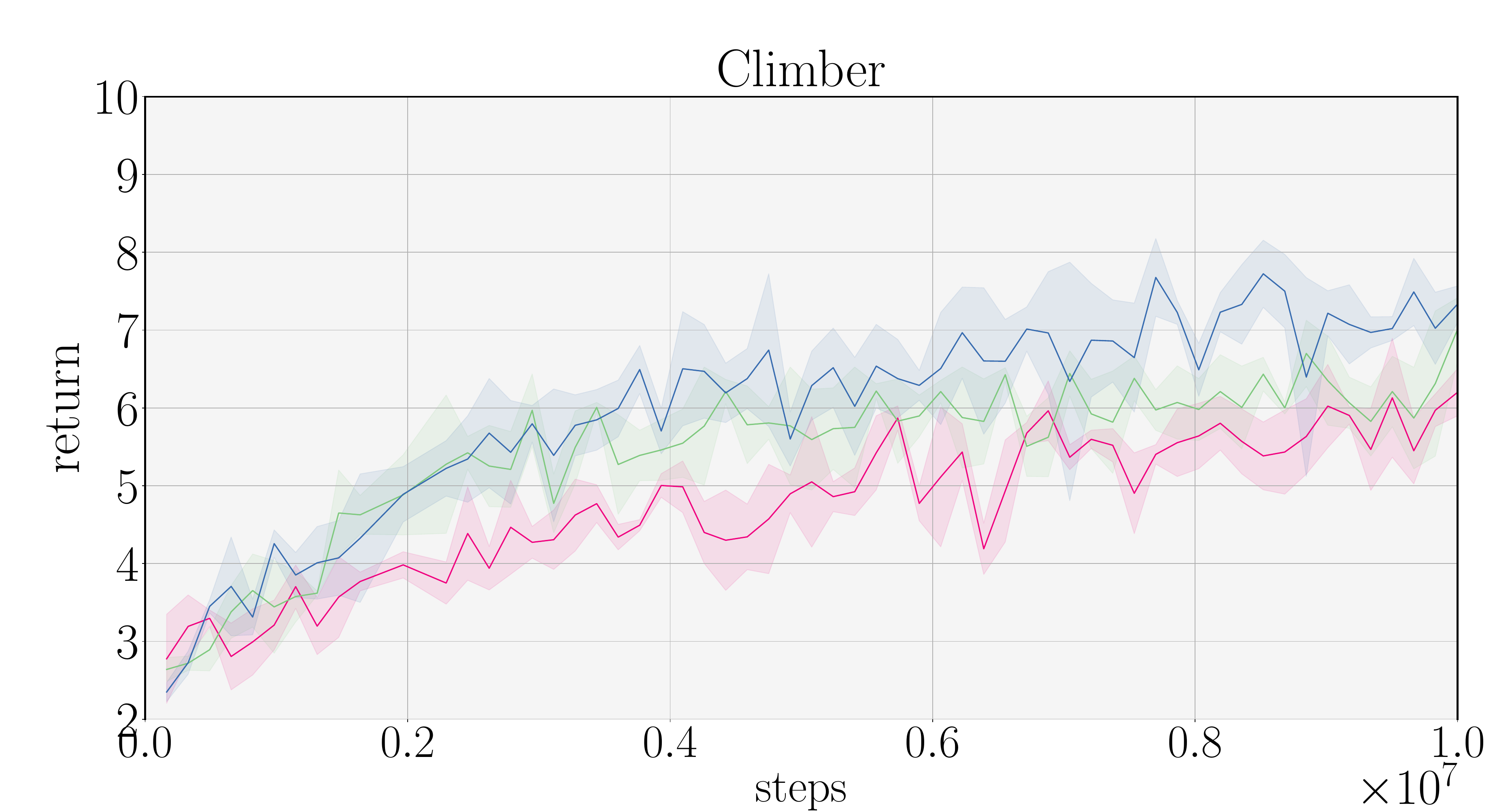} &
        \includegraphics[width=0.45\textwidth]{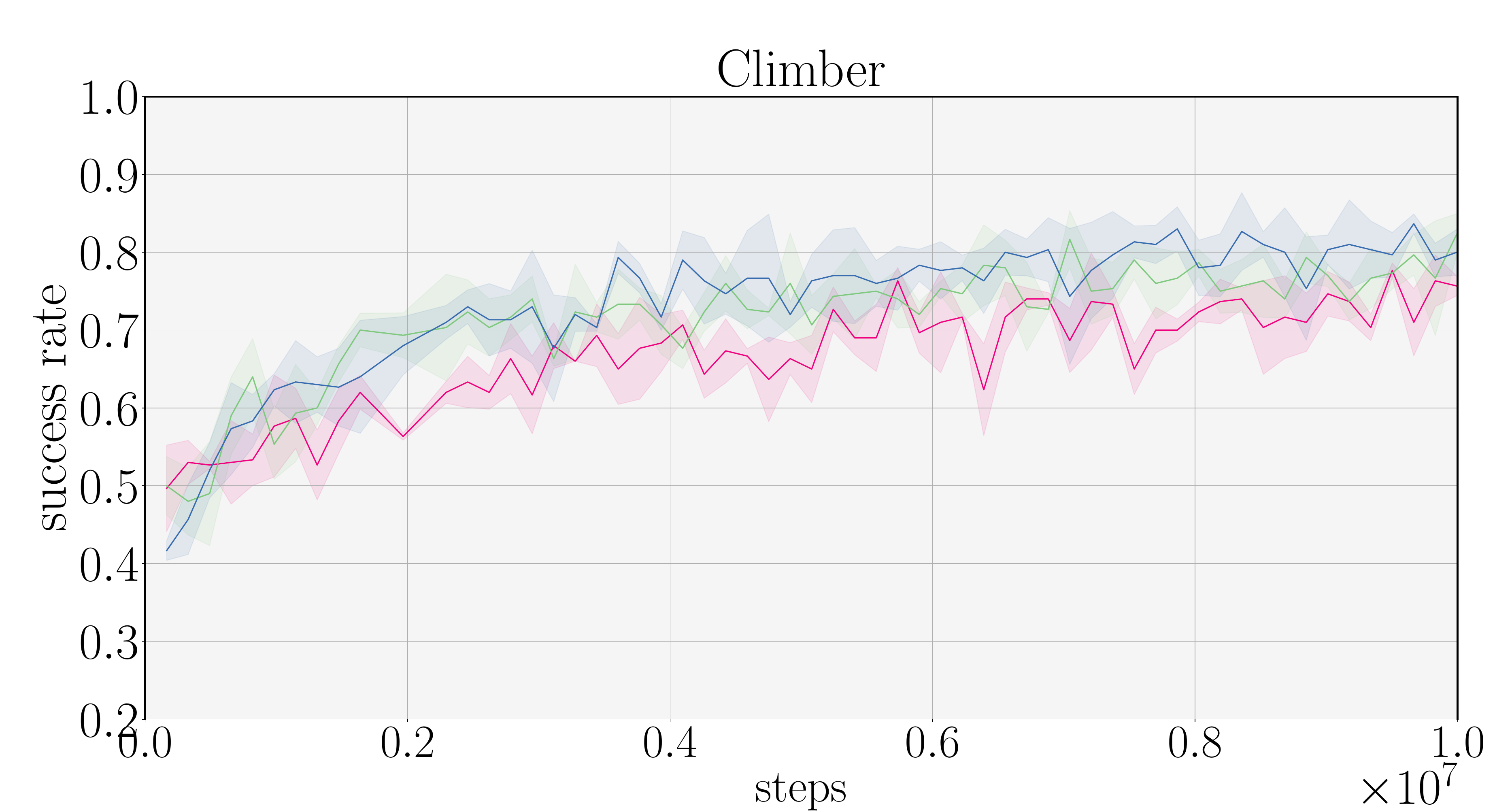}
        \\
        \toprule
        \\
        \multicolumn{2}{c}{\includegraphics[width=0.4\textwidth]{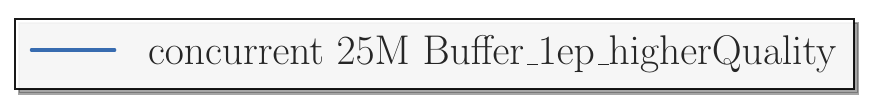}}
        \\
        \includegraphics[width=0.45\textwidth]{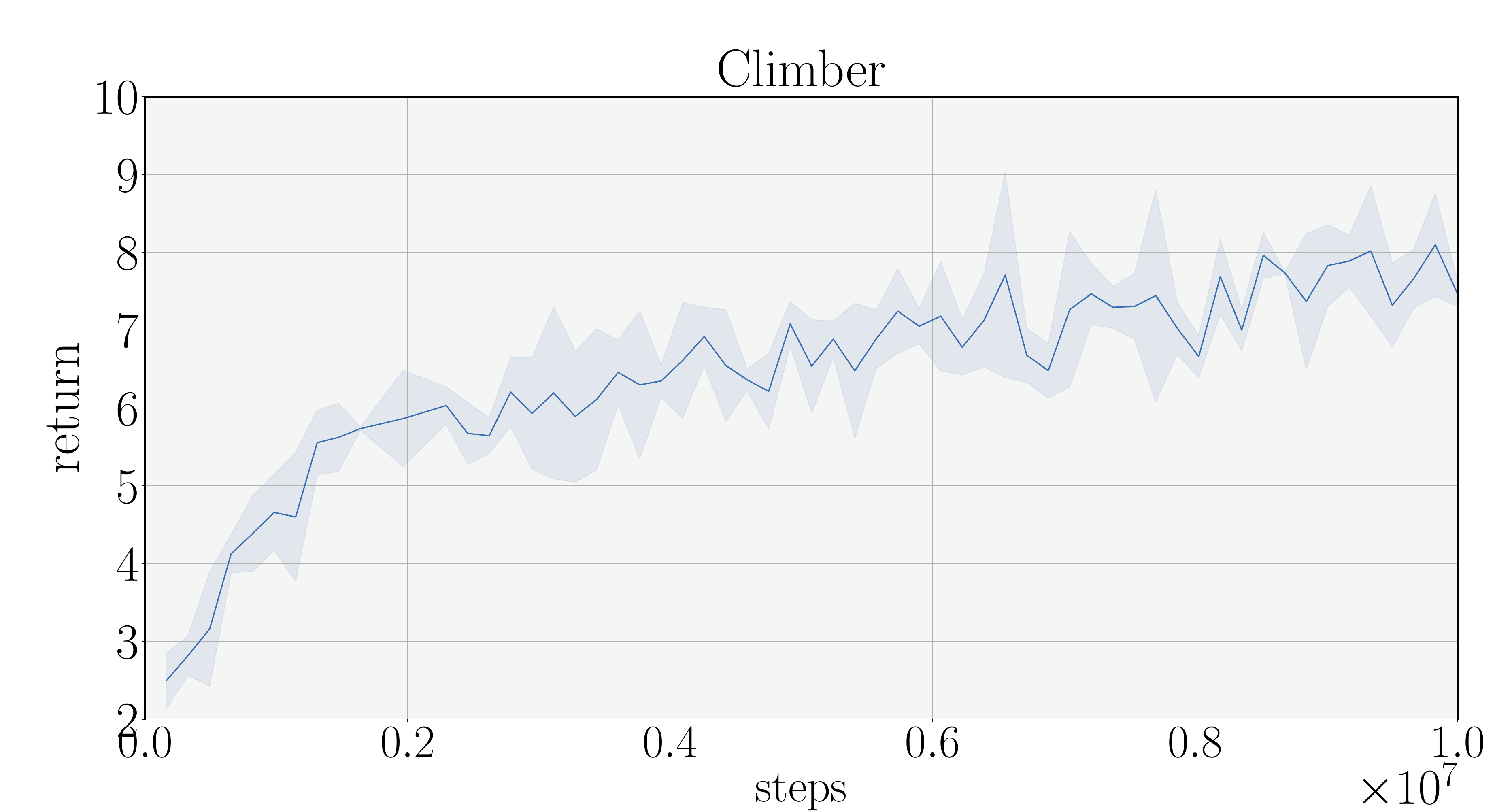} &
        \includegraphics[width=0.45\textwidth]{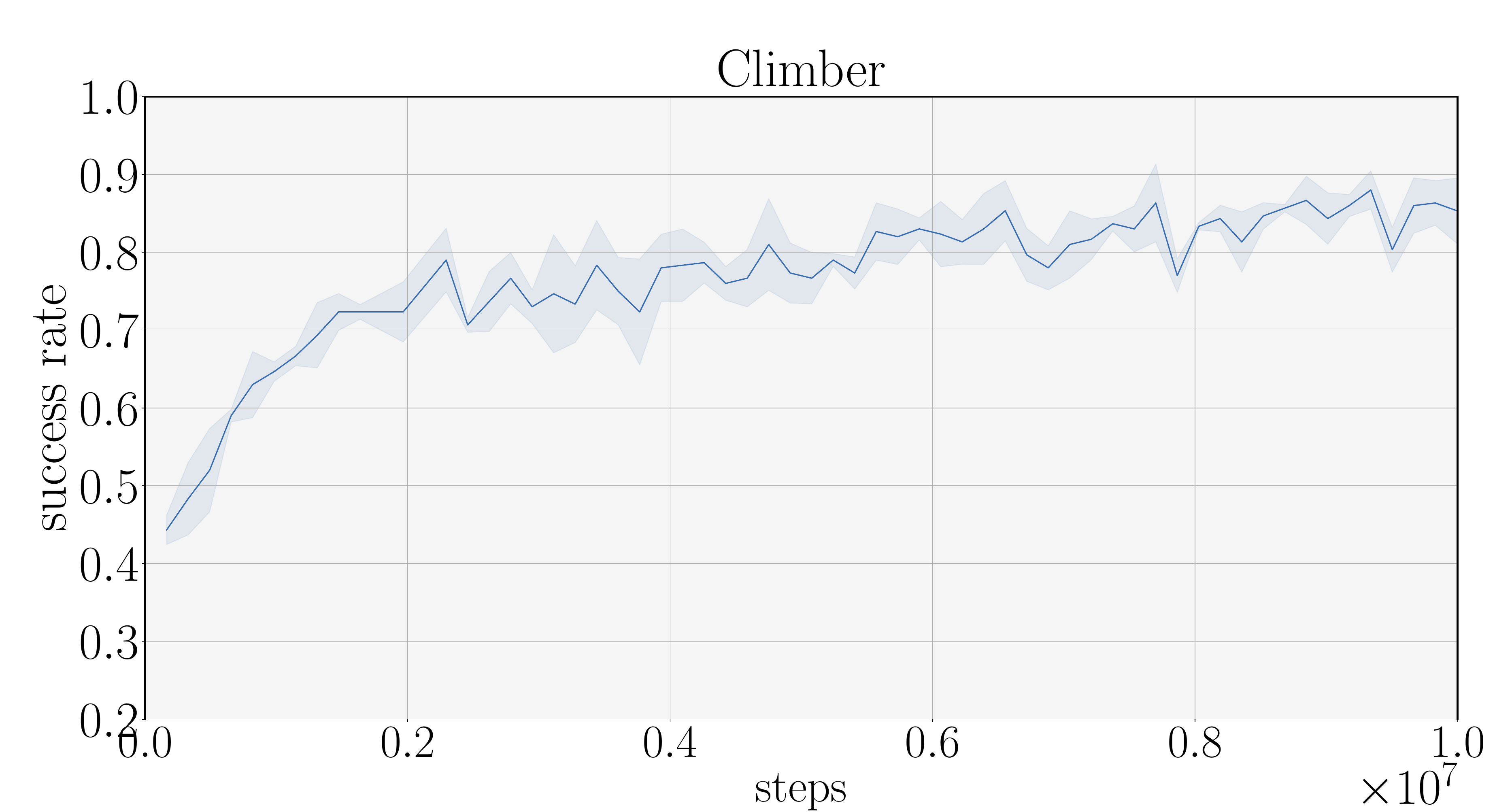}

    \end{tabular}
    \caption{Return (left) and Success ratio (right) of the agent when concurrently training the agent with RL and IL in \texttt{Ninja} and \texttt{Climber} when using different buffers. It can be seen that the success ratio is different from the obtained return in \texttt{Climber} due to its the reward function ($\mathcal{R}$) design.}
    \label{fig:concurrent_training_procgen_successratio}
\end{figure*}

Figure  \ref{fig:concurrent_training_procgen_successratio} shows the obtained return (left) and the success rate (right) of the agent when being trained concurrently with RL and IL with different buffers. Although in \texttt{Ninja} both the return and the success ratio are highly correlated with each other, this does not hold in the case of the \texttt{Climber} task. In fact, the success ratio in the latter tends to be slightly higher than the return. It is interesting to note that even a random agent is able to consistently solve $\sim$25\% and $\sim$40\% of the levels in \texttt{Ninja} and \texttt{Climber}, respectively. 

Interestingly, even if we use any of the considered variants of \textit{Buffer\_1ep} for the \texttt{Climber}, where we provide 99\% of successful episodes, the agent barely achieves an 80-85\% success ratio. This is, the agent is not able to acquire the required knowledge in certain levels, even providing demonstrations that leverage the completion of the task in that level.

\begin{figure*}[t]
    \centering
    \begin{tabular}{ccc}
        $\pi_0^{ninja}$ & $\pi_1^{ninja}$ & $\pi_2^{ninja}$
        \\
        \includegraphics[width=0.3\textwidth]{successfullevels_ninja_1ep_seed0.pdf} &
        \includegraphics[width=0.3\textwidth]{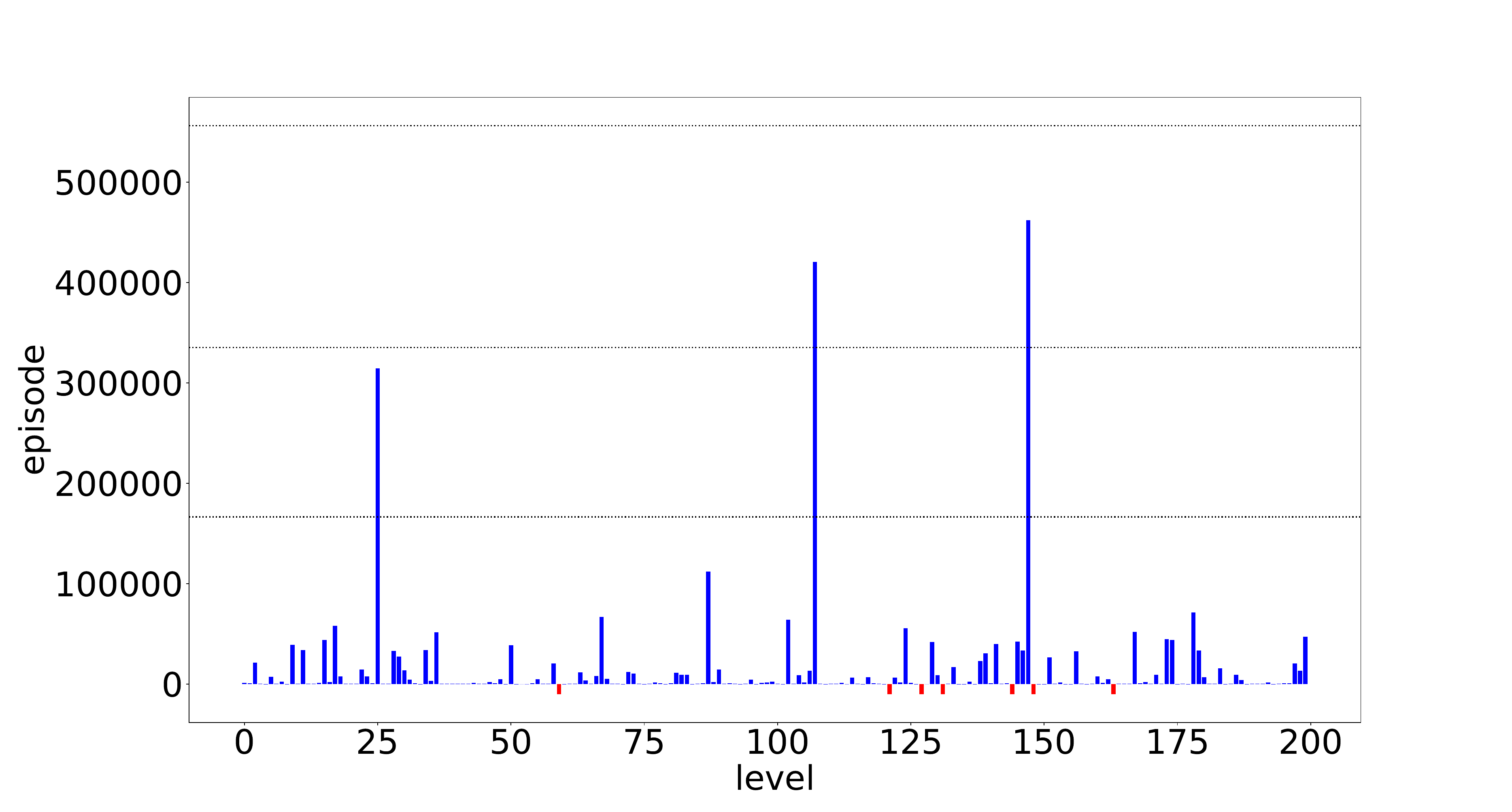} &
        \includegraphics[width=0.3\textwidth]{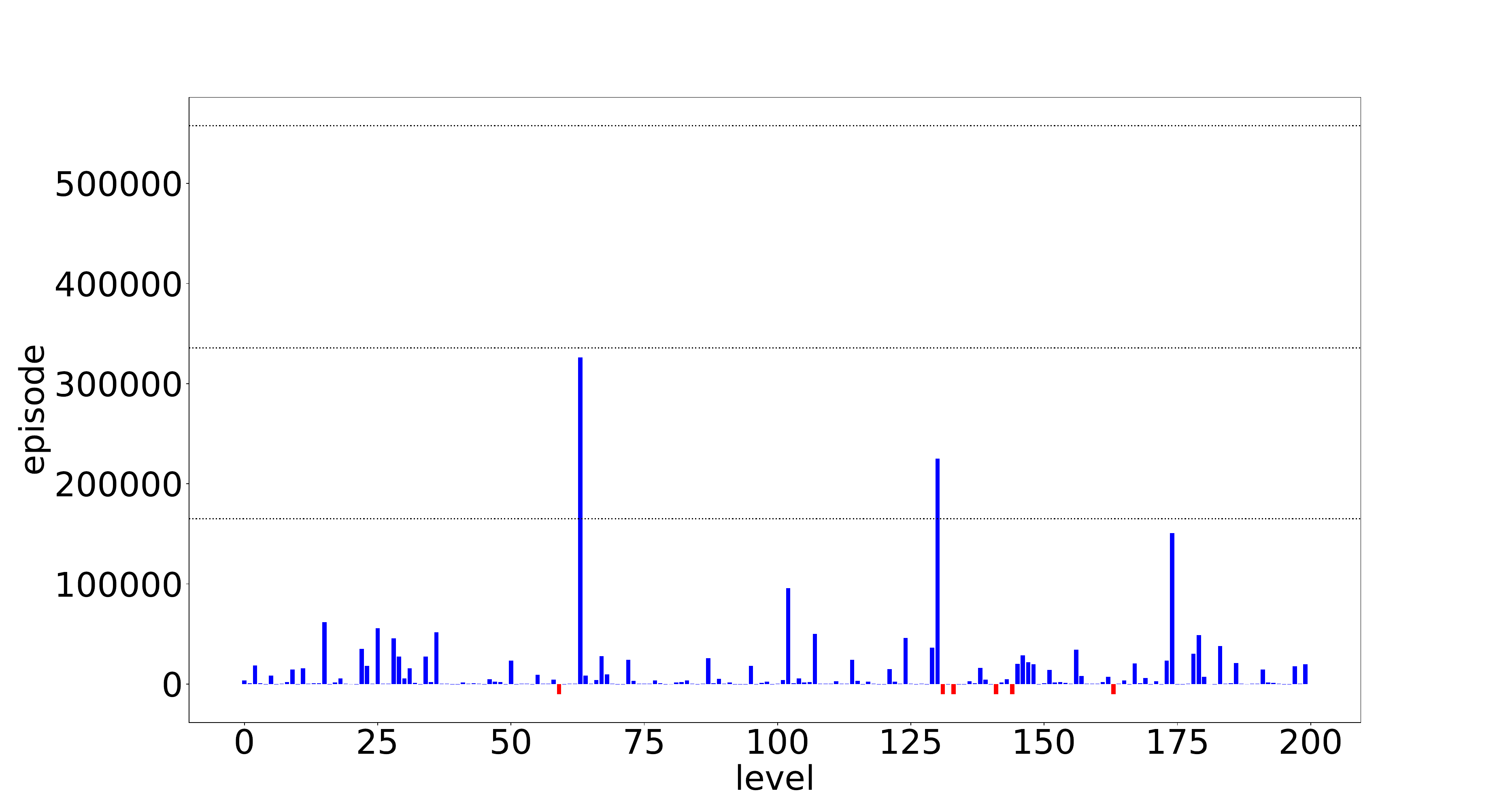}
        \\
        $\pi_0^{climber}$ & $\pi_1^{climber}$ & $\pi_2^{climber}$
        \\
        \includegraphics[width=0.3\textwidth]{successfullevels_climber_1ep_seed0.pdf} &
        \includegraphics[width=0.3\textwidth]{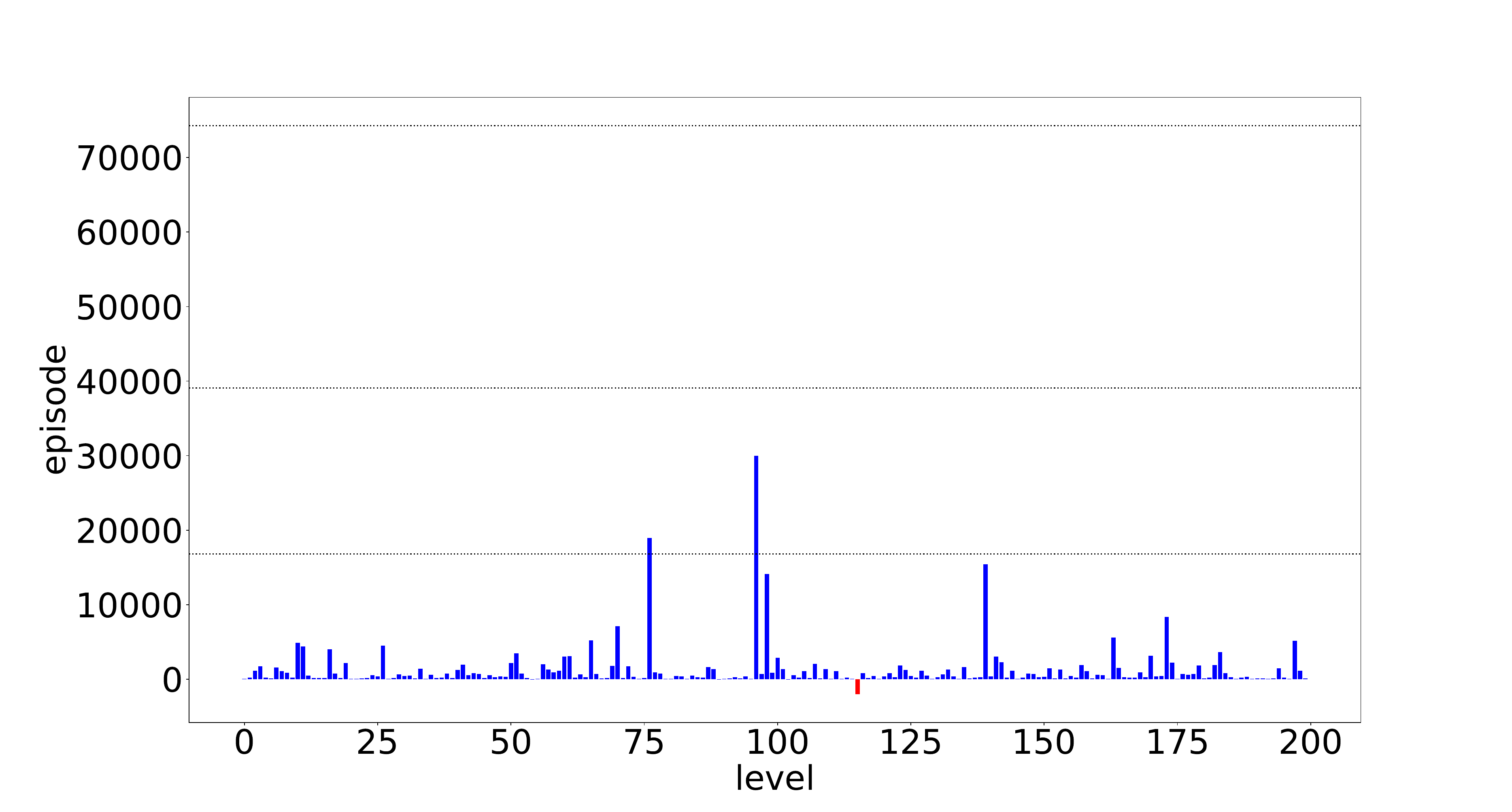} &
        \includegraphics[width=0.3\textwidth]{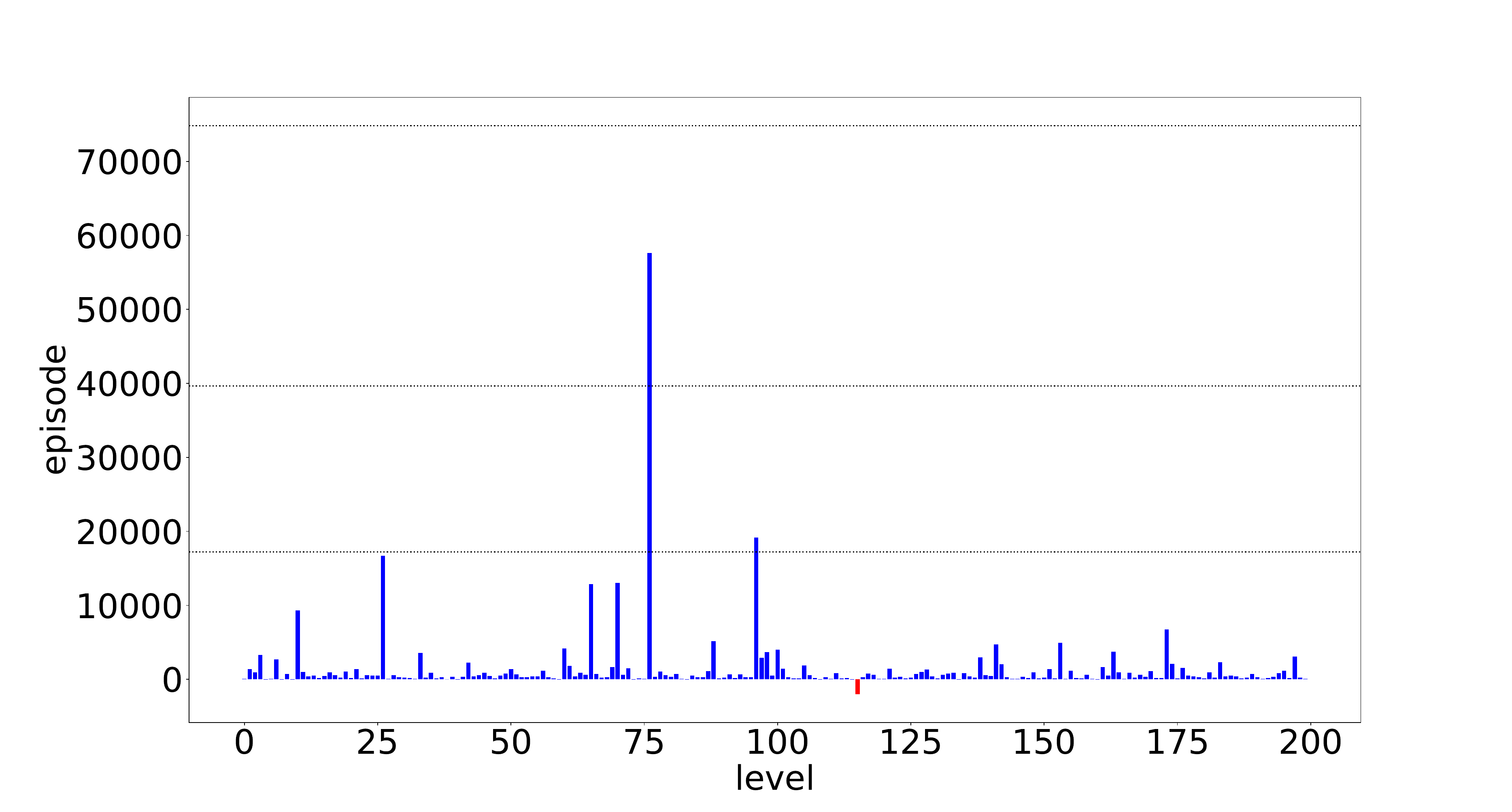} 
        \\
        \multicolumn{3}{c}{(a) Resulting policies after training the agent with \textit{Buffer\_1ep}}\\
        \\
        \\
        $\pi_0^{ninja}$ & $\pi_1^{ninja}$ & $\pi_2^{ninja}$
        \\
        \includegraphics[width=0.3\textwidth]{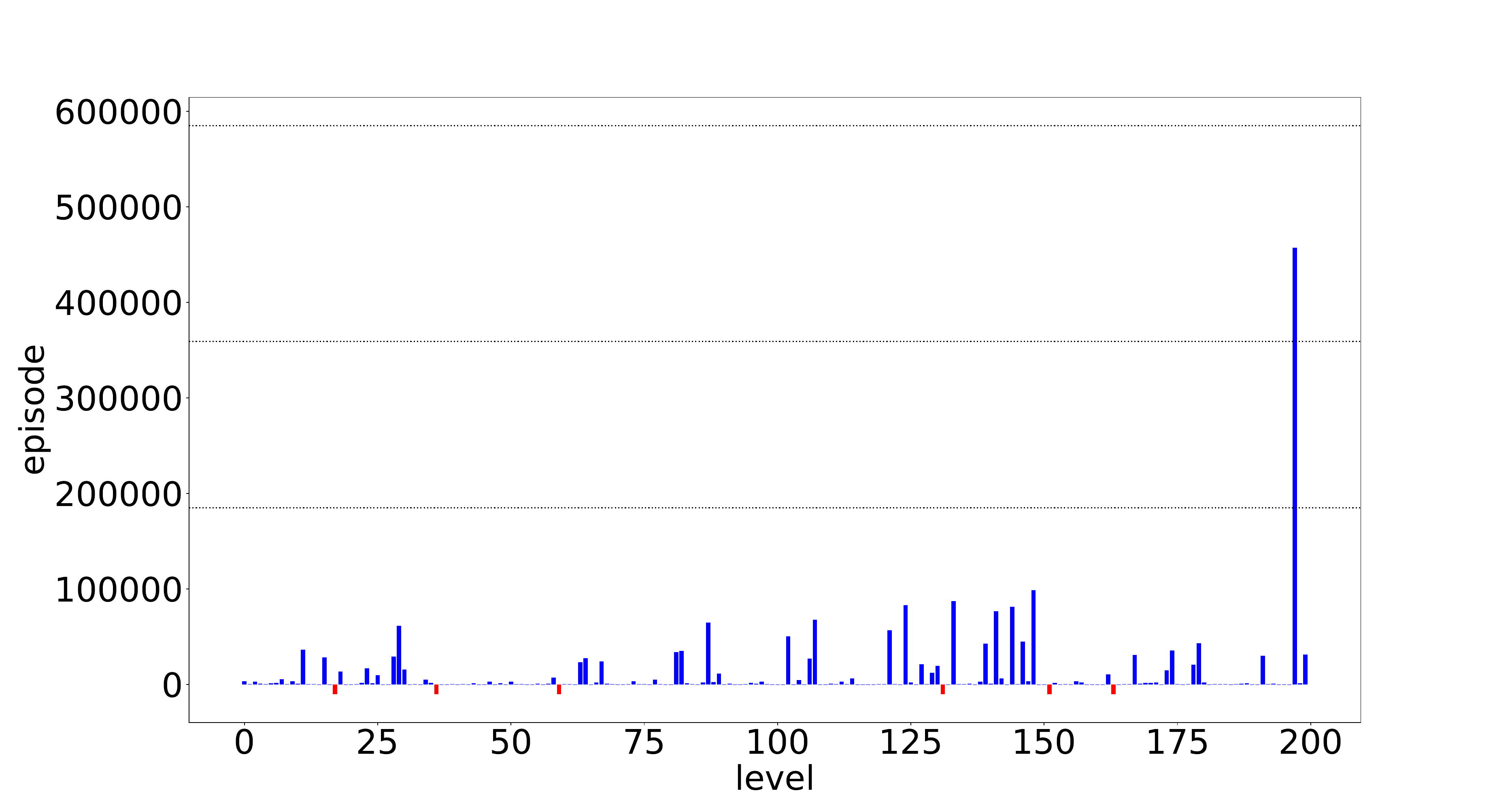} &
        \includegraphics[width=0.3\textwidth]{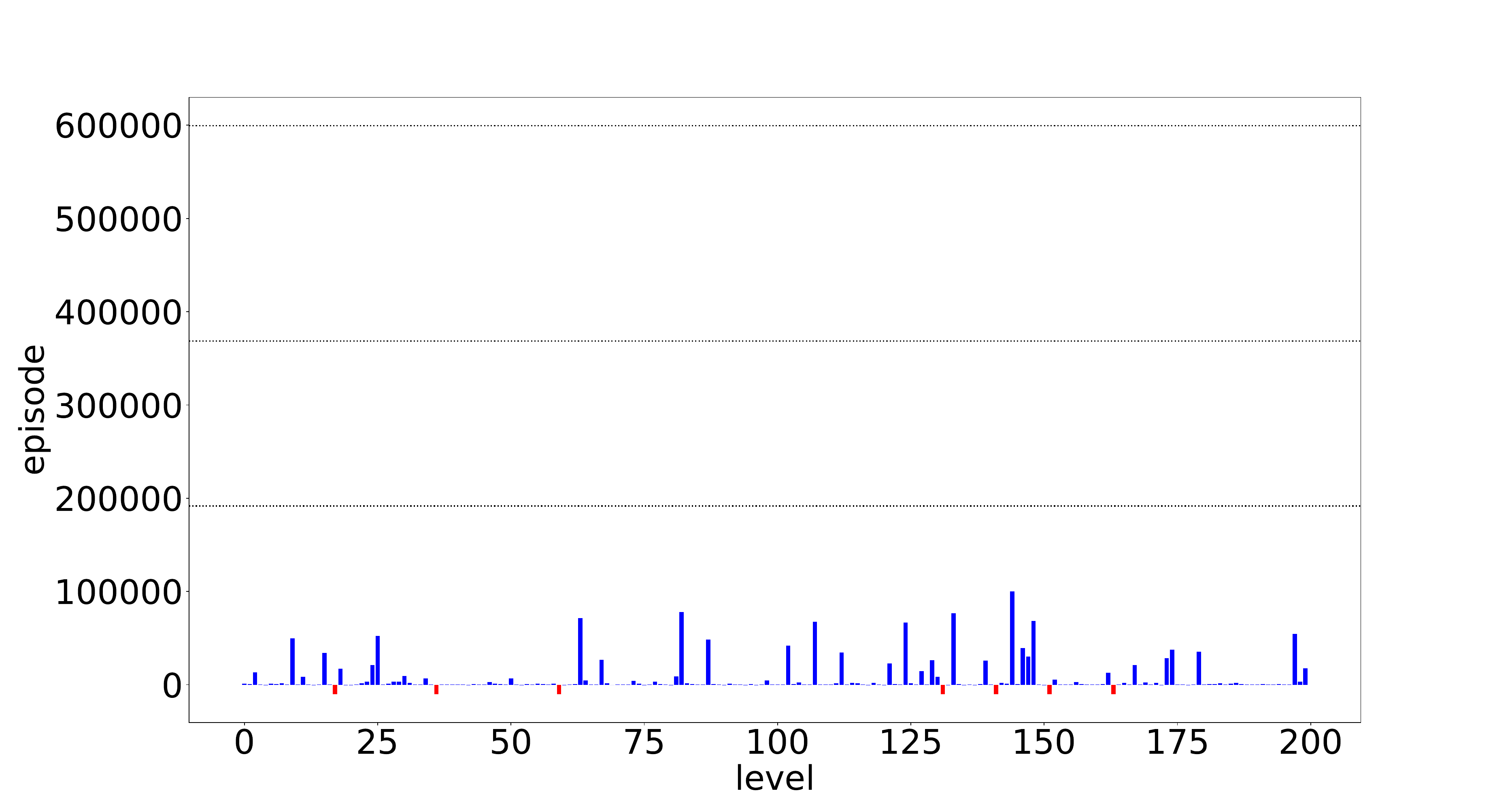} &
        \includegraphics[width=0.3\textwidth]{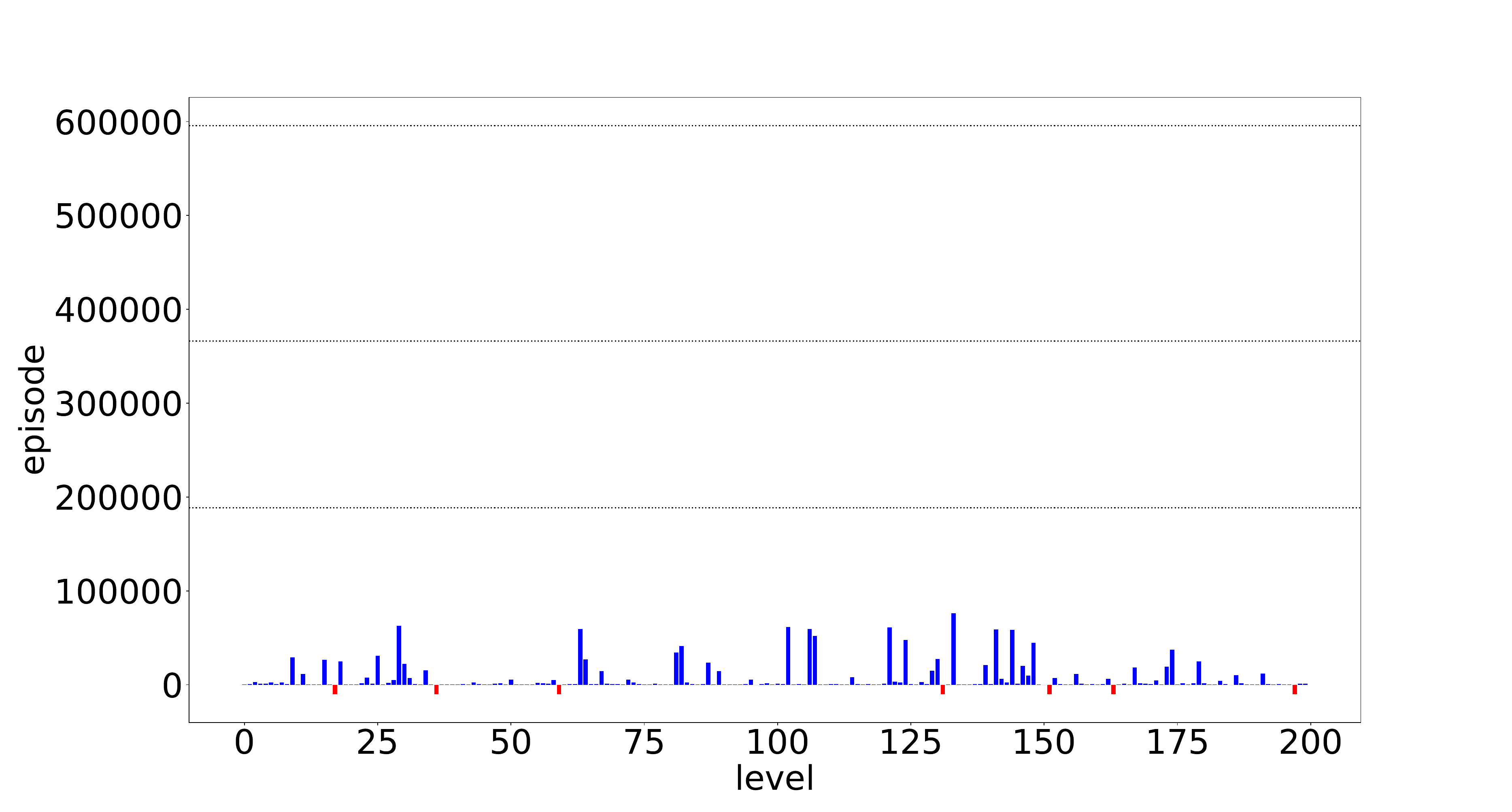}
        \\        
        $\pi_0^{climber}$ & $\pi_1^{climber}$ & $\pi_2^{climber}$
        \\
        \includegraphics[width=0.3\textwidth]{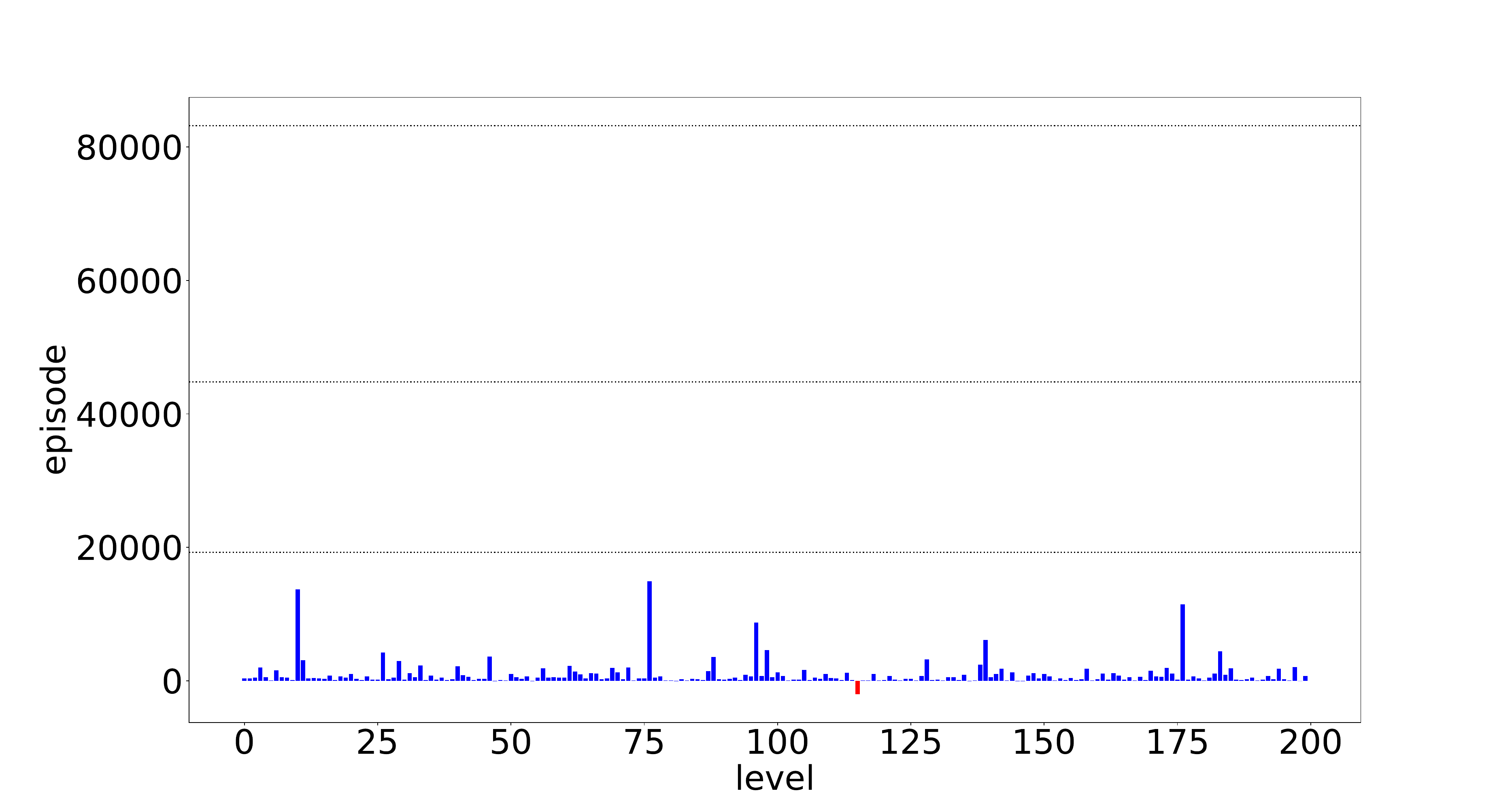} &
        \includegraphics[width=0.3\textwidth]{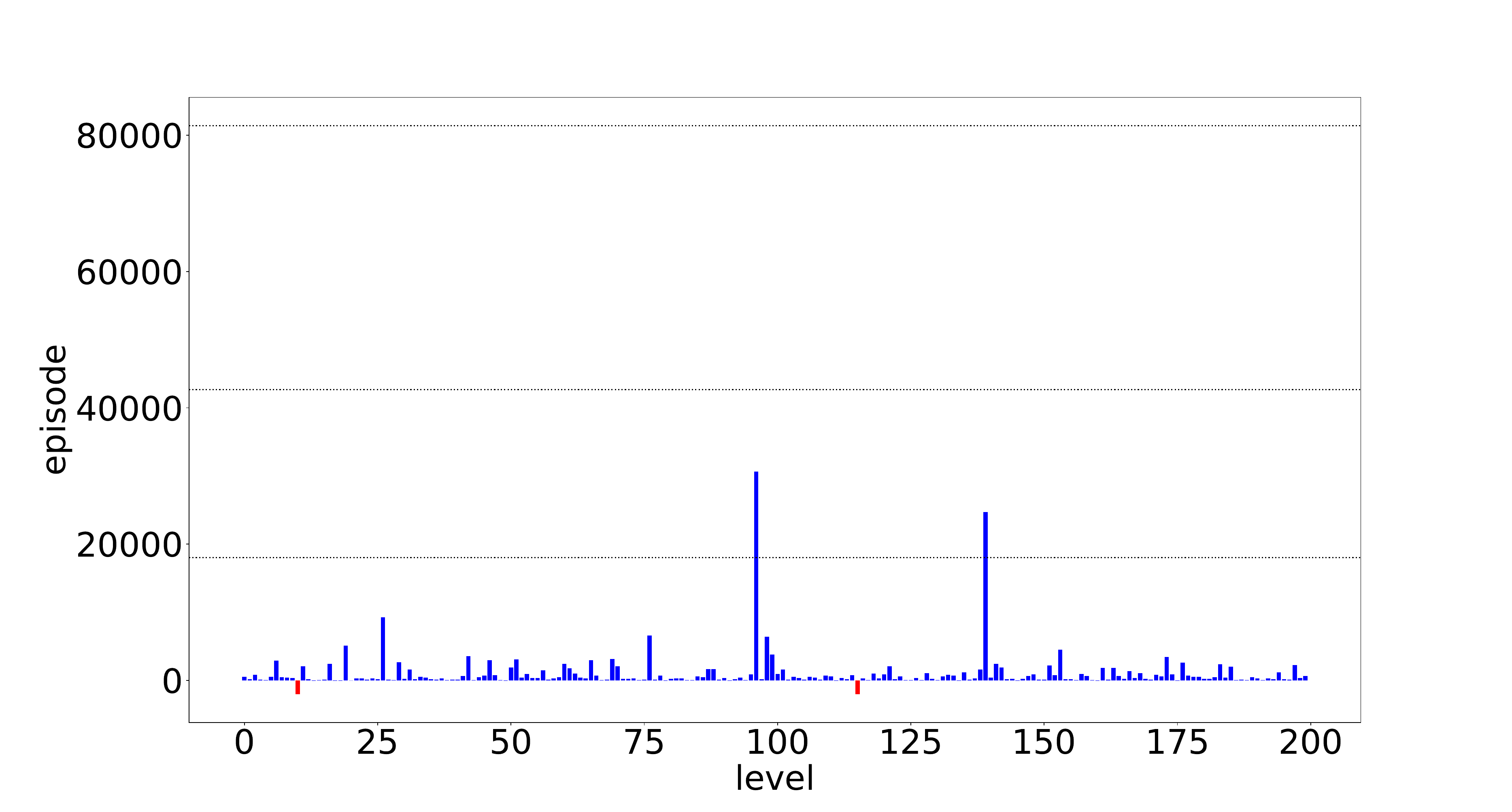} &
        \includegraphics[width=0.3\textwidth]{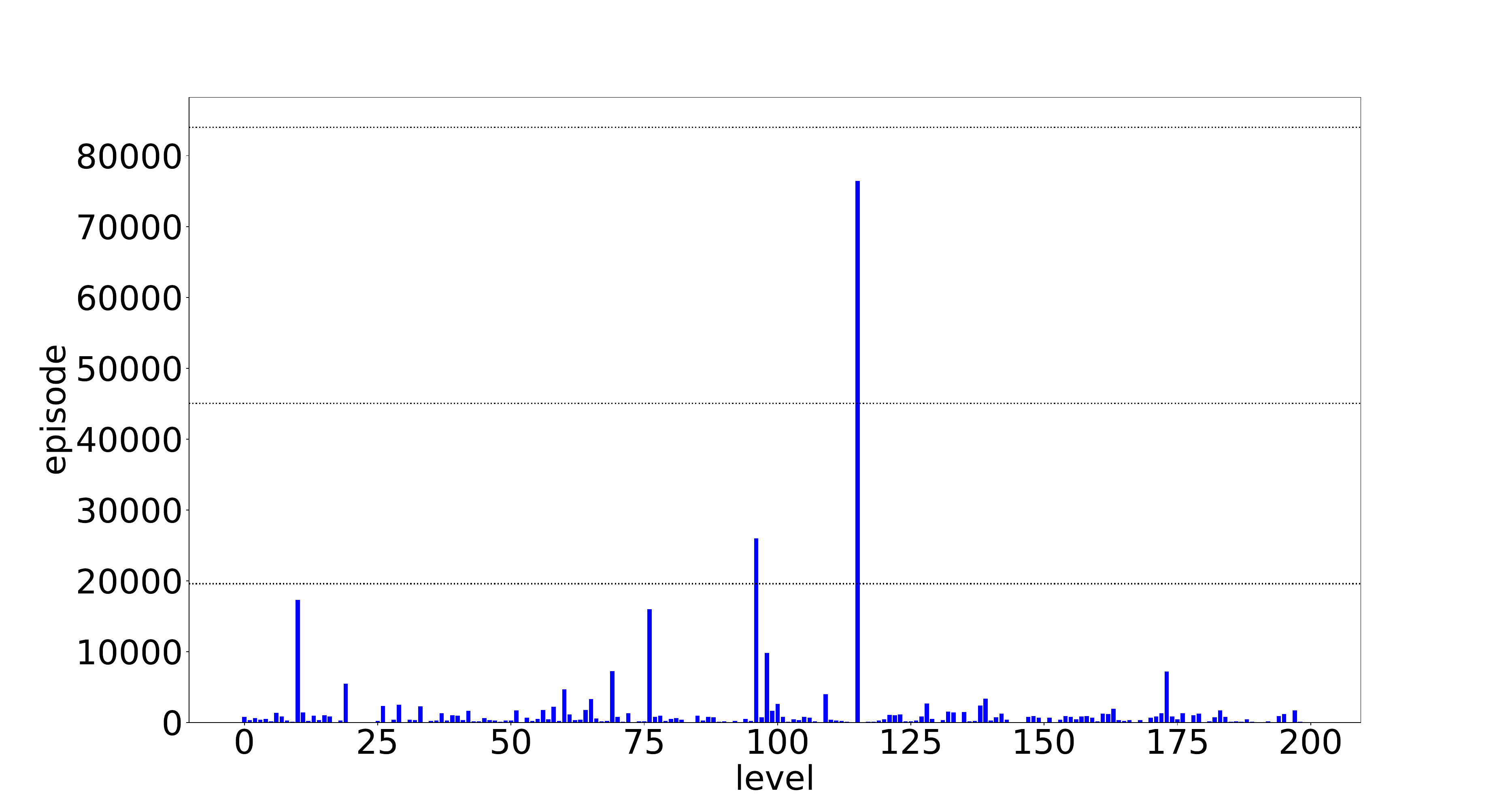} 
        \\
        \multicolumn{3}{c}{(b) Resulting policies after training the agent with \textit{Buffer\_1ep\_higherQuality}}
    \end{tabular}
    \caption{Graphical representation of the first time a non-zero return is obtained at each of the 200 training levels in \texttt{Ninja} and \texttt{Climber} through training. Dashed lines denote milestones at 7.5M, 15M, and 25M steps. Levels unsolved throughout the training are highlighted in red, indicating the persistent challenge in mastering these levels.
    Each subfigure represents the results of different simulation runs (0,1,2) when training the agent concurrently with RL and IL considering (a) \textit{Buffer\_1ep} and (b) \textit{Buffer\_1ep\_higherQuality}.
    }
    \label{fig:procgen_succesful_levels_extended}
\end{figure*}

Figure \ref{fig:procgen_succesful_levels_extended} shows the first time a successful trajectory was collected at each of the 200 training levels through a training of 25M time steps. This plot allows assessing the complexity exhibited by each level, and how their demonstrations might impact on the agent's training process. It can be seen that some levels are solved early during training (those represented with low amplitude bars), whereas others are more complex to solve and require more interactions. Some levels are never solved (shown in red). 

On the one hand, when using \textit{Buffer\_1ep} to concurrently train the agent with IL and RL, we see that\footnote{The agent's learned policy for run $r$ is denoted as $\pi_r$.}:
\begin{itemize}
    \item In \texttt{Ninja} the number of unsolved levels is between 6 to 8 levels, showing a clear improvement with each resulting policy:
    \vspace{2mm}
    \begin{itemize}
        \item $\mathcal{L}_{non-solved}^{ninja}|\pi_0 = \{17,59,121,141,144,151,163,174\}$
        \item $\mathcal{L}_{non-solved}^{ninja}|\pi_1 = \{59,121,127,131,144,148,163\}$
        \item $\mathcal{L}_{non-solved}^{ninja}|\pi_2 = \{59,131,133,141,144,163\}$
    \end{itemize}
    \vspace{2mm}
    \item Similarly, in \texttt{Climber} the agent is incapable of solving 1 to 2 levels:
    \vspace{2mm}
    \begin{itemize}
        \item $\mathcal{L}_{non-solved}^{climber}|\pi_0 = \{26, 115\}$
        \item $\mathcal{L}_{non-solved}^{climber}|\pi_1 = \{115\}$
        \item $\mathcal{L}_{non-solved}^{climber}|\pi_2 = \{115\}$
    \end{itemize}
\end{itemize}

On the other hand, when using a allegedly higher quality buffer, \textit{Buffer\_1ep\_higherQuality}, we can see that:
\begin{itemize}
    \item In \texttt{Ninja} the number of unsolved levels slightly decreases to 6-7 levels:
    \vspace{2mm}
    \begin{itemize}
        \item $\mathcal{L}_{non-solved}^{ninja}|\pi_0 = \{17,36, 59,131,151,163\}$
        \item $\mathcal{L}_{non-solved}^{ninja}|\pi_1 = \{17,36,59,131,141,151,163\}$
        \item $\mathcal{L}_{non-solved}^{ninja}|\pi_2 = \{17,36,59, 131,151,163,197\}$
    \end{itemize}
    \vspace{2mm}
    \item Conversely, in \texttt{Climber} the improvement is negligible, with 1 to 2 levels still unsolvable in some cases:
    \vspace{2mm}
    \begin{itemize}
        \item $\mathcal{L}_{non-solved}^{climber}|\pi_0 = \{115\}$
        \item $\mathcal{L}_{non-solved}^{climber}|\pi_1 = \{10,115\}$
        \item $\mathcal{L}_{non-solved}^{climber}|\pi_2 = \emptyset$
    \end{itemize}
\end{itemize}

\end{appendix}

\end{document}